\documentclass[twoside,11pt]{article}

\usepackage{blindtext}
\usepackage[utf8]{inputenc} 
\usepackage[T1]{fontenc}    
\usepackage{url}            
\usepackage{booktabs}       
\usepackage{amsfonts}       
\usepackage{nicefrac}       
\usepackage{microtype}      
\usepackage{multirow}
\usepackage{array}
\usepackage{graphicx}
\usepackage{subcaption}
\usepackage{xurl}
\usepackage[dvipsnames]{xcolor}
\usepackage{amsmath}
\usepackage{longtable}
\usepackage{graphicx}
\usepackage{booktabs}
\usepackage{ragged2e}
\usepackage{textcomp}
\usepackage{lastpage}
\usepackage{dmlr2e}
\usepackage{listings}
\usepackage{caption} 
\usepackage{tipa}
\usepackage{tabularx}
\usepackage{comment}
\usepackage{hyperref}
 \usepackage{float}

\ShortHeadings{DARai: Daily Activity Recordings for AI and ML applications}{Kaviani, Yarici, Kim, Prabhushankar, AlRegib, Solh and Patil}
\firstpageno{1}

\begin{document}

\title{Hierarchical and Multimodal Data for Daily Activity Understanding}
\author{\name Ghazal Kaviani \textsuperscript{1} \email gkaviani@gatech.edu
       \AND
       \name Yavuz Yarici \textsuperscript{1}  \email yavuzyarici@gatech.edu 
       \AND
       \name Seulgi Kim \textsuperscript{1} \email seulgi.kim@gatech.edu 
       \AND
       \name Mohit Prabhushankar \textsuperscript{1} \email mohit.p@gatech.edu 
       \AND
       \name Ghassan AlRegib \textsuperscript{1} \email alregib@gatech.edu  \thanks{Corresponding author.}
       \AND
       \name Mashhour Solh\textsuperscript{2} \email mashhour@amazon.com 
       \AND
       \name Ameya Patil \textsuperscript{2} \email ameyapat@amazon.com \AND
       \addr \textsuperscript{1} OLIVES at the Center for Signal and Information Processing CSIP, School of Electrical and Computer Engineering, Georgia Institute of Technology, Atlanta, GA, USA \textsuperscript{1} \\
       \addr \textsuperscript{2} Amazon Lab126, San Francisco, CA, USA }
       
\def\openreview{https://openreview.net/forum?id=0E7a6RTBqM}
\editor{Sergio Escalera}
\maketitle
\begin{abstract}
\textbf{D}aily \textbf{A}ctivity \textbf{R}ecordings for \textbf{a}rtificial \textbf{i}ntelligence (\texttt{DARai}, pronounced /Dahr-ree/), is a multimodal, hierarchically annotated dataset constructed to understand human activities in real-world  settings. \texttt{DARai} consists of continuous scripted and unscripted recordings of 50 participants in 10 different environments, totaling over 200 hours of data from 20 sensors including multiple camera views, depth and radar sensors, wearable inertial measurement units (IMUs), electromyography (EMG), insole pressure sensors, biomonitor sensors, and gaze tracker. To capture the complexity in human activities, \texttt{DARai} is annotated at three levels of hierarchy: (i) high-level activities (L1) that are independent tasks, (ii) lower-level actions (L2) that are patterns shared between activities, and (iii) fine-grained procedures (L3) that detail the exact execution steps for actions.
The unscripted nature of \texttt{DARai} enables the collection of action counterfactuals, defined as observed alternative executions of the same activity under different conditions (e.g., lifting a heavy versus a light object). Experiments with various machine learning models showcase the value of \texttt{DARai} in uncovering important challenges in human-centered applications. Specifically, we conduct unimodal and multimodal sensor fusion experiments for recognition, temporal localization, and future action anticipation across all hierarchical annotation levels. To showcase the shortcomings of individual sensors, we conduct domain-variant experiments that are possible because of \texttt{DARai}'s multi-sensor and and its inclusion of action counterfactuals, i.e., observed alternative executions of the same activity. The code, documentation, and dataset is available at the dedicated \href{https://alregib.ece.gatech.edu/software-and-datasets/darai-daily-activity-recordings-for-artificial-intelligence-and-machine-learning/}{\texttt{DARai website}}.\\


\end{abstract}


\section{Introduction}
\label{Introduction}

\begin{figure}[ht]
    \centering
    \scriptsize
    \includegraphics[width=0.8\textwidth]{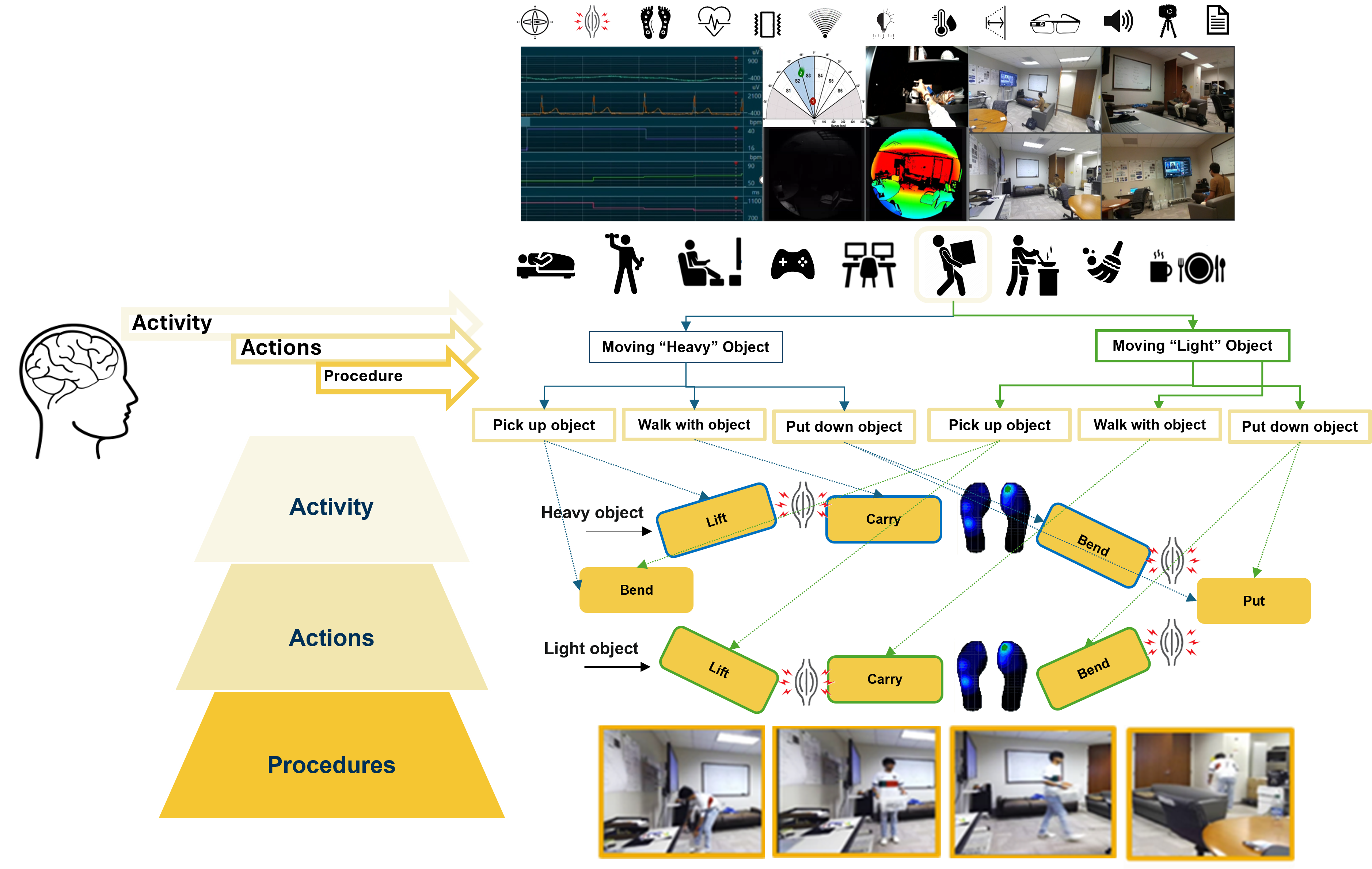}
    \caption{ \scriptsize \texttt{DARai} multimodal data collected to capture and reflect \textit{action counterfactual} variations across modalities. Hierarchical structure of \texttt{DARai} data, demonstrated using an example activity: \textsf{Moving an Object}. Two variations are shown, moving a heavy box (blue) and a light box (green), along with their corresponding lower hierarchy levels. For instance, variations in insole pressure and forearm muscle interactions are observed when lifting or carrying a box.}
    \label{fig:main-figure}
\end{figure}


\paragraph{Challenges in visual data understanding} Visual information has traditionally driven research in understanding human activities. Progress in computer vision techniques for tasks such as recognition and scene understanding~\citep{https://doi.org/10.1111/exsy.13680}, image classification and detection~\citep{carion2020end}, image segmentation~\citep{quesada2024benchmarking}, video action recognition~\citep{ma2019ts}, and traffic surveillance~\citep{kokilepersaud2023focal} are centered around visual data. However, in real-world settings, visual processing methods face significant practical and ethical constraints, including partial occlusions \citep{gunasekaran2023now}, acquisition and environmental errors~\citep{Temel2017_NIPSW,temel2019traffic,prabhushankar2022introspective, kokilepersaud2023exploiting}, sensitivity to angles and viewpoints~\citep{ozturk2024intelligent,dong2022viewfool,garg2022semantic}, privacy concerns \citep{mujirishvili2023acceptance,ravi2024review}, and the impracticality of maintaining fixed or unobtrusive camera perspectives. In contrast, widely available wearable sensors like inertial measurement units (IMUs), electromyography (EMG), and pressure insoles capture body-centric signals unaffected by camera placement. Furthermore, they provide additional information than visual data alone.

\paragraph{Value in multimodal and hierarchical data acquisition} Consider the activity of \textsf{Moving an object} demonstrated in Figure~\ref{fig:main-figure}. This simple activity can differ substantially in muscle interaction, foot pressure distribution, and physiological responses depending on whether the object is light or heavy. The differences in moving a heavy or light object may be visually imperceptible but are clearly evident in insole pressure data as demonstrated in Figure~\ref{fig:main-figure}. Therefore, integrating various data modalities help mitigate the limitations of a single modality approach~\citep{lin2023multimodality}. Such multimodal integration becomes essential given that human activities follow a hierarchical structure. The activity hierarchy starts with a specific goal which we term as Level 1 or L1 Activity. \textsf{Moving an object} in Figure~\ref{fig:main-figure} is an L1 activity. Next, intermediate actions, termed Level 2 or L2 are planned. These actions include \textsf{pick up}, \textsf{carry} and \textsf{put down} in Figure~\ref{fig:main-figure}. Finally, the specific procedures, including \textsf{bend} and \textsf{lift}, that complete each action are executed. Human-centric datasets require capturing multimodal and nested L1, L2, and L3 hierarchies that reflect both high-level and fine-grained level sensor signals and temporal organization~\citep{robinson2010development}.

We use \textit{action counterfactuals} to denote observed alternative executions of the same L1 activity. Two cases arise in \texttt{DARai}: (a) L1 activity and the L3 procedures are the same and the L2 action sequence varies (e.g., L1 \emph{Working on a computer} with L3 \emph{Open a program}, where L2 may be \emph{take a quiz}, \emph{attend a virtual meeting}, or \emph{type an email}; L1 \emph{Carrying an object} with L3 \emph{Pick up}, where L2 varies by moving a light, heavy, large, or small box), and (b) L1 is the same and the selection or order of L2 actions and their L3 procedures varies (e.g., L1 \emph{Using handheld smart devices}). Actions (L2) states what steps are taken and procedures (L3) specifies how each step is carried out; any overlap between L2 and L3 labels comes from decomposing of different complex activities into similar components, not from similar annotations for same activity. This definition is descriptive and does not imply causal counterfactuals.

\paragraph{Dataset requirements for daily activity understanding} Research on human-centered applications like robotics~\citep{chen2021framework}, assistive technologies~\citep{s21186037}, health monitoring~\citep{Wang_Wang_Luo_2023}, and disease diagnosis~\citep{prabhushankar2022olives} utilize visual data and additional sparse data modalities. These additional modalities include skeleton data~\citep{10143178}, clinical and biomarker data~\citep{kokilepersaud2023clinically}, and more recently language models~\citep{Damen_2018_ECCV} among others. However, commonly used datasets in daily activity and action understanding emphasize single-modality~\citep{byrne2023fine} or scripted scenarios~\citep{stein2013combining} and rarely incorporate hierarchical annotations across multiple sensors~\citep{liu2019ntu}. Addressing these gaps by designing datasets that feature continuous, unscripted recordings from both visual and non-visual sensors can enable researchers to tackle significant challenges related to privacy, multimodal learning strategies, and hierarchical activity recognition, thus promoting robust, real-world human activity understanding.

In this paper, we propose the Daily Activity Recordings for artificial intelligence (\texttt{DARai}) dataset. \texttt{DARai} is an open source, multimodal, hierarchical, and continuously recorded dataset. \texttt{DARai} marks the largest available dataset interms of the number of sensors and data modalities. The contributions of \texttt{DARai} are listed as follows. 
\begin{enumerate} 
\item \texttt{DARai} is a comprehensive daily activity dataset consisting of $200$ hours of recordings across $20$ data modalities from $12$ sensors. \texttt{DARai} is collected from $50$ participants in five different kitchens and living spaces, totaling 10 unique environments.
\item \texttt{DARai} is annotated to have $160$ classes across three levels of hierarchies - activities (L1), actions (L2), and procedures (L3). Furthermore, an action level language description of the L1 data is provided. 
\item The hierarchical structure and continuously recorded data allows construction of action counterfactual scenarios- empirically observed alternative executions of the same activity- and long-range temporal dependencies, thereby showcasing the utility of specific sensors under different conditions.
\item \texttt{DARai} is anonymized, preprocessed and curated for machine learning applications and is open sourced at \href{https://ieee-dataport.org/open-access/darai-daily-activity-recordings-ai-and-ml-applications}{IEEE Dataport}~\citep{ecnr-hy49-24}.
\item Benchmarks for activity, action, and procedure recognition, action and procedure temporal segmentation, and action anticipation are provided on \texttt{DARai} and the codes are available on the \href{https://alregib.ece.gatech.edu/software-and-datasets/darai-daily-activity-recordings-for-artificial-intelligence-and-machine-learning/}{project page}. Across all tasks, our baseline set includes transformer-based models alongside convolutional and recurrent approaches.
\end{enumerate} 

\paragraph{Challenges in \texttt{DARai} data collection} Gathering continuous, real-world data across multiple environments using various sensor platforms is a significant challenge. Notably, each of the $12$ sensors used in \texttt{DARai} and listed in Table~\ref{tab:sensors}, have different sampling rates and timestamps' protocol, making device synchronization challenging. This is addressed by implementing a multi-threaded system with a global timestamp signal to align all start and end time of data streams effectively. More information on sensor setup and calibration is provided in Appendix ~\ref{Appendix: Sensor-Setup}. We achieve a synchronization drift of 1 ms in 24-hour timespan. Depth sensors overheat during long recording sessions, reducing depth estimation quality and requiring periodic cool downs. Wearable sensors are connected wirelessly, and sometimes encounter connectivity drops. Such connectivity issues are overcome with post-recording recovery from device internal memory, thereby necessitating using sensors that provide significant internal memory. A further challenge is the large volume of data. Each participant in a session (spanning between 30 minutes to 1 hour) generates up to 300 GB of data. This necessitates local storage before transferring processed subsets to the servers. Details on data size and format are available in Appendix ~\ref{Appendix: Structure}.

The novelty of \texttt{DARai} is compared against existing datasets in Section~\ref{Section:related-work}. The specifics of data collection, annotation, and processing are described in Section~\ref{Section:Dataset}, followed by machine learning experiment setup and benchmarks in Sections~\ref{Section:Setup-ML},~\ref{Section:Visual-Robus-Limit},~\ref{Section:Hierarchical}, and~\ref{Section:Temporal}.

\section{Related Work}
\label{Section:related-work}

In this section, we examine $28$ multimodal and hierarchical datasets to establish the novelty of \texttt{DARai} dataset.

\paragraph{Multimodal Datasets}
\begin{table}[htb]
\scriptsize
\caption{Datasets Comparison Summary}
\label{tab:dataset_multi}
\centering
\resizebox{\textwidth}{!}{%
\begin{tabular}{m{0.5cm} m{3.5cm} c c c m{4cm} m{2cm}}
\toprule
\textbf{Setup} & \textbf{  Dataset} & \textbf{\# Classes} & \textbf{\# Subjects}  & \textbf{\# Modality} & \textbf{Activities} & \textbf{Multiple \newline Environment}  \\
\toprule
\multirow{16}{*}{\rotatebox{90}{Scripted Activities}} & CMU-MMAC \citep{de2009guide} & 5 & 18 & 4 & Food preparation  & No  \\
& UT-Kinect \citep{xia2012view} & 10 & 10 &  3 & Daily arm motions  & No \\
& Multiview 3D Event \citep{wang20143d} & 8 & 8& 3 & Indoor actions  & No \\
& UTD-MHAD \citep{chen2015utd} & 27 & 8 &  3 & Exercise, Arm motions  & No  \\
& Stanford ECM \citep{nakamura2017jointly}& 24 & 10  & 3 & Physical activities  & No  \\
& Daily Intention \citep{wu2017anticipating}& 34 & 12  & 2 & Indoor daily motions  & No  \\
& UI-PRMD \citep{vakanski2018data}& 10 & 10 &  3 & Physical rehabilitation  & No  \\
& NTU RGB+D 120 \citep{liu2019ntu}& 120 & 106  & 4 & Indoor activities  & No \\
& MMAct \citep{kong2019mmact}& 37 & 20 &  3 & Indoor daily actions  & Yes \\
& UESTC RGB-D \citep{ji2019large}& 118 & 40 & 2 & Indoor aerobic exercise  & No  \\
& OREBA \citep{rouast2020oreba} & 4 & 102 &  3 & Eating  & No  \\
& LaRa \citep{niemann2020lara}& 8 & 14 & 3 & Logistic and packaging  & No  \\
& BON \citep{ehzr-w794-21} & 18 & 25 & 1 & Office activities  & Yes \\
& ActionSense \citep{delpreto2022actionsense} & 20 & 10  & \underline{8} & Kitchen activities  & No  \\
& StrokeRehab \citep{kaku2022strokerehab} & 24 & 41 & 2 & Stroke rehabilitation  & No  \\
& MPHOI-72 \citep{qiao2022geometric} & 13 & 5 & 2 & Indoor Human and Object Interaction  & No  \\
\midrule
\multirow{2}{*}{\rotatebox{90}{Hybrid}}
& Breakfast Actions \citep{kuehne2014language} & 10 & 52 &  1 & Food preparation  & No  \\
& \textcolor{blue}{\texttt{DARai (2026)}} & \textbf{160} & 50 &  \textbf{20} & Indoor activities, Office tasks, Kitchen and Household activities  & \textbf{Yes}  \\

\midrule
\multirow{4}{*}{\rotatebox{90}{Unscripted Activities}}
 & Epic kitchen-100 \citep{Damen_2018_ECCV} & 97 & 37 & 3 & Kitchen activities  & No  \\

& Ego4D \citep{grauman2022ego4d} & Varies* & 931 &  4 & Daily activities \newline (household, outdoor, workplace, etc.)  & Yes  \\
& Ego-Exo4D \citep{grauman2023ego} & Varies* & 839 &  6 & Indoor skilled human activities \newline (sports, music, dance, etc.)  & Yes  \\
& CAP \citep{byrne2023fine}& 512 & 780 &  1 & Indoor activities  & Yes  \\ \\ \\
\bottomrule
\end{tabular}%
}
\hfill 
\noindent\textit{*Varies} indicates that the exact set of activities present in the dataset is unknown and can only be determined by accessing the annotations.
\end{table}
Several widely used multimodal datasets are shown in Table~\ref{tab:dataset_multi}, organized by their data collection setups: scripted activities, unscripted activities, or a hybrid approach. The data collection setup influences trade-offs between the number of activity classes, subjects, sensors setup and environments, as well as activity categories.
\begin{itemize}
    \item Scripted Activities: Participants receive instructions to perform scripted activities. Generally, participants carry out similar activities. Such setup limits the variety in activity scenarios and environments, enabling the use of stationary and cumbersome sensors (e.g., motion capture suits).
    \item Unscripted Activities: Participants act freely without a scripted set of activities. Such a set-up results in unscripted or "in-the-wild" data. This setup often results in diverse set of activities and environments. Though it is difficult to maintain both consistency and variety in the sensor setup under these conditions.
    \item Hybrid: Participants are instructed to perform all or a subset of high-level activities from a predefined list while acting freely without a scripted plan for the lower level intermediate steps. Hybrid setups allow for more natural recordings of human activity in diverse environments while ensuring a diverse set of activities and supporting larger sensor setup.
\end{itemize}


In addition to the categorization provided in the columns of Table~\ref{tab:dataset_multi}, \texttt{DARai} includes activities of varying lengths, from 30 seconds to over 5 minutes. This is opposed to existing datasets that provide short activity clips, without accounting for long-term temporal dependencies. 



\paragraph{Hierarchical Datasets}




\begin{table}[htb]
\centering
\scriptsize
\caption{ \scriptsize Comparison of Taxonomy in Hierarchical Datasets}
\label{table:taxonomy_hierarchy}
  \resizebox{\textwidth}{!}{%
\begin{tabular}{l >{\raggedright\arraybackslash}p{4cm} >{\raggedright\arraybackslash}p{4cm}>{\raggedright\arraybackslash}p{5cm}}
\toprule
\textbf{Setup} & \textbf{Dataset} & \textbf{Hierarchy Structure} &\textbf{Hierarchy Construction Method}\\ 
\toprule
\multirow{6}{*}{Post-Collection Hierarchy Design}
 & 50 Salads \citep{stein2013combining} & High-level activity, Low-level activity & Annotator decision\\
 & Breakfast Actions \citep{kuehne2014language}  & Coarse actions, Fine-grained actions & Sentences from audio transcription
 \\
& YouCook2 \citep{zhou2018towards}  & Activity , Steps & Sentences from subtitles
 \\
&  FineGym \citep{shao2020finegym} & Event, Set, Element 
& Annotator decision, Decision-tree based process
\\
& MoMa \citep{luo2021moma}  & Activity , Sub-Activity, Atomic Actions  & Annotator decision
 \\

& FineSports \citep{Xu_2024_CVPR} & Action, Sub-Action & Annotator decision, MixSort-OC (\citep{cui2023sportsmotlargemultiobjecttracking})
\\

& Ego4D Goal-Step \citep{song2024ego4d}  & Goal, Steps, Sub-Steps& Annotator decision, wikiHow
\\

\midrule
\multirow{2}{*}{Predefined Hierarchy Design}
& Assembly 101 \citep{sener2022assembly101}  & Coarse actions, Fine-grained actions& Verb and noun from activity labels
\\

& \textcolor{blue}{\texttt{DARai (2024)}}  & Activity, Action, Procedure& Predefine decomposition rules
 \\
\bottomrule
\end{tabular} %
}
\end{table}

Unlike object detection and image classification datasets, which benefit from standard taxonomies like WordNet~\citep{10.1145/219717.219748}, human activity datasets often lack a widely adopted hierarchical or semantic framework. Existing datasets group activities under broad scenarios (e.g., daily tasks, exercise, kitchen activities) without defining multi-level hierarchies beyond activity categories. Table \ref{table:taxonomy_hierarchy} surveys several datasets that offer explicit hierarchies and categorizes them by hierarchy setup. Post-collection hierarchy design implies that hierarchies are constructed at the annotation level after data is collected. In contrast, predefined hierarchy design allows data collection with hierarchies, inspired by activity planning and reasoning by humans. Predefined hierarchies allow variations in certain activities through multiple instances of the same activity with minor changes. For example, watching different TV programs, moving objects of varying sizes and weights, using a specific recipe to prepare food, or preparing food without a recipe are all examples of \textit{action counterfactuals}. \texttt{DARai} consists of these action counterfactuals performed by the same participant. Predefined hierarchical structure supports more detailed analyses of complex tasks by breaking them down into smaller actions and procedures~\citep{sloutsky2010perceptual}.

\section{Dataset Description and Annotation}
\label{Section:Dataset}
\subsection{Data collection process and considerations}
\begin{table}[htb]
\centering
\scriptsize
\caption{Sensors and Modalities Information}
\resizebox{\textwidth}{!}{%
\begin{tabular}{@{}lll l@{}}
\toprule
\textbf{Category} & \textbf{Sensor} & \textbf{Data Modalities} & \textbf{Views} \\ 
\toprule
\multirow{2}{*}{3rd POV \footnote{POV: point of view}} & Kinect Camera\cite{azure_kinect_dk} & RGB, Depth, IR & 2 \\
                         & Lidar Camera\cite{intel_realsense_l515_datasheet} & RGB, Depth, Depth Confidence, IR & 2/1 \\                          
                         \midrule

\multirow{6}{*}{1st POV} & Microphone & Audio & 1 \\
                         & Wearable IMU & Wrist acceleration, gyro and magnetic Field & 2 \\
                         & Wearable EMG & Forearm Muscle Signal & 2 \\
                         & Eye Tracker & RGB, Audio, Head IMU and Gaze & 1 \\
                         & Insole Pressure & Total foot pressure, Multi zone pressure & 1 \\
                         & Wearable Bio monitors & Respiration rate, ECG, Heartbeat rate and intervals & 1 \\ 
                         \midrule
\multirow{3}{*}{Ambient and Remote} & Stationary IMU & Surface vibration & 2 \\
                                    & Radar & Doppler cubes & 1 \\
                                    & Environmental Sensors & CO2, Humidity, Temperature, Light & 1 \\                 
                                    \midrule

BMI \footnote{BMI = body mass index}, Exhaustion Level & Self Report Form & Text & N/A \\
\bottomrule
\end{tabular}
}
\label{tab:sensors}
\end{table}
Fifty participants were recruited through a volunteer call and recorded in ten indoor environments across five locations (living spaces, home offices, kitchens) with variation in lighting, time of day, air conditioning, and background noise. Each session includes \textit{action counterfactual} instances where participants perform enforced variations (e.g., moving a heavy box instead of a light one), some of which are captured more clearly by specific sensors (e.g., insole pressure; see Fig.~\ref{fig:main-figure}). All recordings were collected under an Institutional Review Board (IRB)–approved protocol with written informed consent; privacy safeguards and use policy are summarized in Sec.~\ref{sec:ethics}.
Data is collected from 12 sensor devices, including Kinect, LiDAR, wearable IMUs on both hands, and stationary IMUs positioned on desks and floors in two environments, covering motion, physiological responses, and environmental conditions. Table \ref{tab:sensors} lists the sensors, their output modalities, and number of sensors used. Cameras are placed at different angles to capture standing and sitting postures, allowing natural movement rather than requiring participants to face a fixed camera, unlike other datasets~\citep{delpreto2022actionsense, liu2019ntu, qiao2022geometric, sener2022assembly101}. After the recordings, participants complete two questionnaires: the first gathers information about their familiarity with the tasks, exhaustion levels, and BMI, while the second assesses their experience with the sensors. More information about the participants and the survey results are provided in the Appendix~\ref{Appendix: Participants}.
\subsection{Annotations}

\begin{figure}[htbp]
  \centering
  \includegraphics[width=0.8\textwidth]{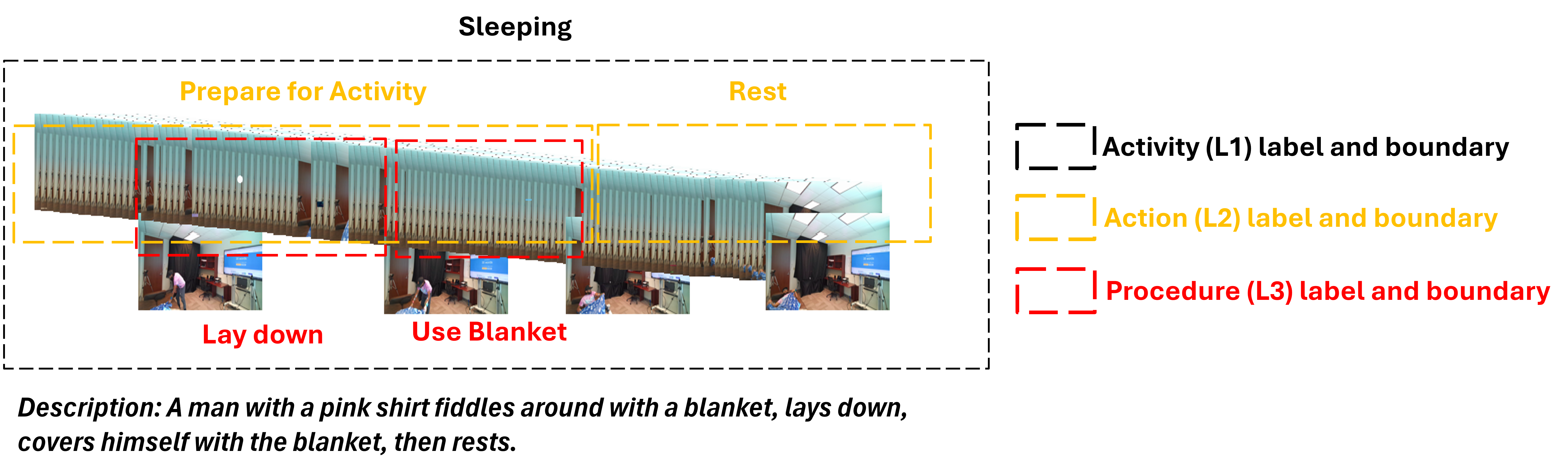}
  \caption{ \scriptsize A sequence of frames depicting an example of an annotated activity sample: L1 activity 'Sleeping' with a sequence of L2 actions: 'Prepare for Activity' and 'Rest', and L3 procedures: 'Lay Down' and 'Use Blanket'. Boundaries indicate the start and end times of each annotation level within the corresponding upper-level video sample.}
  \label{fig:hierarchy_seg}
  \vspace{0.5em} 
  \captionsetup{type=table, font=scriptsize} 
  \resizebox{0.6\textwidth}{!}{%
  \begin{tabular}{lcl}
    \toprule
    \textbf{Hierarchy Level} & \textbf{Number of Unique Labels} & \textbf{Example} \\
    \midrule
    $L\_1$: Activities & 18 & "Moving an Object" \\
    $L\_2$: Actions    & 44 & "Pick up object" , "Put down object" \\
    $L\_3$: Procedures & 98 & "Bend" , "Lift" , "Put" \\
    \bottomrule
  \end{tabular}%
  }
  \caption{ \scriptsize Number of unique labels at each level of hierarchy}
  \label{tab:unique_labels}
\end{figure}

The annotation description of \texttt{DARai} hierarchy is provided below:

\begin{itemize}
\item Level 1 (L1): Activities \\
High-level, independent tasks that broadly describe actions without specifying detailed goals or variations. For instance, \textsf{Moving an object} without defining its size or weight. Figure \ref{fig:hierarchy_seg} shows the hierarchy for the L1 activity \textsf{Sleeping}.

\noindent
\item Level 2 (L2): Actions \\
Actions are shorter segments that can frequently occur across multiple activities. A complete instance of an activity consists of a sequence of these actions. The order of actions or inclusion of specific steps may vary. For example, the action \textsf{Prep ingredients} might be skipped if ingredients are already prepared. Actions alone may lack sufficient context to definitively identify the activity. For instance, the L2 action \textsf{Prepare for Activity} in Figure \ref{fig:hierarchy_seg} could indicate either preparing to sleep or organizing the bed after waking, depending on temporal context.

\noindent
\item Level 3 (L3): Procedures \\ 

Procedures specify the detailed fine-grained context in which actions are performed, clarifying differences between individual instances of the same action. For instance, the L3 procedures for the action \textsf{Prepare for Activity} in Figure \ref{fig:hierarchy_seg} include \textsf{Lay down} and \textsf{Use blanket}.
\noindent
\item Natural language descriptions:\\ Natural-language description provided by human annotators, describing activities and interactions and the scene explicitly at the L2 level.
\end{itemize}
        

Some activities in the dataset do not span over all levels of the hierarchy. In Figure \ref{fig:hierarchy_seg}, the L2 action \textsf{Rest} does not contain any L3 procedures. This flexibility allows both procedural activities such as \textsf{Exercise} or \textsf{Playing a game}, and non-procedural activities including \textsf{Sleeping} or \textsf{Watching}, to fit within the same hierarchy structure and definition. In contrast, other hierarchical datasets, including~\citep{song2024ego4d,sener2022assembly101,Zhou_Xu_Corso_2018,10.1007/978-3-031-19778-9_38,Kuehne_2014_CVPR}, have only considered procedural activities in their design. We provide the number of unique labels at each level of the hierarchy, along with examples, in Table \ref{tab:unique_labels}.

Further details about the annotation workflow are provided in Appendix \ref{Appendix: Annotation}.

\subsection{Data Preprocessing}

The \texttt{DARai} multisensory setup records over 200 hours of data across 20 modalities from 12 devices, totaling more than 20~TB of raw recordings. Appendix~\ref{Appendix: Structure} and Table~\ref{tab:sensors spec} in Appendix~\ref{Hardware} list formats, sampling rates, and resolutions for each sensor.%

\textbf{Synchronization and alignment.} All streams are time-aligned using a global timestamp from the acquisition hosts systems and saved in addition to synchronized internal clocks, which provide consistent start/elapsed/end times for each stream. This allows segmentation and resampling without relying on device-specific timestamp protocols.%
\footnote{Global host timestamps and multi-threaded capture are described in Section~1 (Challenges) and Appendix D in more details.}

\textbf{Imputation and normalization.} Short missing intervals in physiological and wearable signals are imputed by linear interpolation; gaps longer than a modality-specific threshold are left as missing and marked by a binary mask released with the data. Each channel is standardized per subject with z-score scaling, 
$\tilde{x}_{s,c}(t)=\frac{x_{s,c}(t)-\mu_{s,c}}{\sigma_{s,c}}$, 
using that subject's recordings.

Processed data are segmented into activity-level samples using L1 boundaries, with each sample corresponding to one subject performing a labeled activity within a session. Visual frames are organized with zero-padded indices; time-series signals are stored as structured CSV files with aligned timestamps across all modalities. The dataset and codes are publicly available on \href{https://ieee-dataport.org/open-access/darai-daily-activity-recordings-ai-and-ml-applications}{IEEE Dataport} and \href{https://github.com/olivesgatech/DARai}{Project Repository}.

\subsection{Ethics and Responsible Use}\label{sec:ethics}
All recordings were collected under an IRB-approved protocol with written informed consent. Raw videos are anonymized frame by frame using a face detection pipeline~\citep{ORB-HD_deface_2023} and reviewed by annotators to flag errors; natural speech is anonymized to prevent speaker identification; Other personally identifiable information is removed except in explicitly approved demonstration cases. The dataset is released to support open-source research; users must not attempt re-identification or linkage with external sources and must follow the license terms. 

\section{Experiment Setup for ML Applications}
\label{Section:Setup-ML}
\begin{figure}[ht]
    \centering
    \captionsetup{font=small} 
    \footnotesize
    \includegraphics[width=0.8\textwidth]{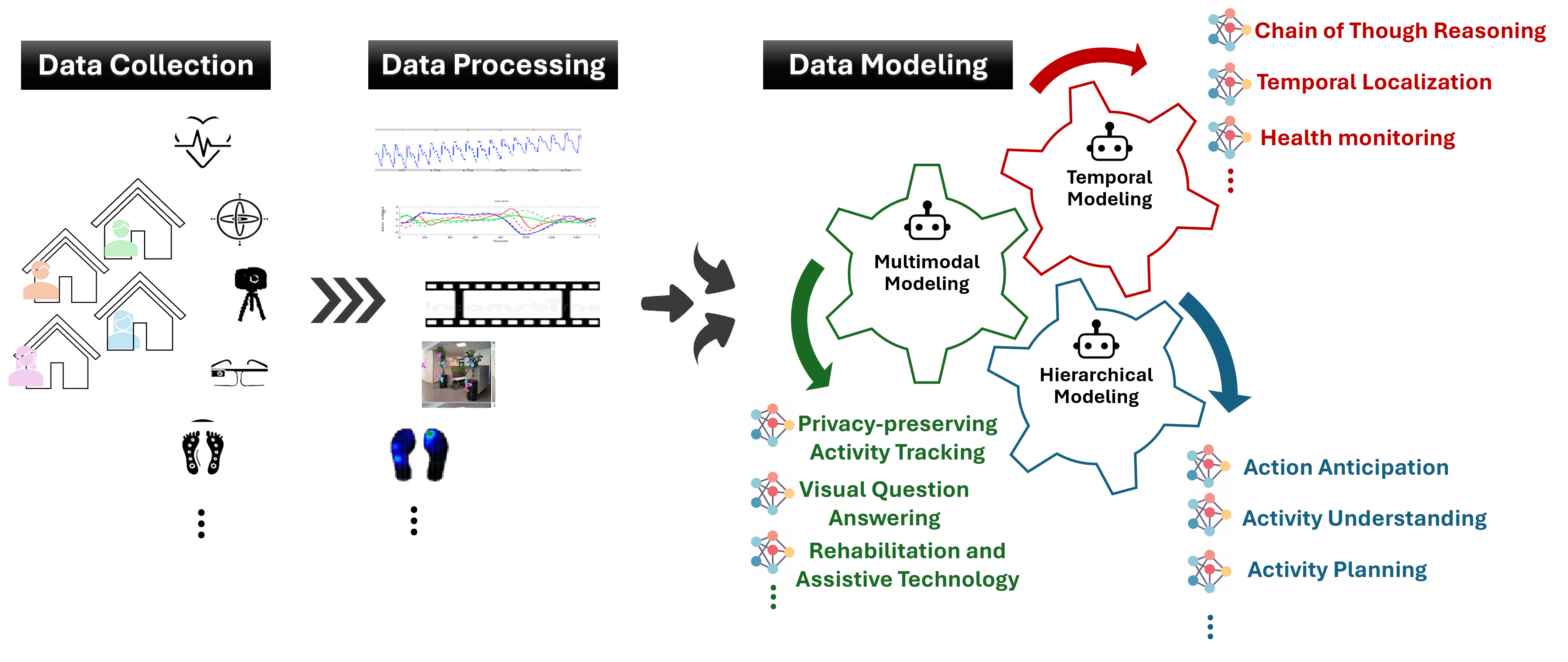}
    \caption{ \scriptsize Overview of machine learning applications using the multimodal hierarchical \texttt{DARai} dataset. This figure highlights key aspects explored in this work, including temporal modeling, multimodal modeling, and hierarchical modeling, as well as potential applications for the machine learning community as direct or combined downstream tasks of these approaches.}

    \label{fig:ML-applications}
\end{figure}
As shown in Figure~\ref{fig:ML-applications}, \texttt{DARai} allows multimodal, hierarchical, and temporal modeling of human daily activities. In this paper, we choose three exemplar applications and demonstrate the challenges and utility of learning from human daily activity data in the following three sections. Section~\ref{Section:Visual-Robus-Limit} evaluates the robustness of visual and wearable modalities under cross-view and cross-body settings, comparing fine-tuned and from-scratch deep learning models. Section~\ref{Section:Hierarchical} investigates unimodal and multimodal performance across the hierarchical levels of \texttt{DARai} for fine-grained activity recognition. Finally, Section~\ref{Section:Temporal} explores temporal dependencies in unscripted human activities, addressing both lower-level action localization and future activity anticipation. These tasks demonstrate \texttt{DARai}'s diverse downstream applications in machine learning.
\paragraph{Task interfaces and benchmark templates:}
To support applications beyond those evaluated in this paper, we release standardized data interfaces that work with the provided cross-subject train/test split and the temporally aligned multimodal streams. The interfaces include:
\begin{itemize}
  \item \textbf{Time-series}: loaders and configuration templates for \textit{supervised} and \textit{self-supervised} tasks on IMU, EMG, insole, biomonitor, and gaze signals, that can be used individually or jointly with other modalities.
  \item \textbf{RGB/Depth video}: loaders and templates for video understanding tasks including but not limited to recognition, temporal segmentation, and action anticipation. 
  \item \textbf{Video question answering}: video loaders and configuration files for open set question and answers.
\end{itemize}
The interfaces do not constrain model design or evaluation protocol. New tasks can be instantiated by adding a configuration file without modifying the dataset. The dataloaders and reference configurations are provided in the codebase.\footnote{\url{https://github.com/olivesgatech/DARai}}


\begin{figure}[ht]
\centering
\begin{minipage}{0.25\textwidth}
    \includegraphics[width=\linewidth]{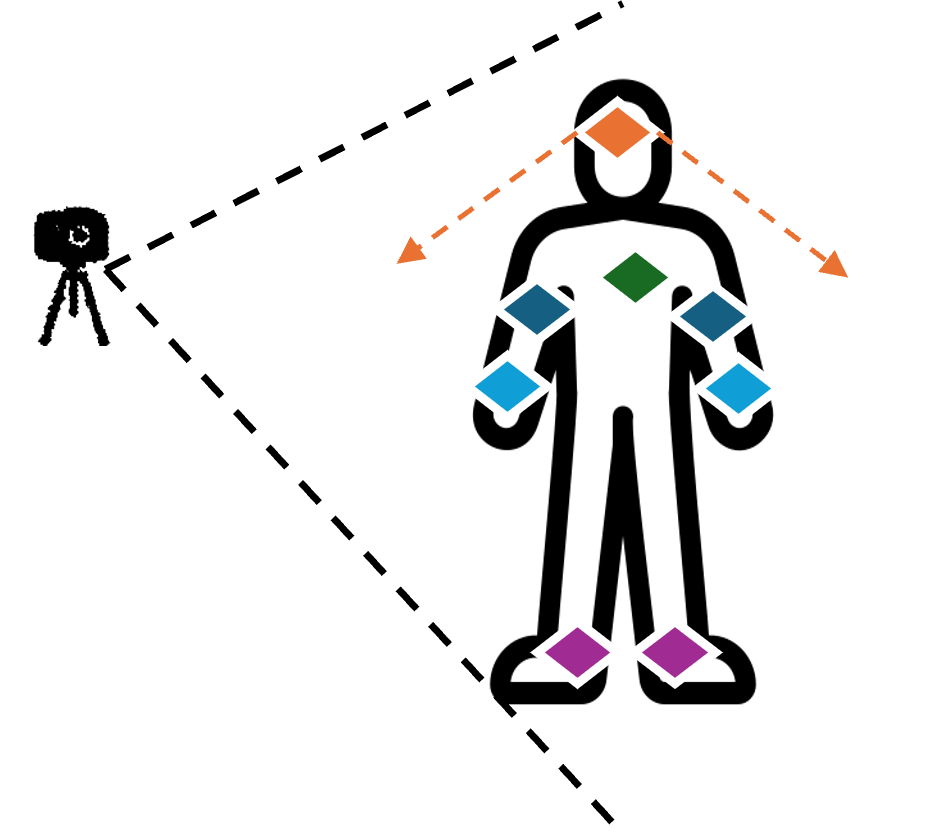} 
\end{minipage}%
\begin{minipage}{0.65\textwidth}
    \centering
    \footnotesize
    \resizebox{\linewidth}{!}{%
        \begin{tabular}{cp{1.5cm}cp{6cm}}
        \hline
        \textbf{Modality} & \textbf{Dimension} & \textbf{Resolution} & \multicolumn{1}{c}{\textbf{Information Provided}} \\ \hline
        RGB, D & 3 Channel, 1 Channel & 15 fps, 30 fps & Visual appearance, depth perception, spatial information \\ \hline
        \textcolor{Cyan}{IMU} & 9 Channel & 12 Hz, 200 Hz & Tracking motion dynamics and hand interactions \\ \hline
        \textcolor{MidnightBlue}{EMG} & 1 Channel & 2000 Hz, 4000 Hz & Hand muscle activity , griping and holding \\ \hline
        \textcolor{Purple}{Insole} & 8 Channel & 500 Hz & Insole pressure distribution and weight balancing, moving patterns \\ \hline
        \textcolor{OliveGreen}{Bio} & 5 Channel & 1000 Hz & Physiological signals \\ \hline
        \textcolor{Orange}{Gaze} & 2 Channel & 200 Hz & Visual focus tracking \\ \hline
        \end{tabular}%
    }
\end{minipage}
\caption{ \scriptsize Illustration of the unique information provided by various data modalities, highlighting the distinct insights each contributes to modeling human activity and state patterns.}
\label{fig:modality_info}
\end{figure}

\paragraph{Experiment Setup}
In our experiments, we use a subset of the data modalities available in \texttt{DARai}, including RGB and Depth data from different camera views, IMU from both hands, EMG of both forearm muscles, insole pressure from feet, five biomonitoring signals, and gaze tracking data. Figure \ref{fig:modality_info} summarizes the insights offered by each data modality, along with their dimensionality and resolution. The insights include motion dynamics, hand orientation, muscle activation patterns, foot pressure distribution and balance, physiological responses to varying tasks, and visual attention focus and patterns. Additional sensor data provided beyond this subset can further support research in IoT and home assistive robotics. For all experiments, we adopt a cross-subject evaluation method. Individual subjects are exclusively assigned to either the train or test set, thereby ensuring that no subject's data is used in both training and testing. To maintain consistency across all experiments, the sets of subject IDs used for training and testing remain fixed. For reproducibility and further customization, our specific train/test subject split configuration is provided as a Python script in our GitHub repository \footnote{\url{https://github.com/olivesgatech/DARai}}. For all classification tasks, we use class-wise top-1 accuracy. The class-wise accuracy and weighted F1 metric is chosen because the hierarchical structure of human activities does not necessarily include all actions or procedures in every instance, resulting in a nonuniform distribution of lower-level classes. More details about the class distribution can be found in Appendix \ref{Appendix: Annotation}.

We highlight the key observations from our data collection, sensor setup and each experiment in Table~\ref{tab:key_observations_summary}. A more comprehensive version of this table is provided in Appendix~\ref{Appendix: observation-challenges}. 
{\scriptsize
\begin{longtable}{l p{5in}}
\toprule
\textbf{Topic} & \textbf{Key Findings} \\
\midrule
\multirow{3}{1in}{\hyperref[sec:visual-training]{Visual Model Training Strategies}}
 & Finetuned \emph{visual} models \emph{consistently outperform} those trained from scratch, across all class categories and all three levels of hierarchy. \\
 & Within kitchen and living room subsets, \emph{depth} models trained from scratch \emph{outperform} their RGB counterparts. \\
 & At levels L2 and L3 of the hierarchy, \emph{depth} models continue to outperform \emph{RGB} models trained from scratch, likely due to the stability of depth-based spatial representations across hierarchy levels. \\
\midrule
\multirow{1}{1in}{\hyperref[sec:visual-training]{Environment Differences}}
 & Activity recognition accuracy within the living room is \emph{consistently higher} than those within the kitchen. \\
\midrule

\multirow{3}{1in}{\hyperref[sec:Viewpoint-placement-effect]{Viewpoint and Placement Effect}}
 & Visual modalities suffer a \emph{severe drop} in accuracy when tested on the cross view. \\
 & Non-visual modalities are \emph{robust} when the sensors capture symmetric interaction such as insole. \\
 & Wearable cross-body models show \emph{less performance degradation} when compared to cross-view camera models. \\
\midrule

\multirow{6}{1in}{\hyperref[sec:unimodal-strength]{Unimodal Sensor Strengths}}
 & Gaze data are effective at \emph{separating fine-grained activities} when other data modalities that rely on physical movements suffer from performance degradation. \\
 & Bio-monitoring data are good at \emph{identifying changes in physiology}, such as increased heart rate in \textsf{Playing video games} or reduced activity in \textsf{Sleeping}. However, they lack the resolution to separate activities with similar effort, such as different cooking activities. \\
\midrule

\multirow{5}{1in}{\hyperref[sec:hierachical-multimodal]{Hierarchical Granularity}}
 & Changes in performance across granularity levels are \emph{not} linear. \\
 & The visual model with the highest accuracy, \(\sim 56\%, \sim 59\%\), experiences a \emph{significant performance drop} when transitioning from L1 to L2 and L3. \\
 & Insole pressure and Hand EMG data modalities do \emph{not} exhibit a significant decline in performance from L1 to L2 as visual data does. \\
 & All modality combinations experience a decline in accuracy from L1 to L3, but the magnitude of degradation \emph{varies significantly} by sensor modality combination. \\
 & Gaze data improves L3 (procedure) activity recognition fused with other modalities. \\
 & Biomonitoring signals do not offer fine-grained insights at Level 3, even when combined with insole, EMG, and IMU data modalities. \\
\midrule

\multirow{2}{1in}{\hyperref[sec:app-localization]{Temporal Localization}}
 & In temporal activity localization, both the camera view and the length of the input video play a crucial role in accurately segmenting an untrimmed activity sample into lower-level segments. \\
 & Actions and procedures are not isolated events. They are interconnected components of larger sequences, with a strong dependence on the temporal context of those that come before and after. \\
\midrule

\multirow{5}{1in}{\hyperref[sec:anticipation]{Short- and \ Long-Horizon Anticipation}}
 & At L2 (Action level), a moderate observation period provides sufficient temporal context for immediate action anticipation, while excessive observation may introduce redundant or less relevant information. \\
 & Long-horizon anticipation at the L2 level, however, remains consistently lower and exhibits minimal improvement as the observation rate increases. This indicates that action-level predictions do not benefit significantly from extended temporal context, possibly due to the more independent nature of individual actions. \\
 & At the L3 (Procedure level), short-horizon anticipation improves as the observation rate increases, reaching optimal performance around a 70\% observation rate. \\
 & Long-horizon anticipation at this level consistently improves with increasing observation rates, eventually surpassing short-horizon performance beyond a 70\% observation rate. This trend suggests that fine-grained procedural activities exhibit stronger temporal dependencies, benefiting from longer observation windows that provide richer contextual information. \\
\bottomrule
\caption{Key observations}
\label{tab:key_observations_summary}
\end{longtable}
}

\section{Visual Data Robustness and Real-World Limitations}
\label{Section:Visual-Robus-Limit}

Visual sensing modalities, such as RGB and depth, are effective for human activity recognition~\citep{ma2019ts}.  However, such modalities suffer from robustness challenges in uncontrolled settings~\citep{hara2021rethinking,ponbagavathi2024probing}. In particular, existing benchmarks that rely on vision modalities are generally limited to coarse-grained actions and do not address real-life scenarios. In contrast, \texttt{DARai}, with diverse modalities and multiple levels of hierarchical labels, was designed and curated to provide a benchmark for real-life deployment. This section presents a benchmark for evaluating the robustness of visual activity recognition models across varying viewpoints and environments.

\subsection{Visual Data Benchmarking}
\label{sec:visual-training}

Given the rise of large-scale pretrained models, we examined whether pretraining compensates for the diversity of daily activities. We compared models trained from scratch on \texttt{DARai} with models first pretrained on Kinetics 400 and then fine-tuned on \texttt{DARai}. For this benchmark, we evaluated R3D~\citep{tran2018closer}, MViT-small~\citep{li2022mvitv2}, and Swin-tiny~\citep{liu2022video}. Table~\ref{tab:visual-only-l1} in Appendix~\ref{Appendix: ablation tables} compares their top-1 accuracy on combined and individual environment classes, while Table~\ref{tab:visual-only-l2l3} reports performance by camera view at hierarchy levels L2 and L3. Although accuracy is high at the activity level (L1), all models show a notable drop at finer-grained levels (L2, L3).

\begin{table}[htb]
\centering
\scriptsize
\captionsetup[table]{font=scriptsize}
\resizebox{\textwidth}{!}{%
\begin{tabular}{cccccccc}
\hline
\multicolumn{4}{c}{\textbf{Class Category}} & \multicolumn{4}{c}{\textbf{All}} \\ 
\cline{5-8}
\textbf{Data} & \textbf{Camera View} & \textbf{Level} & \textbf{Model} & \textbf{Accuracy - Finetuned} & \textbf{Accuracy - Trained from scratch} & \textbf{F1 - Finetuned} & \textbf{F1 - Trained from scratch} \\ \\
\multirow{12}{*}{\rotatebox[origin=c]{90}{RGB}} & $1$ & \multirow{6}{*}{L2} & \multirow{2}{*}{ResNet} & 0.46 $\pm$ 0.006 & 0.31 $\pm$ 0.01 & 0.458 & 0.295 \\
 & $2$ &  &  & 0.39 $\pm$ 0.01 & 0.25 $\pm$ 0.02 & 0.375 & 0.266\\
 & $1$ &  & \multirow{2}{*}{Mvit-s} & 0.43 $\pm$ 0.032 & 0.07 $\pm$ 0.004 & 0.418 & 0.303\\
 & $2$ &  &  & 0.38 $\pm$ 0.016 & 0.14 $\pm$ 0.07 & 0.382 & 0.12\\
 & $1$ &  & \multirow{2}{*}{Swin-t} & 0.42 $\pm$ 0.006 & 0.18 $\pm$ 0.02 & 0.411 & 0.143\\
 & $2$ &  &  & 0.33 $\pm$ 0.03 & 0.10 $\pm$ 0.021 & 0.325 & 0.064\\
& $1$ & \multirow{6}{*}{L3} & \multirow{2}{*}{ResNet} & 0.52 $\pm$ 0.006 & 0.35 $\pm$ 0.01 & 0.489 & 0.318\\
 & $2$ &  &  & 0.45 $\pm$ 0.007 & 0.28 $\pm$ 0.01 & 0.431 & 0.255\\
 & $1$ &  & \multirow{2}{*}{Mvit-s} & 0.53 $\pm$ 0.01 & 0.14 $\pm$ 0.07 & 0.507 & 0.070\\
 & $2$ &  &  & 0.44 $\pm$ 0.02 & 0.05 $\pm$ 0.003 & 0.418 & 0.073\\
 & $1$ &  & \multirow{2}{*}{Swin-t} & 0.47 $\pm$ 0.02 & 0.25 $\pm$ 0.02 & 0.446 & 0.213\\
 & $2$ &  &  & 0.34 $\pm$ 0.01 & 0.17 $\pm$ 0.016 & 0.321 & 0.138\\
\midrule
\multirow{12}{*}{\rotatebox[origin=c]{90}{Depth}} & $1$ & \multirow{6}{*}{L2} & \multirow{2}{*}{ResNet} & \multirow{6}{*}{-} & 0.56 $\pm$ 0.02 & \multirow{6}{*}{-} & 0.525\\
 & $2$ &  &  &  & 0.53 $\pm$ 0.01 &  & 0.487\\
 & $1$ &  & \multirow{2}{*}{Mvit-s} &  & 0.55 $\pm$ 0.02 &  & 0.560\\
 & $2$ &  &  &  & 0.53 $\pm$ 0.02 &  & 0.510\\
 & $1$ &  & \multirow{2}{*}{Swin-t} &  & 0.41 $\pm$ 0.01 &  & 0.401\\
 & $2$ &  &  &  & 0.40 $\pm$ 0.05 &  & 0.404\\
& $1$ & \multirow{6}{*}{L3} & \multirow{2}{*}{ResNet} & \multirow{6}{*}{-} & 0.42 $\pm$ 0.01 & \multirow{6}{*}{-} & 0.395\\
 & $2$ &  &  &  & 0.38 $\pm$ 0.01 &  & 0.344\\
 & $1$ &  & \multirow{2}{*}{Mvit-s} &  & 0.33 $\pm$ 0.11 &  & 0.290\\
 & $2$ &  &  &  & 0.29 $\pm$ 0.16 &  & 0.259\\
 & $1$ &  & \multirow{2}{*}{Swin-t} &  & 0.25 $\pm$ 0.12 &  & 0.211\\
 & $2$ &  &  &  & 0.24 $\pm$ 0.02 &  & 0.215\\
 \bottomrule
\end{tabular}%
}
\caption{ \scriptsize Top-1 accuracy results for visual data experiments using three model architectures on Action and Procedure level of DARai hierarchy. Since activities across subsets may share similar actions or procedures, they cannot be separated by activity class subset. Results are reported for all available actions and procedures classes. Fine-tuned experiments are performed using models pretrained on Kinetics 400 dataset.}
\label{tab:visual-only-l2l3}
\end{table}

Analyzing these results presented in Table~\ref{tab:visual-only-l1} and Table~\ref{tab:visual-only-l2l3} reveals a number of key observations:
\paragraph{Finetuned vs. Trained from scratch}
Pretrained models consistently outperform those trained from scratch at all hierarchy levels. This advantage is most evident when training data are limited, where fine-tuned models leverage diverse prior features for higher accuracy.
\noindent
\paragraph{ Environment-Specific Activities}
Recognition accuracy in the living room is consistently higher than in the kitchen. Even after fine-tuning on kitchen data, pretrained models still maintain a notable performance gap, suggesting stronger alignment with living-room-style activities.
\noindent
\noindent
\paragraph{Fine-Grained Levels (L2 and L3)}
At L2 and L3, trained from scratch depth models surpasses RGB counterparts, likely due to stable spatial representations. However, once using pretrained weights, RGB models achieve higher performance.

\subsection{Robustness to Real-World Conditions}
\label{sec:Viewpoint-placement-effect}
\subsubsection{Visual Cross-View Evaluation}

\begin{figure}[h!]
    \centering
    \includegraphics[width=0.7\textwidth]{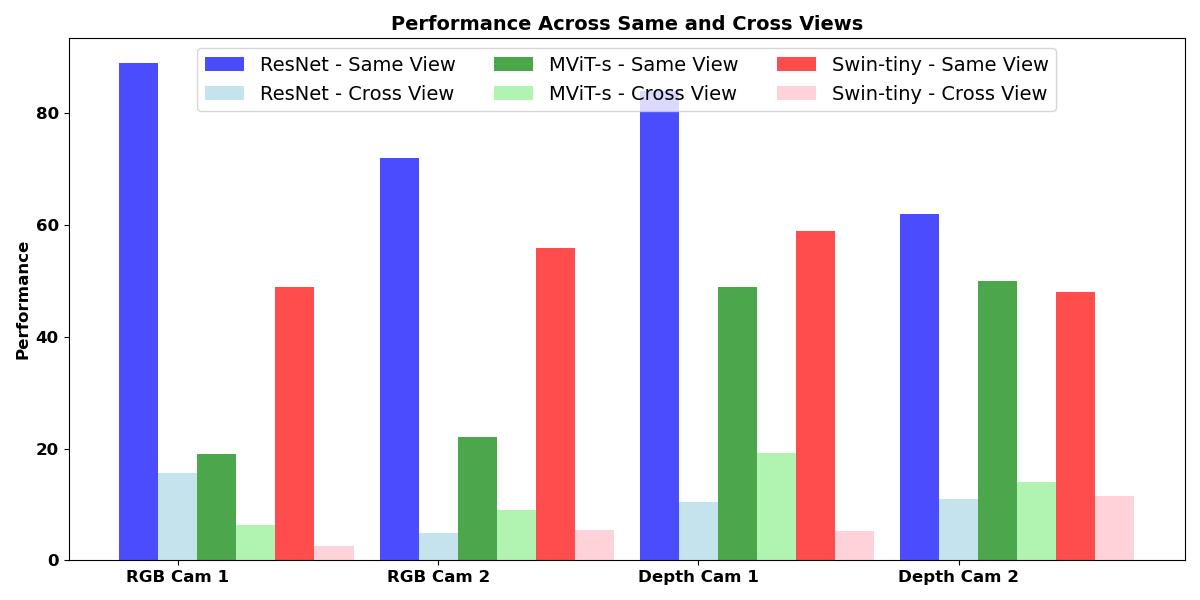}
    \caption{ \scriptsize Comparison of same-view and cross-view inference performance for ResNet, MViT-s, and Swin-t using visual data. The results show top-1 accuracy for each model when tested on the same camera view and cross views.}

        \label{fig:stacked_robustness}
\end{figure}

\texttt{DARai} provides real-world conditions to test model robustness. For example, Figure~\ref{fig:stacked_robustness} shows cross-view evaluations on three models, namely ResNet, MViT-s, and Swin-t, trained on a camera view and tested on same-view versus cross-view using RGB and depth data, respectively. As expected, these results show that models suffer a severe drop in accuracy when tested on the cross view. Objects and subjects that are fully within the viewpoint of a camera may not necessarily be viewable by the other camera. Thus, a significant drop in performance, e.g., ResNet performance drops from ~89\% to ~15\% and same model trained on the second view accuracy drops from ~72\% to 5\%, is expected. Interestingly, depth data shows a similar trend although depth provides view-invariant information such as distances. This can be explained by the occlusion across views. Overall, vision-based recognition is tightly coupled to the camera perspective, a serious drawback for daily life applications within a home environment.

\subsubsection{Wearable Cross-Body Evaluation}
\begin{figure}[h!]
    \centering
    \includegraphics[width=0.6\textwidth]{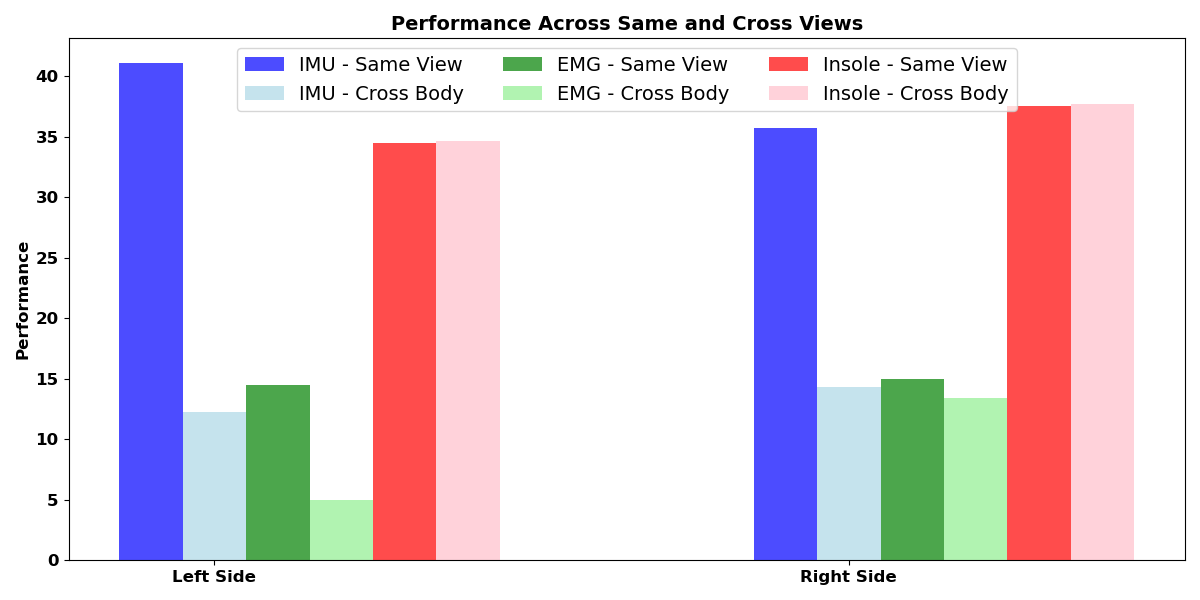}
    \caption{ \scriptsize Comparison of same-view and cross-view inference performance using IMU, forearm muscle EMG, and insole pressure wearable data. The results show top-1 accuracy for each sensor when tested on the same side of the body and the opposite side. Models were trained separately on left-side and right-side data for this experiment.}
    \label{fig:wearable_robustness}
\end{figure}

While cross-view generalization is a common concern for RGB and depth modalities, wearable sensors face both cross-subject and cross-body challenges~\citep{yarici2025subject}. In this analysis, data from a sensor worn on one side of the body, e.g., a hand, an arm, or a foot, is used for training while testing is performed on data from the same type of sensor on the other side of the body. This setup creates a “cross-body” domain shift. In such settings, the challenges are different from those in vision-based cross-view systems. Certain body activities are mostly symmetric across the left and right parts of the body. Those activities are generally related to movement patterns and balancing body weights and pose, represented by foot-pressure insoles as depicted in Figure~\ref{fig:wearable_robustness} for the Insole sensors. In contrast, hands interactions are generally asymmetric and one hand is usually dominant. As a result, a model trained on right-wrist or right-hand signals may struggle when tested on data from the left side, which can have a weaker or distinct signal, leading to higher error rates. This is clearly depicted in Figure~\ref{fig:wearable_robustness} for the IMU and EMG sensors.

In general, based on \texttt{DARai}, wearable cross-body models show less performance degradation when compared to cross-view camera models. This observation leads to the question about the contribution of every modality to recognition at every hierarchy. There seems to be a collaboration across these modalities to create a representation space that outperforms the counterparts created by a single or a subset of these modalities. \texttt{DARai} provides the proper settings to conduct such investigation. 
\section{Multimodal and Hierarchical Human Activities Benchmark}
\label{Section:Hierarchical}

\begin{figure}[ht]
    \begin{subfigure}[b]{0.45\textwidth}
        \centering
        \includegraphics[height=7cm]{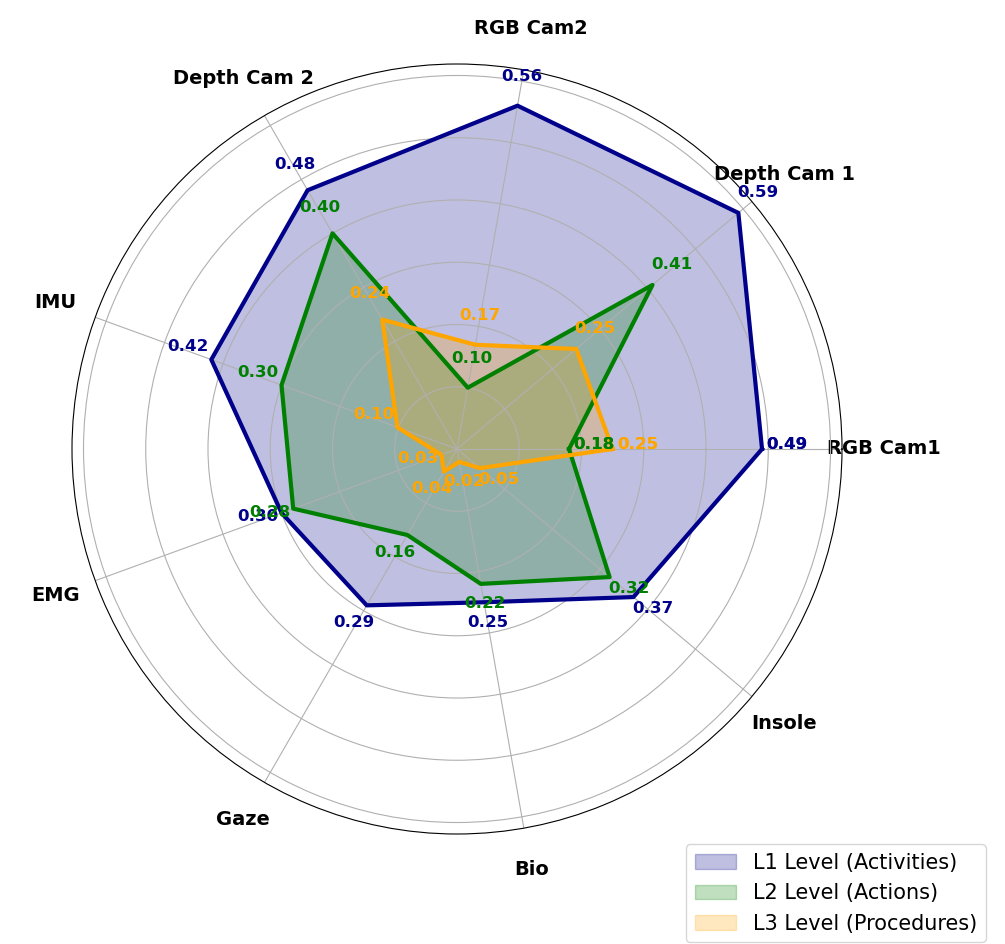}
        \caption{\scriptsize A transformer-based architecture was trained from scratch on each modality separately for this experiment, using Swin-t for visual data modalities and the modified transformer introduced in Appendix~\ref{Appendix: ablation tables} for other data modalities.}
        \label{fig:poly-unimodal}
    \end{subfigure}
    \quad
    \begin{subfigure}[b]{0.45\textwidth}
        \centering
        \includegraphics[height=7cm]{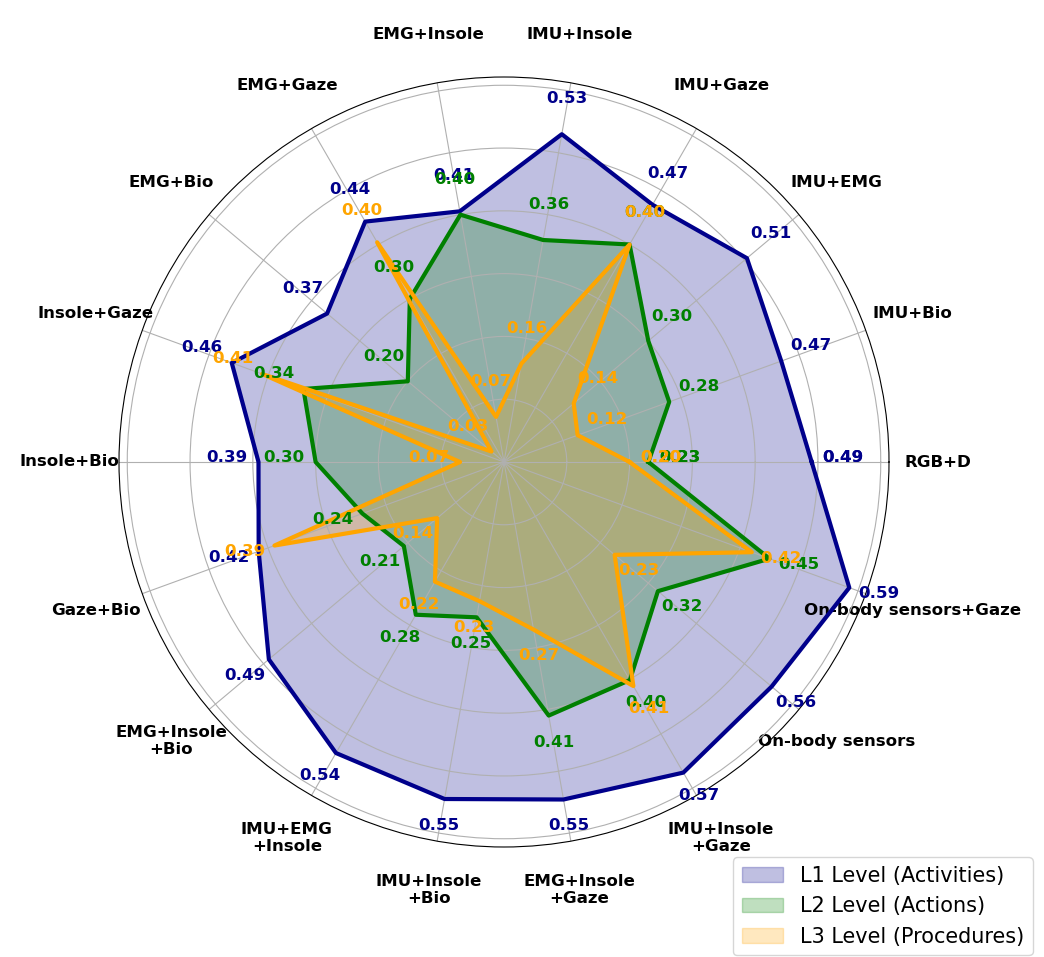}
        \caption{\scriptsize A transformer-based architecture was trained to fuse each modality pairs separately for this experiment, using Swin-t for visual data modalities and the modified transformer introduced in Appendix~\ref{Appendix: ablation tables} for other data modalities.}
        \label{fig:multimodal-granularity}
    \end{subfigure}
    \caption{ \scriptsize Comparison of Top-1 accuracy results 
        for individual data modalities and data modality pairs across 3 level of \texttt{DARai} 
        hierarchy.}
    \label{fig:polygon-plots}
\end{figure}

In this section, we evaluate unimodal and multimodal approaches on \texttt{DARai} across the three hierarchical levels (L1, L2, L3). We employ a transformer-based architecture, introduced in Appendix~\ref{Appendix: ablation tables}, to analyze each modality alone and in pairwise combinations. Figure~\ref{fig:polygon-plots} presents these experiments results, highlighting both individual performance differences and the advantages of fusing complementary data modalities.

\subsection{Unimodal Analysis and Comparison}
\label{sec:unimodal-strength}

\begin{figure}[H]  
    \centering
    \setlength{\tabcolsep}{3pt}
    \renewcommand{\arraystretch}{0.9}
    \resizebox{0.94\columnwidth}{!}{%
    \begin{tabular}{cc}
        \begin{subfigure}[b]{0.48\textwidth}\centering
            \includegraphics[width=\linewidth]{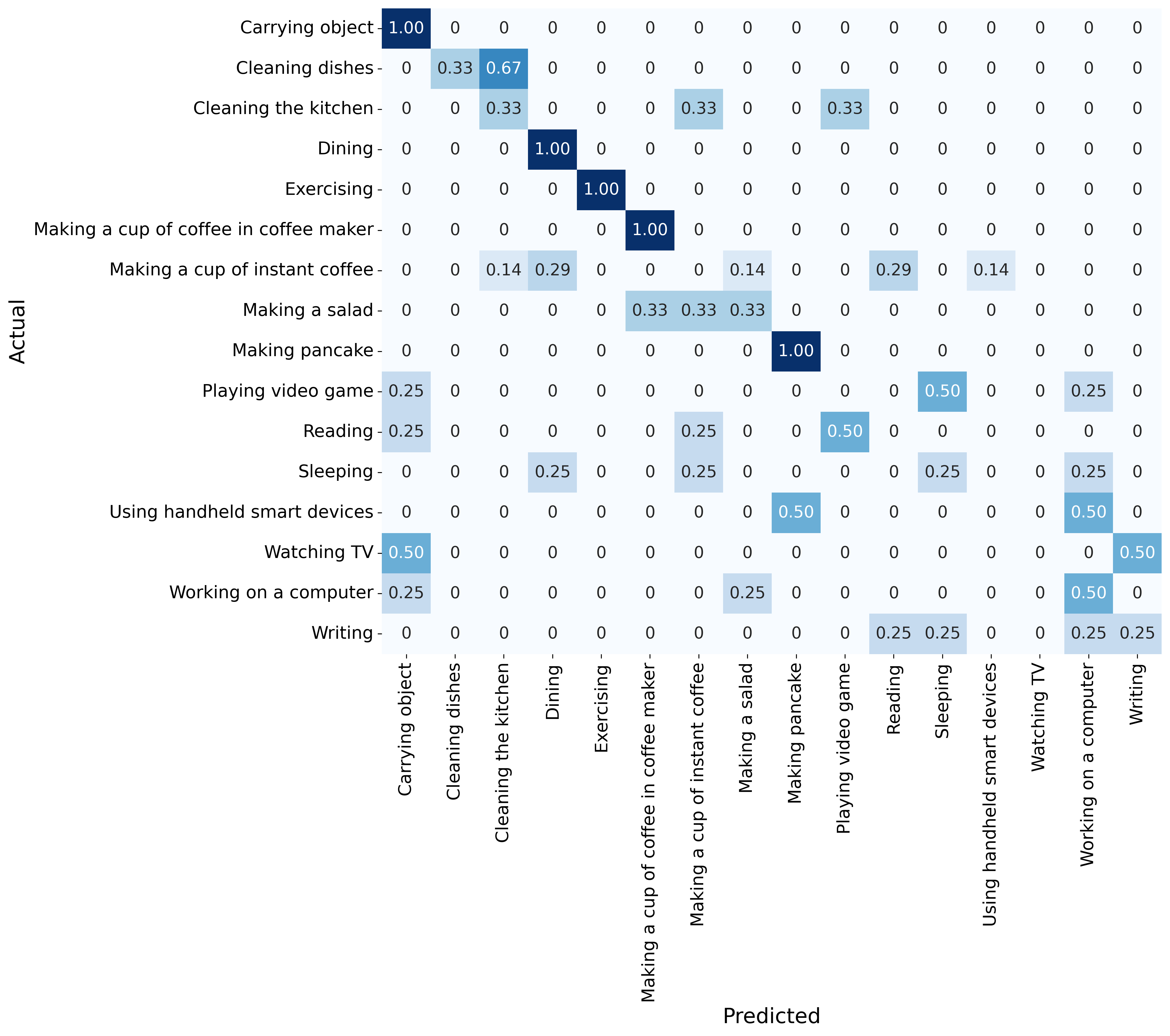}
            \caption{Insole}
        \end{subfigure} &
        \begin{subfigure}[b]{0.48\textwidth}\centering
            \includegraphics[width=\linewidth]{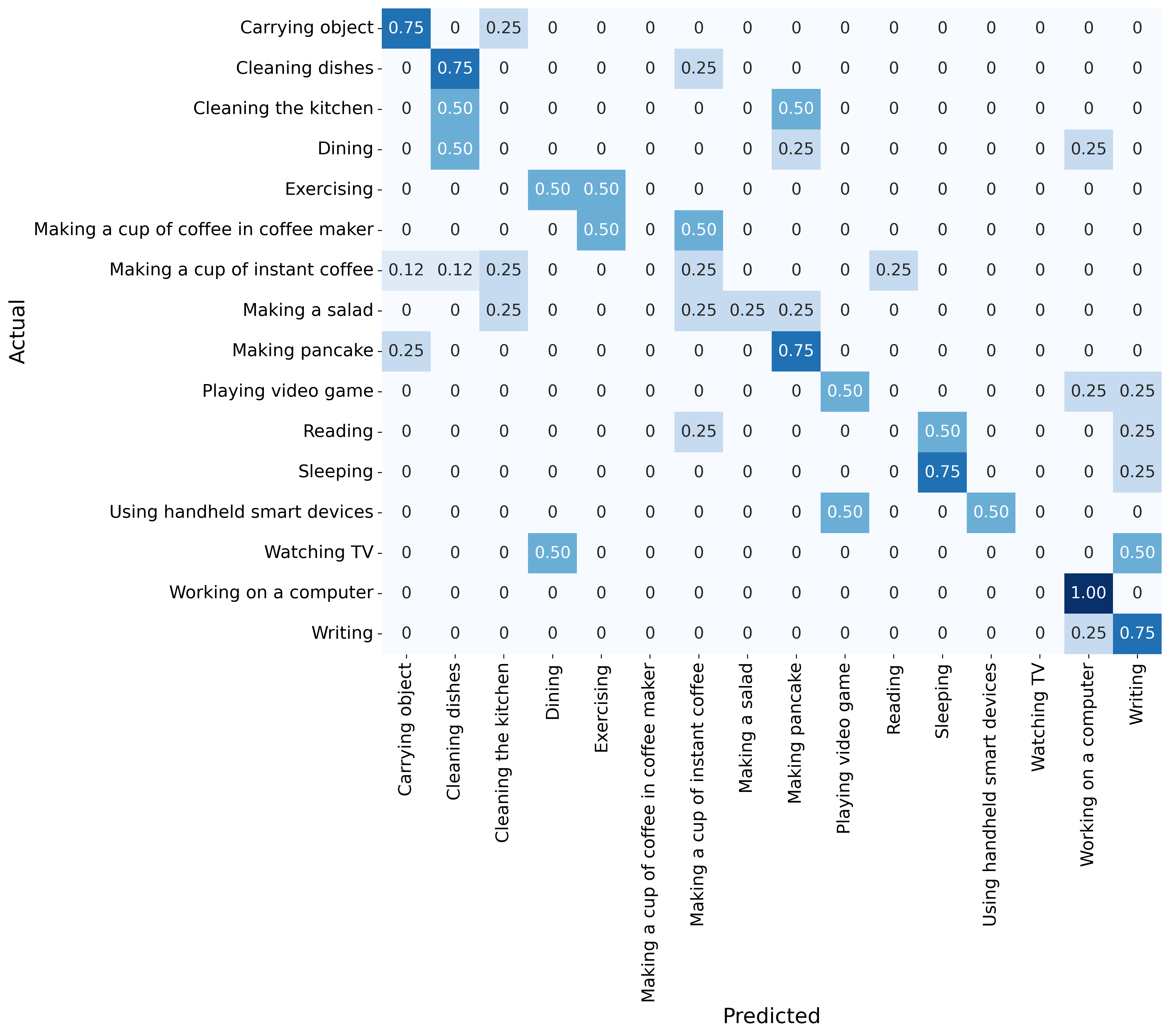}
            \caption{IMU}
        \end{subfigure} \\[2pt]
        \begin{subfigure}[b]{0.48\textwidth}\centering
            \includegraphics[width=\linewidth]{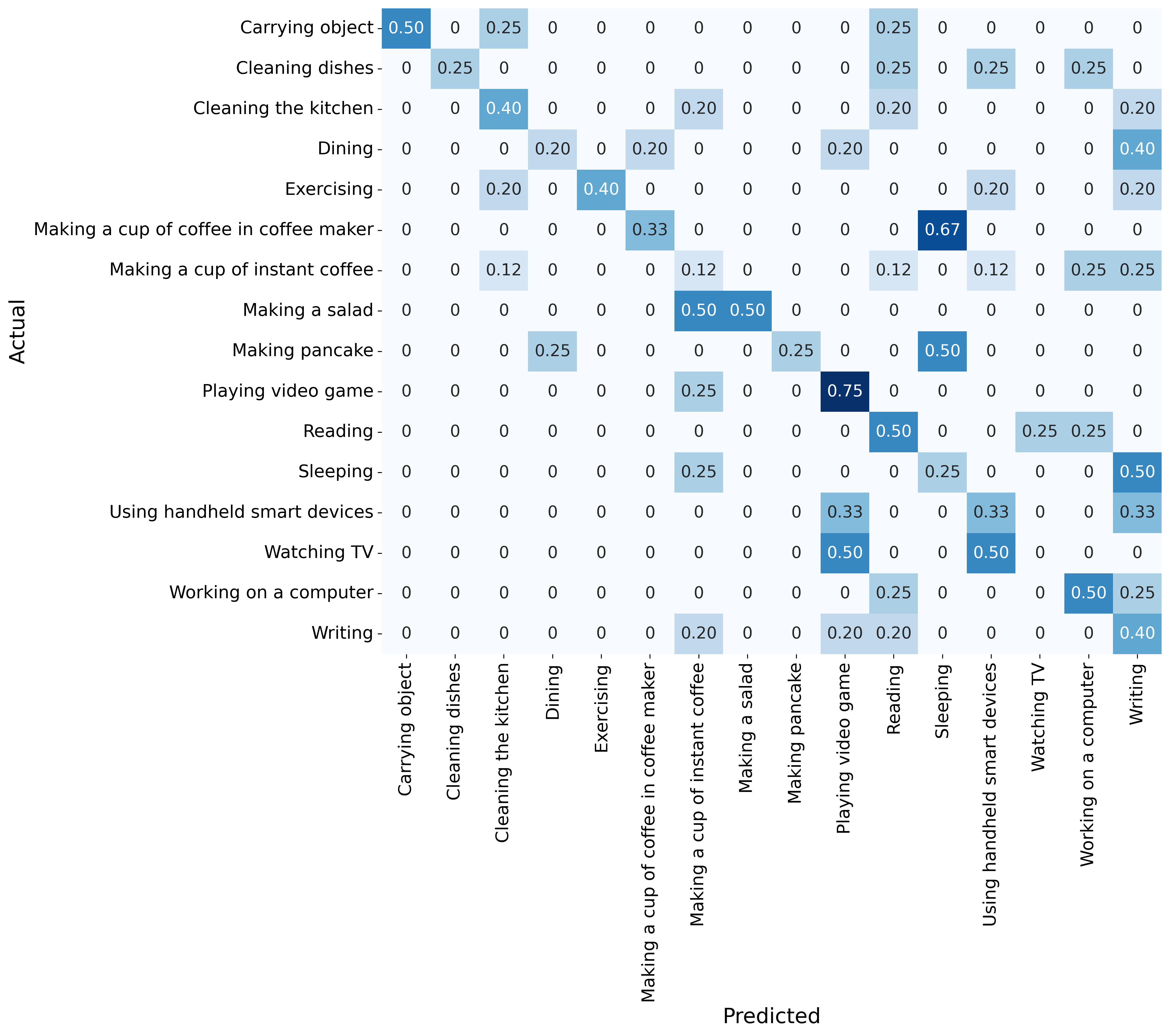}
            \caption{EMG}
        \end{subfigure} &
        \begin{subfigure}[b]{0.48\textwidth}\centering
            \includegraphics[width=\linewidth]{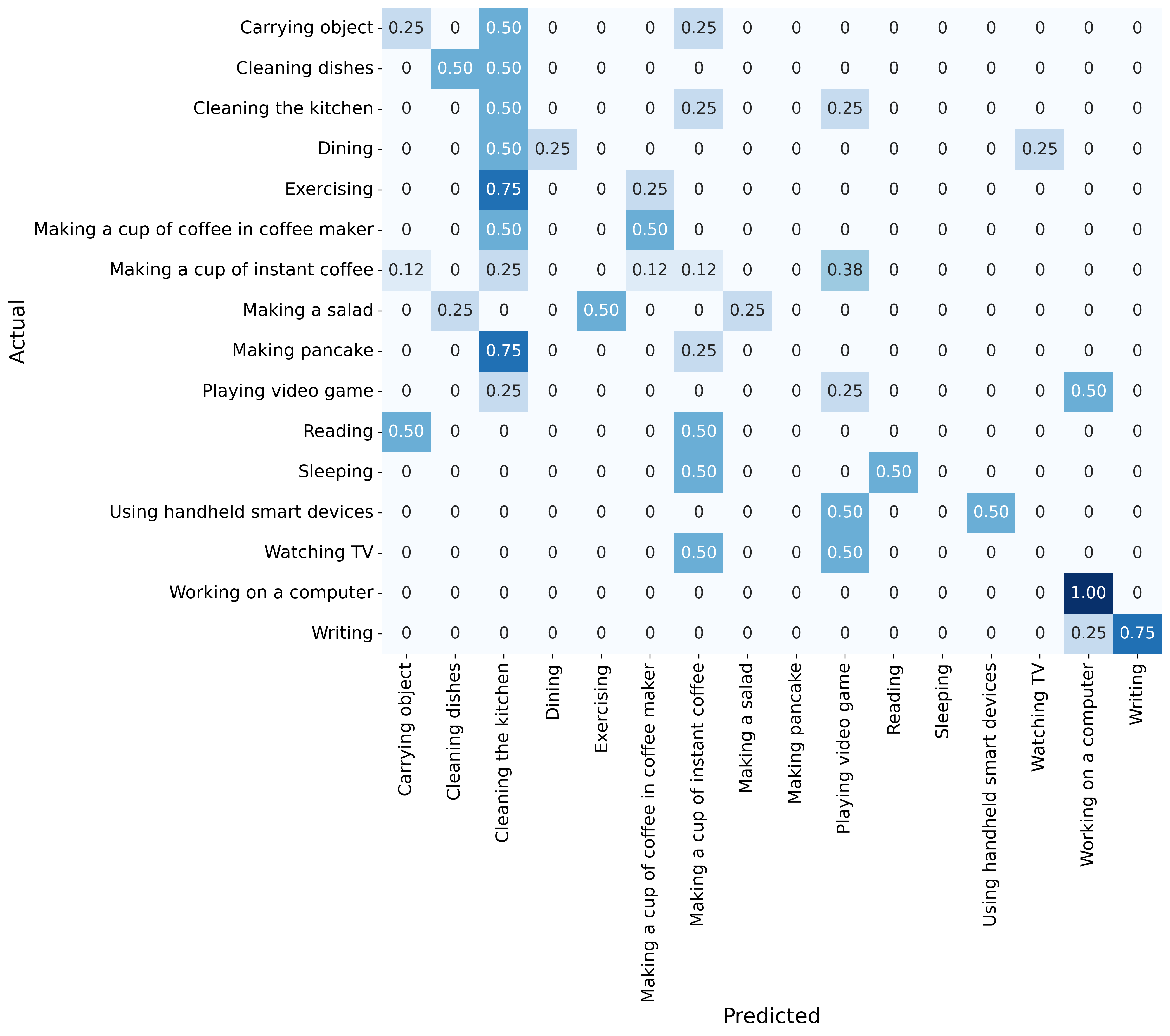}
            \caption{Gaze}
        \end{subfigure} \\[2pt]
        \begin{subfigure}[b]{0.48\textwidth}\centering
            \includegraphics[width=\linewidth]{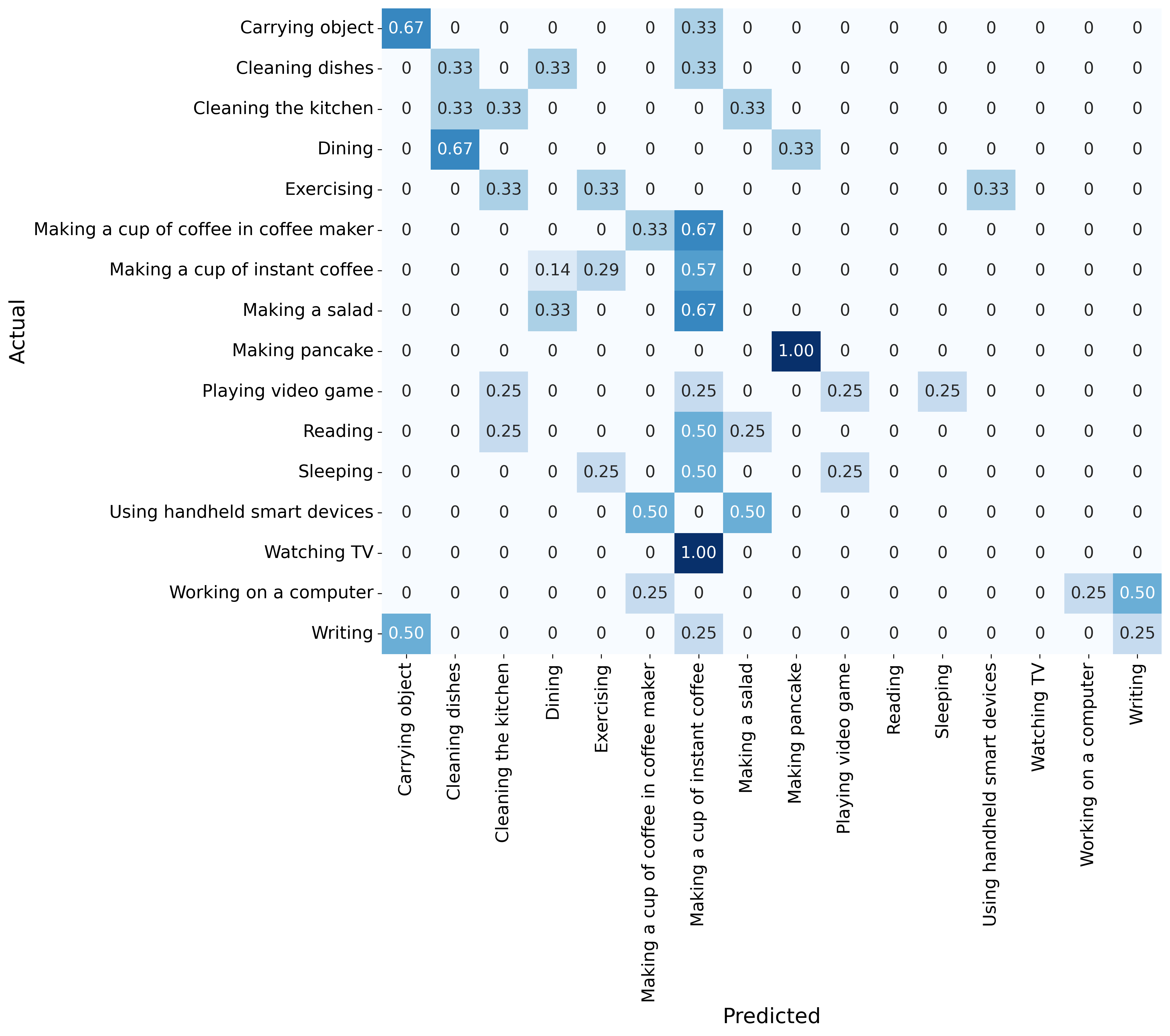}
            \caption{Bio}
        \end{subfigure} &
        \begin{subfigure}[b]{0.48\textwidth}\centering
            \includegraphics[width=\linewidth]{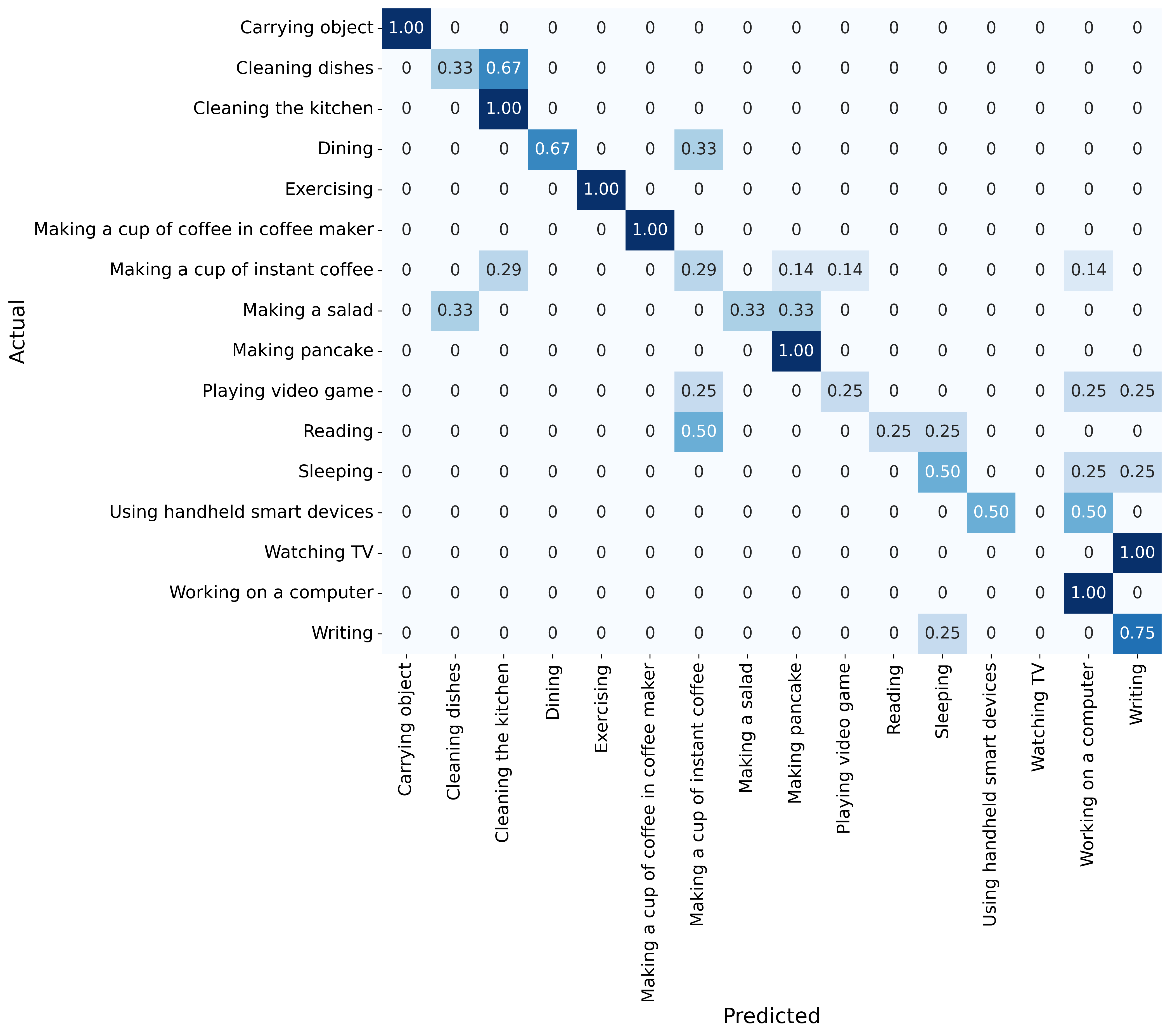}
            \caption{Fusion}
        \end{subfigure}
    \end{tabular}%
    }
    \vspace{-4pt}
    \caption{\scriptsize Activity confusion matrices for unimodal and multimodal settings.}
    \label{fig:unimodal-cm-main}
\end{figure}

Table~\ref{tab:modality_comparison} summarizes the effectiveness of different sensing modalities in recognizing human activities. Each modality excels in specific contexts but also has limitations. For example, gaze data is effective for activities involving sustained visual fixation, while EMG signals capture distinct muscle activation patterns. Biomonitoring data provide insights into physiological changes but lack fine-grained resolution for differentiating similar exertion levels. Given such complementary effectiveness, it is common to fuse these modalities to improves accuracy and robustness. We show the confusing matrices of uni-modal data and fusion of five uni-modal data on Figure \ref{fig:unimodal-cm-main}. The confusion matrices show that each sensor captures unique aspects of activity but struggles when used alone. For instance, Insole data effectively captures unique foot pressure patterns and performs well in recognizing activities such as \textsf{Carrying objects}, where additional weight affects pressure distribution. On the other hand, relying on Insole modality only introduces uncertainty in differentiating between activities such as \textsf{Cleaning dishes} and \textsf{Making a salad} that have similar standing posture. Another example is IMU data, which are  effective at separating activities with dynamic movements such as \textsf{Exercising} and \textsf{Working on a computer}, which is a primarily static activity. However, IMU data are less effective in separating activities with similar hand movements such as \textsf{Cleaning dishes} and \textsf{Cleaning the kitchen}. More detailed study on fusion based on \texttt{DARai} is in Appendix \ref{Appendix: ablation tables}.
\begin{table}[htb]
    \centering
    \scriptsize
    \resizebox{\textwidth}{!}{
    \begin{tabular}{l l l}
        \toprule
        \textbf{Modality} & \textbf{Best Suited For} & \textbf{Example Activities} \\
        \toprule
        Gaze & Sustained visual fixation & Reading, Watching TV, Working on a computer \\
        EMG & Muscle activation patterns & Making pancakes (whisking motion), Playing video games (controller grasping) \\
        Bio-Monitoring & Physiological variations & Playing video games (increased heart rate), Sleeping (reduced activity) \\
         Insole & Planter pressure patterns & Carrying objects (added weight changes foot pressure), Standing postures \\
        IMU & Dynamic body movements 
            & Exercising (dynamic motion), Working on a computer (static) \\
        \bottomrule
    \end{tabular}
    }
    \caption{ \scriptsize Comparison of Different Modalities for Activity Recognition}
    \label{tab:modality_comparison}
\end{table}

\texttt{DARai}'s unique hierarchical and multimodal aspects provide a space to investigate modality contribution to every level of the hierarchy. Figure~\ref{fig:poly-unimodal} shows the performance of each of the modalities. The best-performing unimodal models for non-visual data modalities are shown on this plot. The input length, window size, and other parameters are kept consistent across all levels of hierarchy, with samples drawn exclusively from one hierarchy level at a time (L1, L2, or L3). As shown in Figure \ref{fig:poly-unimodal}, changes in performance across granularity levels are \emph{not} linear. Furthermore, different modalities respond differently to decreasing abstraction, shorter sample sequences, and finer-grain details. Previous studies have shown that task granularity negatively impacts natural image models~\citep{prabhushankar2020contrastive,kokilepersaud2024taxes, kokilepersaud2025hex}. Our findings confirm that even the visual model with the highest accuracy (i.e., 56\%-59\%), experiences a significant performance drop when transitioning from L1 to L2 and L3. In contrast, insole pressure, forearm muscle EMG and biomonitoring modalities do not exhibit the same steep decline in performance, experiencing only 5\%, 2\%, and 3\% drops from their L1 accuracy, respectively. Notably, the decline in visual model performance is smaller between L2 and L3, with both camera views performing better at L3 than at L2. This trend is expected, as 20\% of the procedures (L3) are shared between higher-level activities (L1) and actions (L2), indicating more overlap at these levels.

In contrast, biomonitor signals and hand muscle EMG signals demonstrate greater resilience to increased granularity between L1 and L2. Additionally, insole pressure outperforms other non-visual modalities at the action level (L2) with 32\% accuracy. The alignment between plantar patterns and the structure of actions makes insole data effective for recognizing fine-grain actions at this level. However, from L2 to L3, none of these modalities maintain the same level of resilience. In naturalistic daily life activities, data are not trimmed around a single action, so wearable time-series often include extended inactive intervals, such as minimal hand movement or prolonged standing in one signal, while other modality signals remain more active. These intervals can overshadow distinct activity patterns within a single modality and lead to performance decline. At lower levels of the hierarchy, the increased granularity magnifies this effect, making it harder to capture pattern differences.

\subsection{Multimodal Analysis and Comparison}
\label{sec:hierachical-multimodal} 

We evaluate multimodal modeling across the three hierarchical levels of \texttt{DARai}, considering both pairwise combinations (Figure~\ref{fig:multimodal-granularity}) and structured sensor groups (Table~\ref{tab:sensor_groups}). All modality combinations exhibit decreasing accuracy from L1 to L3, but the extent of the decline varies by modality type. Pairwise results (Figure~\ref{fig:multimodal-granularity}) show that some modalities complement each other effectively, such as IMU with Insole or Gaze, while others provide limited benefit when fused with others, such as EMG with biomonitoring. These examples illustrate how the information carried by different signals overlaps or reinforces each other, but the overall trends are clearer when modalities are considered in broader groups.

Table~\ref{tab:sensor_groups} summarizes performance when sensors are ablated into functional groups. Physiological signals (EMG and biomonitoring) are informative at coarse granularity (L1) but their performance declines sharply at finer levels, indicating limited discriminative power for detailed activities. Biomechanical signals (IMU and Insole) achieve stronger results and degrade more gradually, reflecting their utility in capturing motion dynamics. Gaze contributes to attentional cues but performs modestly when used alone. Third-person vision (RGB and Depth) provides competitive performance at L1 but deteriorates substantially at finer levels, highlighting the difficulty of translating scene context into fine-grain distinctions. The best overall performance is achieved when combining all wearable sensors (physiological, biomechanical and behavioral), which sustains relatively high accuracy across all levels, reaching over 42\% at L3. 

These findings suggest that while individual modalities have characteristic strengths and weaknesses, grouping them by functional role highlights which categories are robust to task granularity. Physiological signals alone are fragile, external vision is coarse-grain biased, and biomechanical dynamics are more stable, but fusing multiple wearable signals consistently yields the most reliable performance across the hierarchy.
\begin{table}[htb]
\centering
\resizebox{\textwidth}{!}{%
\begin{tabular}{cllrrr}
\hline
Placement                & Sensor Group           & Modalities                      & L1 Accuracy & L2 Accuracy & L3 Accuracy \\ \hline
\multirow{4}{*}{On-body} & Physiological          & EMG + Bio Monitor               & 0.368       & 0.200       & 0.026       \\      
                         & Biomechanical          & IMU + Insole Pressure           & 0.530       & 0.359       & 0.159       \\
                         & Behavioral (attention) & Gaze                            & 0.298       & 0.159       & 0.052       \\
                         & All                    & EMG + Bio + IMU + Insole + Gaze & 0.585       & 0.451       & 0.423       \\ \hline
Off-body                 & Third-Person Visual    & RGB + Depth                     & 0.49        & 0.23        & 0.20         \\ \hline
\end{tabular}%
}
    \caption{ \scriptsize Comparison of Different Sensor Modality Groups. Missing entries correspond to cases where no single-modality results were available.}
    \label{tab:sensor_groups}
\end{table}

\section{Temporal Dependencies Benchmark}
Understanding temporal dependencies is crucial for modeling human activities, which unfold hierarchically over time. Unlike existing datasets with predefined step boundaries, DARai presents a continuous, unstructured setting where activities transition fluidly across different granularities, posing challenges in recognizing hierarchical structure and temporal dependencies. To assess these dependencies, we focus on temporal activity localization and activity anticipation. Performance is evaluated using Mean over Classes (MoC), which measures per-class performance and averages across all classes, following the long-term action anticipation protocol~\citep{gong2022future,abu2021long,sener2020temporal,ke2019time, kim2026countering}. For consistency, results are averaged over three different seeds. For the activity localization result, refer to~\ref{sec:app-localization}.
\label{Section:Temporal}

\begin{figure}[htb]
    \centering
    \includegraphics[width=0.9\textwidth]{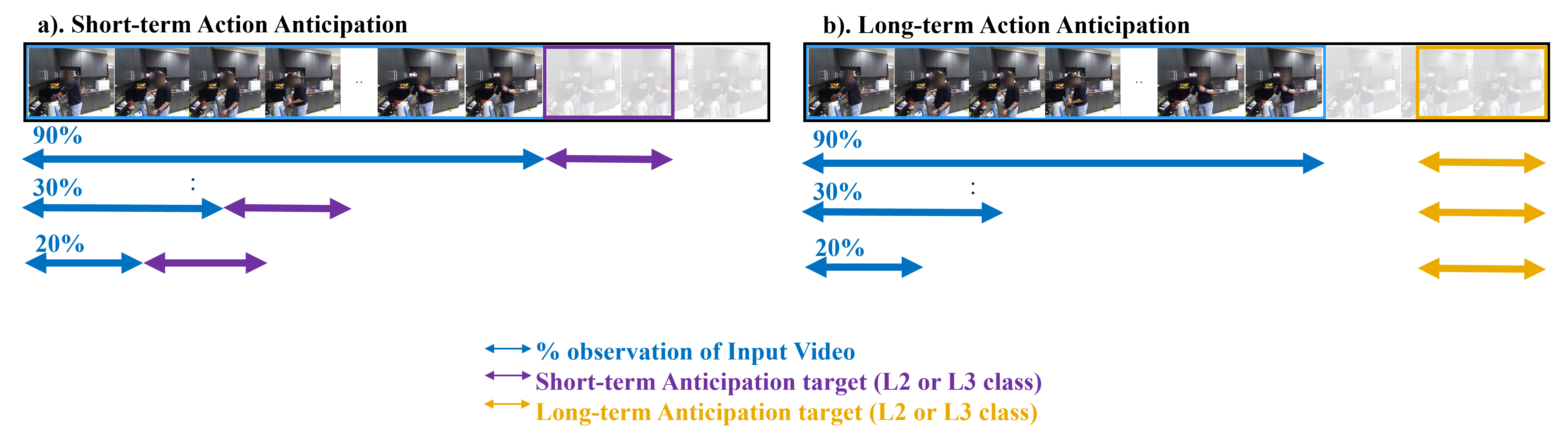}
    \caption{ \scriptsize Feeding different length of sample sequence for action anticipation task. (a) shows short-term action anticipation task, and (b) shows long-term action anticipation task.}
    \label{fig:horizon-anticipation}
\end{figure}

\subsection{Short- and Long-Horizon Activity Anticipation}
\label{sec:anticipation}
Action anticipation predicts future actions from partially observed sequences. \texttt{DARai} presents extended, fine-grained actions and procedures from a third-person view, enabling two tasks: \textbf{short-horizon}, which focuses on immediate upcoming actions, and \textbf{long-horizon}, which targets more distant actions. Figure~\ref{fig:horizon-anticipation} illustrates the setup, and we vary the observation ratio \(\alpha\) from 0.1 to 0.9.

\begin{figure}
    \centering
    \includegraphics[width=0.8\textwidth]{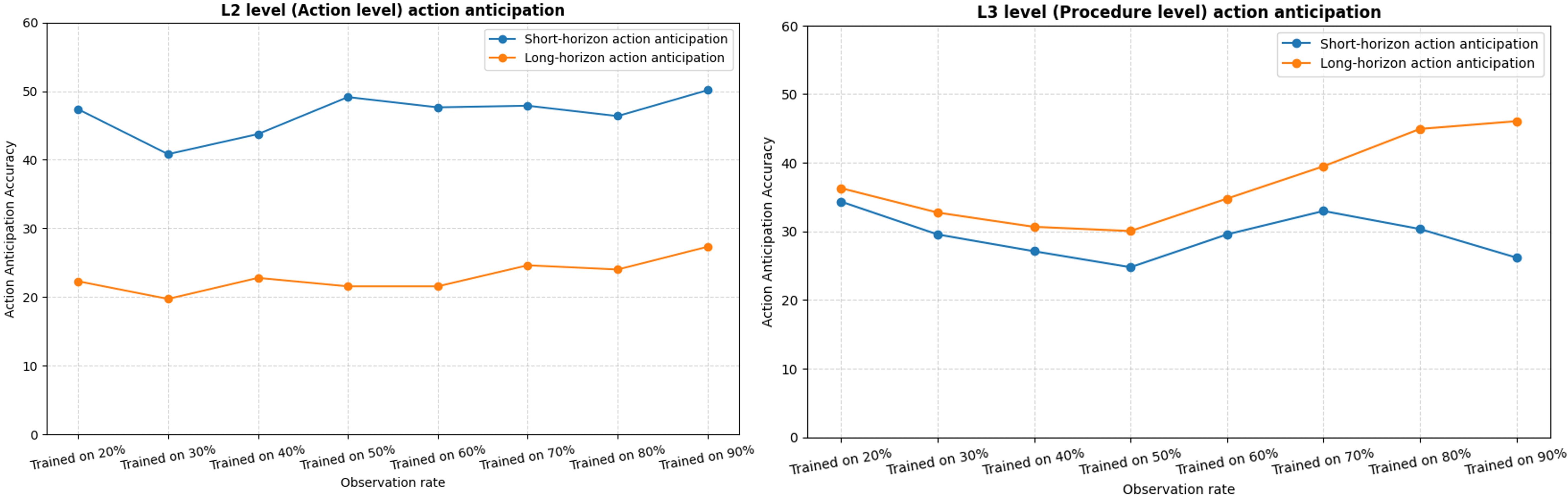}
    \caption{ \scriptsize Action anticipation accuracy across different observation rates based on partially observed L1-level sequences at L2 (Action level) and L3 (Procedure level). The left graph represents L2-level action anticipation for both anticipating over the next 8 seconds (short-term action anticipation) and the latest 8 seconds of each sequence (long-term action anticipation). The right graph represents L3-level action anticipation for both short-term and long-term action anticipation. The accuracy represents Mean over Classes (MoC) accuracy.}
    \label{fig:short-long-ant-var-obs}
\end{figure}

\begin{figure}[ht!]
    \centering
    \includegraphics[width=0.6\textwidth]{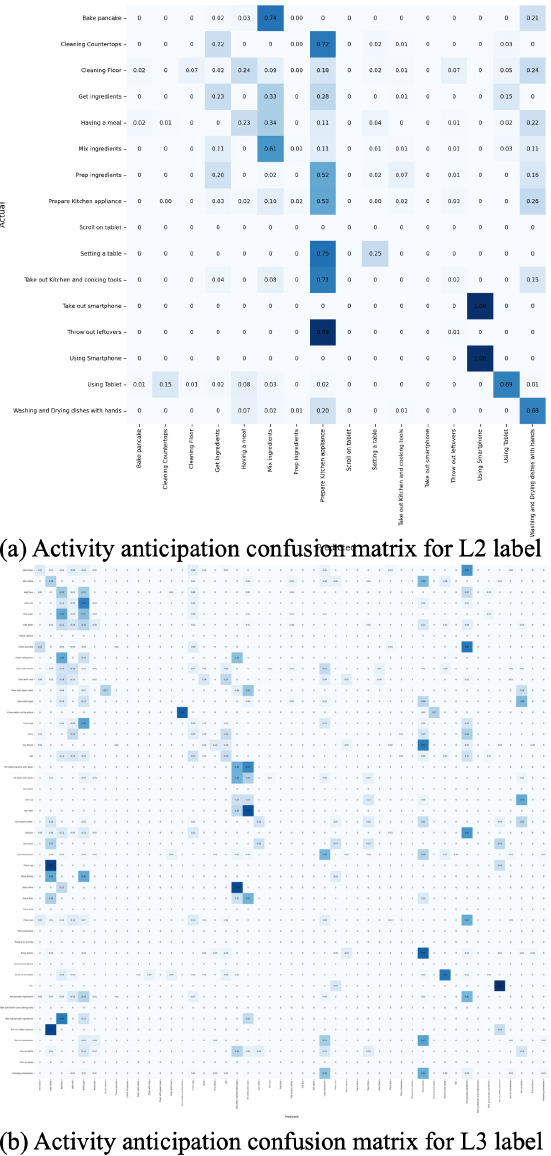}
    \caption{ \scriptsize This figure shows the comparison of confusion matrices demonstrating action anticipation performance when predicting next L2 labels (a) and L3 labels (b).}
    \label{fig:action-ant-confusion-matrix}
\end{figure}

Figure~\ref{fig:short-long-ant-var-obs} shows that short-horizon anticipation improves steadily with more observation—up to about 50\%—but gains little beyond that. This indicates that a moderate window captures enough context, and extra observations can introduce redundancy. In contrast, long-horizon anticipation consistently benefits from increasing the observation ratio, reflecting the strong temporal dependencies of extended procedural activities in \texttt{DARai}. 
\subsection{Hierarchical Activity Anticipation}
\noindent Finer-grain anticipation tasks add additional complexity, as the model must understand both the hierarchical structure and the sequence of events or actions from higher-level activities~\citep{kim2025multi}. This type of anticipation is far more challenging than traditional feature prediction based on observed features. To address such challenges, \texttt{DARai} is essential for testing and benchmarking models, since it provides multiple levels of hierarchy.

Figure \ref{fig:short-long-ant-var-obs} shows the disparity in anticipation performance across different levels of temporal granularity. At the action level (L2), short-term action anticipation accuracy fluctuates between 42\% and 50\%, while long-term anticipation remains lower, ranging from 20\% to 25\% across different observation rates. This suggests that action anticipation primarily relies on local temporal cues rather than extensive historical context. Conversely, at the procedure level (L3), long-term anticipation accuracy at L3 improves from 36\% to 46\%, while short-term varies more significantly, ranging from 26\% to 34\% as observation rate changes. In addition, the confusion matrices of Figure~\ref{fig:action-ant-confusion-matrix} shows that predicting the next L3 labels (b) is harder than predicting the next L2 labels (a). Specifically, from (a) we infer that the model confuses activities within the same category, such as "Take out smartphone" vs. "Using smartphone" in device-related tasks, and "Prepare ingredients" vs. "Get ingredients" in kitchen-related tasks. These findings emphasize the importance of considering the hierarchical nature of activities when designing anticipation models, as the required observation window and predictive capability vary depending on the level of granularity. Thus, optimizing anticipation strategies should be guided by the underlying temporal structure of the target activities.

\begin{figure}[htb]
    \centering
    \includegraphics[width=1.0\textwidth]{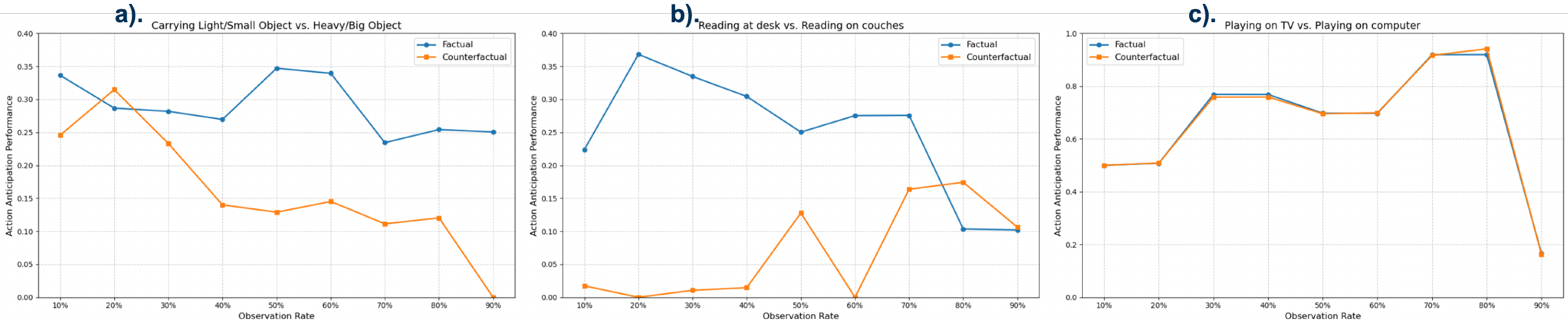}
    \caption{ \scriptsize Action counterfactual cases demonstrating performance comparisons under different observation rates. (a) Comparison between `Carrying light/small object' (factual) and `Carrying heavy/big object' (counterfactual). (b) Comparison between `Reading at desk' (factual) and `Reading on couches' (counterfactual). (c) Comparison between `Playing on TV' (factual) and `Playing on computer' (counterfactual). The graphs show action anticipation performance across varying observation rates from 10\% to 90\%.}
    \label{fig:action-counterfactual}
\end{figure}

\subsection{Activity Anticipation in Action Counterfactual cases}
\noindent We further conduct experiment on action counterfactual cases. Especially, we select pairs of samples that share identical L1 and L3 labels while differing only in their L2 labels. Specifically, both cases in Figure~\ref{fig:action-counterfactual} (a) share the same L1 label "Carrying Object" and L3 labels ("Pick up box", "Walk with box", "Put down box") but differ in L2 labels ("Carrying light/small object" vs. "Carrying heavy/big object"). In (b), both share same L1 label "Reading" and L3 labels ("Get paper/book", "Read paper/book") with L2 differences ("Reading at desk" vs. "Reading on couches"). In (c), both share same L1 label "Playing Video Game" and L3 labels ("Playing reaction game", "Playing speed game") with L2 variations ("Playing on TV" vs. "Playing on computer").

To examine the model's generalization ability to unseen counterfactual scenarios, during training, only the factual cases were included in the dataset, while the counterfactual cases were exclusively used during inference for comparison. The objective of the experiment is to predict the next L3 labels.

The results reveal distinct performance patterns: In cases (a) and (b), significant performance gaps were observed between factual and counterfactual samples, whereas in case (c), the model performed similarly in both scenarios. Notably, in case (b), the model frequently misclassified counterfactual samples as "Conversation on the phone" or "Rinse dishes". We attribute this error to the model's inherent bias towards contextual background patterns, as "Conversation on the phone" in the dataset often occurs in couch settings. This indicates that the current state-of-the-art models, when applied to the \texttt{DARai} dataset, tend to rely heavily on contextual biases rather than focusing on the intrinsic characteristics of the target actions themselves.
\section{Discussion}

\paragraph{Activity structure}
Continuous, unscripted recordings include long inactive intervals and fluid boundaries between actions and procedures. This complicates segmentation and recognition but reflects natural behavior.

\paragraph{Action counterfactuals}
Providing activity categories without scripts leads participants to produce alternative executions of the same activity. This changes temporal relations between actions and procedures and supports analysis of execution variability.

\paragraph{Granularity}
Performance shifts across L1, L2, and L3 are not uniform. Coarse activities are easier to classify than fine-grained actions and procedures. Insole pressure and EMG are less sensitive to granularity than RGB.

\paragraph{Robustness}
Models trained on a specific camera view or body side lose accuracy when tested on a different view or side. This shows the need for methods that handle placement and viewpoint changes. Sensor fusion improves accuracy when single modalities are weak.

\paragraph{Temporal modeling}
Actions show higher temporal diversity than procedures. Longer observation windows help localize actions. Predicting long-horizon procedures is often easier than predicting long-horizon actions.

\paragraph{Limitations}
The large sensor setup reduces flexibility for data collection and expansion. As a result we based our study on a cohort of 50 participants. We acknowledge that while this scale is on par with leading scripted and hybrid datasets, it does not match the participant breadth of recent large-scale, 'in-the-wild' collections. Future work can broaden environment coverage and explore fully wireless sensor setups. While the custom portable system cost was approximately \$22,000, more economical alternatives may be employed, as the setup and results are not contingent upon specific sensor brands.

\paragraph{Broader Impact}
By showing that non-visual modality fusion improves recognition, \texttt{DARai} supports privacy-focused approaches with minimal or no camera use. The dataset encourages development of models that are robust to placement and viewpoint changes. The data release follows informed consent of participants and promotes open research. At the same time, human activity datasets inherently raise privacy considerations. To address this, we exclude personal identifiers and blur faces where possible. These safeguards contribute to, but do not resolve, the broader challenge faced by the research community in balancing data utility with privacy.

\section{Conclusions}
We introduced and open-sourced \texttt{DARai}, a hierarchical, multimodal dataset with continuous, unscripted recordings and with 3 level (L1,L2,L3) temporal annotations. We reported benchmarks on robustness (cross-view and cross-body), multimodal fusion, and temporal modeling (localization and anticipation), and outlined additional tasks enabled by the data and dataset loader and configurations in the project repository. 

\section{Acknowledgments and Disclosure of Funding}

This work is partially funded through the support from the Franklin Foundation via the John and Marilu McCarty Chair professorship and a gift from Lab126 at Amazon.




\vskip 0.2in
\bibliography{sample}
\newpage
\appendix
\section{Dataset and Benchmarks access}
\label{Appendix: Links}

\subsection{Links to Access Dataset}
\label{app: links}
\begin{itemize}
    \item We provide open access to the DARai dataset at \url{https://ieee-dataport.org/open-access/darai-daily-activity-recordings-ai-and-ml-applications}

\item We provide access to the current hierarchical labels at
\url{https://ieee-dataport.org/open-access/darai-daily-activity-recordings-ai-and-ml-applications}

\item Code for data loaders, splits and experiments are provided at
\url{https://github.com/olivesgatech/DARai}
\end{itemize}
\subsection{Licenses and DOI}
The DOI of the dataset is \url{10.21227/ecnr-hy49}~\citep{ecnr-hy49-24}\\
The code is associated with an MIT License. \\
The license associated with our dataset is a Creative Commons International 4 license.
\label{app: license}

\subsection{Maintenance Plan}
\label{app: maintenance}
The dataset is provided on IEEE Dataport as an open access dataset and IEEE subscription is not required to access or download the dataset. Users are required to login to download. Labels are accessible alongwith the data on IEEE Dataport.

The code is hosted on the GitHub repository specified in Section A.1. Instructions and details regarding the dataset is located at this same repository. 


\section{DARai Key Observations and Challenges}
\label{Appendix: observation-challenges}
Collecting naturalistic human daily-life data has several challenges. We highlight the most significant ones below:
\begin{itemize}
    \item Sensor Selection: With many sensors available, choosing the most suitable options involves considering cost, data-stream stability, open-source SDK/API support, and metadata access.
  \item Different Sensor Platforms: Various sensors use different sampling rates, timestamp formats, and software frameworks for data streaming, making it difficult to synchronize and unify their outputs.
    \item High Equipment Costs: Advanced sensors, such as motion suits, eye trackers, and wireless EMG sensors, can be very expensive, which limits the deployment of diverse sensor setups.
    \item Thermal Issues: Depth sensors often overheat when used for a long time, reducing data quality.
    \item Connectivity Problems: Wearable sensors can lose connection in busy wireless networks, leading to missing data.
    \item Large Data Volume: Each session can generate up to 300\,GB, totaling more than 20\,TB overall, which requires extensive storage.
    \item Wearable Adjustments: Participants may move or adjust wearable devices for comfort, potentially interfering with sensor readings and occasionally requiring recalibration.
    \item Room Layout Constraints: Some environments require adjustments to the sensor setup (e.g., using fewer cameras), which can restrict how sensors are placed.\\
\end{itemize}
Our findings from data collection, annotation, and experiments on the \texttt{DARai} dataset are summarized below:
{\scriptsize
\begin{longtable}{l p{5in}}
\toprule
\textbf{Topic} & \textbf{Key Findings} \\
\midrule
\multirow{3}{1in}{Sensors Setup }
& Sensor sampling rates vary significantly (from 15 fps to 1000 Hz), leading to substantial differences in data density. High-frequency sensors (e.g., EMG , Biomonitors) offer dense temporal resolution, beneficial for capturing fine-grain interactions and changes, whereas lower-frequency sensors (e.g., cameras at 15 fps) yield sparse data suited for broader spatial and temporal relations within activities. \\
\midrule
 \multirow{3}{1in}{Hierarchical Annotations}
 & Natural human activities often lack certain boundaries between actions. Unlike structured datasets with predefined segments, \texttt{DARai} captures continuous activities with fluid transitions. It allows studying model performance on realistic temporal uncertainty and activity transitions. \\
 \cmidrule(lr){1-2}
 \multirow{3}{1in}{Subjects }
 &  Participants experienced more fatigue than dehydration after performing daily activity in our data collection. \\
 & Participants rated the eye tracker as most obtrusive among sensors used, while insoles were rated the least noticeable, suggesting insoles may be the optimal sensor choice for minimizing wearer discomfort. \\
 \cmidrule(lr){1-2}
\multirow{3}{1in}{Visual Model Training Strategies}
 & Finetuned \emph{visual} models \emph{consistently outperform} those trained from scratch, across all class categories and all three levels of hierarchy. \\
 & Within kitchen and living room subsets, \emph{depth} models trained from scratch \emph{outperform} their RGB counterparts. \\
 & At levels L2 and L3 of the hierarchy, \emph{depth} models continue to outperform \emph{RGB} models trained from scratch, likely due to the stability of depth-based spatial representations across hierarchy levels. \\
\cmidrule(lr){1-2}
\multirow{1}{1in}{Environment Differences}
 & Activity recognition accuracy within the living room is \emph{consistently higher} than those within the kitchen. \\
\midrule

\multirow{3}{1in}{Viewpoint and Placement Effect}
 & Visual modalities suffer a \emph{severe drop} in accuracy when tested on the cross view. \\
 & Non-visual modalities are \emph{robust} when the sensors capture symmetric interaction such as insole. \\
 & Wearable cross-body models show \emph{less performance degradation} when compared to cross-view camera models. \\
\midrule

\multirow{6}{1in}{Unimodal Sensor Strengths}
 & Insole data effectively captures unique foot pressure patterns and performs well in recognizing activities such as \textsf{Carrying objects}. \\
 & IMU data are effective at separating activities with \emph{dynamic movements} such as \textsf{Exercising} from static activities such as \textsf{Working on a computer}. However, IMU data are less effective in separating activities with similar hand movements such as \textsf{Cleaning dishes} and \textsf{Cleaning the kitchen}. \\
 & Gaze data are effective at separating activities with \emph{explicit visual fixations} such as \textsf{Reading}, \textsf{Watching TV}, or \textsf{Working on a computer}. \\
 & Gaze data are effective at \emph{separating fine-grained activities} when other data modalities that rely on physical movements suffer from performance degradation. \\
 & EMG signals are proficient at identifying \emph{distinctive patterns of muscle interactions} and can consequently recognize activities such as \textsf{Making pancakes} (beating with a whisking motion) or \textsf{Playing video games} (repetitive controller grasping). Yet they have difficulty with activities such as \textsf{Making coffee}, where limited arm or hand interactions are used. \\
 & Bio-monitoring data are good at \emph{identifying changes in physiology}, such as increased heart rate in \textsf{Playing video games} or reduced activity in \textsf{Sleeping}. However, they lack the resolution to separate activities with similar effort, such as different cooking activities. \\
\midrule

\multirow{5}{1in}{Hierarchical Granularity}
 & Changes in performance across granularity levels are \emph{not} linear. \\
 & The visual model with the highest accuracy, \(\sim 56\%, \sim 59\%\), experiences a \emph{significant performance drop} when transitioning from L1 to L2 and L3. \\
 & Insole pressure and Hand EMG data modalities do \emph{not} exhibit a significant decline in performance from L1 to L2 as visual data does. \\
 & All modality combinations experience a decline in accuracy from L1 to L3, but the magnitude of degradation \emph{varies significantly} by sensor modality combination. \\
 & Gaze data improves L3 (procedure) activity recognition fused with other modalities. \\
 & Biomonitoring signals do not offer fine-grained insights at Level 3, even when combined with insole, EMG, and IMU data modalities. \\
\midrule

\multirow{2}{1in}{Temporal Localization}
 & In temporal activity localization, both the camera view and the length of the input video play a crucial role in accurately segmenting an untrimmed, abstract-level activity sample into lower-level segments. \\
 & Actions and procedures are not isolated events. They are interconnected components of larger sequences, with a strong dependence on the temporal context of those that come before and after. \\
\midrule

\multirow{5}{1in}{Short- and \ Long-Horizon Anticipation}
 & At L2 (Action level), a moderate observation period provides sufficient temporal context for immediate action anticipation, while excessive observation may introduce redundant or less relevant information. \\
 & Long-horizon anticipation at the L2 level, however, remains consistently lower and exhibits minimal improvement as the observation rate increases. This indicates that action-level predictions do not benefit significantly from extended temporal context, possibly due to the more independent nature of individual actions. \\
 & At the L3 (Procedure level), short-horizon anticipation improves as the observation rate increases, reaching optimal performance around a 70\% observation rate. \\
 & Long-horizon anticipation at this level consistently improves with increasing observation rates, eventually surpassing short-horizon performance beyond a 70\% observation rate. This trend suggests that fine-grained procedural activities exhibit stronger temporal dependencies, benefiting from longer observation windows that provide richer contextual information. \\
 & While higher observation rates do not necessarily enhance short-horizon anticipation, they are crucial for long-horizon anticipation, particularly when dealing with fine-grained, procedural activities. \\
\bottomrule
\caption{Key observations}
\label{tab:key_observations}
\end{longtable}
}
\section{Overview of Dataset Contents}
\label{Appendix: Structure}
\subsection{Dataset Structure}
Our dataset is organized based on data modalities, with each modality serving as a top-level folder. Under each modality, we have Activity labels (level 1 of the hierarchy). Data samples for each activity are stored within these folders. Considering the number and variety of modalities in our dataset, we have grouped several data modalities under a supergroup name, such as Biomonitors. All related data modalities are included under these supergroup folders. Generally, the data structure is as described below. Interested parties can download a subset of the data or the entire dataset as a single unit.

\begin{verbatim}
LocalDatasetPath/
+-- Modality/
|   +-- ActivityLabel/
|   |   +-- View/
|   |       +-- data_samples/
\end{verbatim}

\subsection{Dataset Size}
Our original recordings, especially from multiple high-resolution cameras, were extensive in size. We began by separating the recordings into modalities (RGB, Depth, IR, Depth confidence) and segmenting the sessions by activity labels (top level of hierarchy). We then extracted frames from clips using a lossless compression of PNG level 6 for the de-identification and annotation processes. These steps reduced the size of the original recordings down to 4.4 TB of visual data.

Considering the requirements of vision models, which only need a downsampled version of images/videos, we decided to compress the RGB data further. By re-encoding the frames into JPG format with lossy compression, we reduced the size by up to 80\% while maintaining the original pixel resolution and frame rate.

Our physiological data, initially sampled at rates ranging from 500Hz to 4000Hz, was resampled to a uniform rate of 400 Hz. This decision was made after comparing these rates with existing datasets and analyzing patterns in the data, allowing us to use the same model and processing pipeline for all data types. This resampling compressed our physiological data by 20

In total, we collected more than 20 TB of data, which we trimmed and compressed to 2 TB for publication. We plan to release the losslessly compressed version of the visual data along with the full hierarchical labels.

\subsection{Data Formats and Organization}
All data modalities have been split based on Activity labels (level 1 of the hierarchy). Consequently, each activity folder contains sample data corresponding to that specific activity. A single sample file in the dataset represents data from one subject recorded during one of the four recording sessions. If a person performed the same activity multiple times during these sessions, it is still counted as a single sample and included in that one sample file.
\paragraph{Sample Files} The identifier for sample files is formatted as {2-digit subject id}\_{session id}.{file format}, e.g., 01\_3.csv. We maintain separate sample files for each view, activity, and data modality.
\paragraph{File Format}The samples for each modality are formatted to facilitate visualization and use in machine learning and deep learning models. Table \ref{tab:data_formats} provides a summary of the file formats for each group of modalities in our dataset.
To facilitate easy segmentation of data in a hierarchical structure, we extracted frames from visual data. Each file in the visual data represents one frame of the original video, denoted by a 5-digit number that indicates its order in the original video, e.g., 01\_3\_00010.png.

\begin{table}[ht]
\centering
\begin{tabular}{lcccccc}
\toprule
\textbf{Data Modality} & \textbf{RGB} & \textbf{Depth} & \textbf{IR} & \textbf{Depth Confidence} & \textbf{Audio} & \textbf{Timeseries} \\
\midrule
\textbf{File Format} & jpg & png & png & png & wav & csv \\
\bottomrule
\end{tabular}
\caption{File formats for data modalities}
\label{tab:data_formats}
\end{table}
\section{Data Collection Setup}
\label{Appendix: Sensor-Setup}
In our experimental setup, we utilized 16 sensor devices to collect data from humans, the environment, and the scene. These sensors were connected to three host computers that collected streams of data and saved them on drives. To ensure synchronization across these sensor setups, we synchronized the internal clocks of the three hosts using the Network Time Protocol (NTP) \cite{mills2006computer}. This protocol aligns the hosts' clocks by connecting them to a single network time server. The time delay between these synchronized clocks is maintained under 0.1 seconds per day. Every 24 hours, the clocks update their synchronization with the server, preventing any cumulative time differences between the server and the hosts.
Given that our sensors come from various manufacturers and do not support a uniform timestamp protocol, we implemented a multithreaded streaming pipeline. This pipeline connects to each device and manages its stream according to the device's specific protocol. Data is ultimately saved based on the synchronized internal clocks of the hosts.
This configuration allows us to capture the start, elapsed, and end times of each stream using a consistent protocol. Consequently, we can segment the data as needed using these globally synchronized timestamps from the hosts, without concerns about the sensors' varying sampling rates or timestamp protocols. Data synchronization across sensors is reliably achieved through these global timings.

To ensure data quality, Kinect camera, Lidar camera, and radar were calibrated prior to recordings for each room separately, and the configurations were hardcoded for the entire recording sessions in that room. IMU sensors were calibrated during a no-activity period before recordings, and physiological sensors were calibrated to each subject's baseline at every recording session. Environmental sensors did not require specific calibration; however, changes in temperature and humidity were minimal during each recording session.

\begin{figure}[h]
    \centering
    \includegraphics[width=0.7\textwidth]{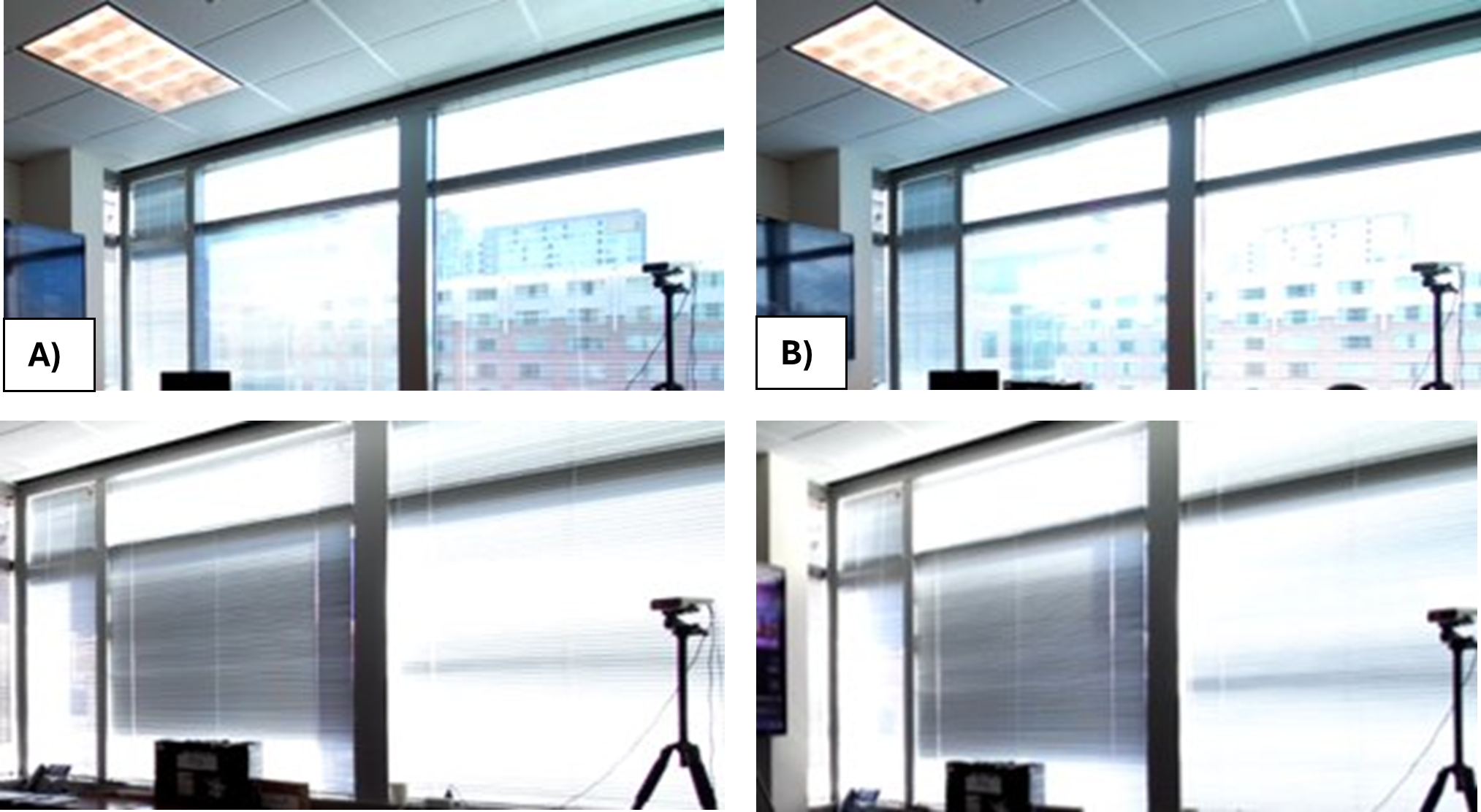}
    \caption{Variation in light condition. A) Shows morning light with open and close curtain. B) shows noon light with open and closed curtain.}
    \label{fig:light condition}
\end{figure}

\paragraph{Room Layouts} We installed four cameras in the living room and office, and three cameras in the kitchen, to collect visual data. Additionally, two stationary IMUs were placed in each room to gather surface vibration data. A single microphone in each room was used to capture audio data separately. We also equipped each room with one light and color sensor, as well as sensors for temperature, humidity, and CO2 levels.
\begin{figure}[!h]
    \centering
    \begin{minipage}{0.48\textwidth}
        \centering
        \includegraphics[width=0.55\textwidth]{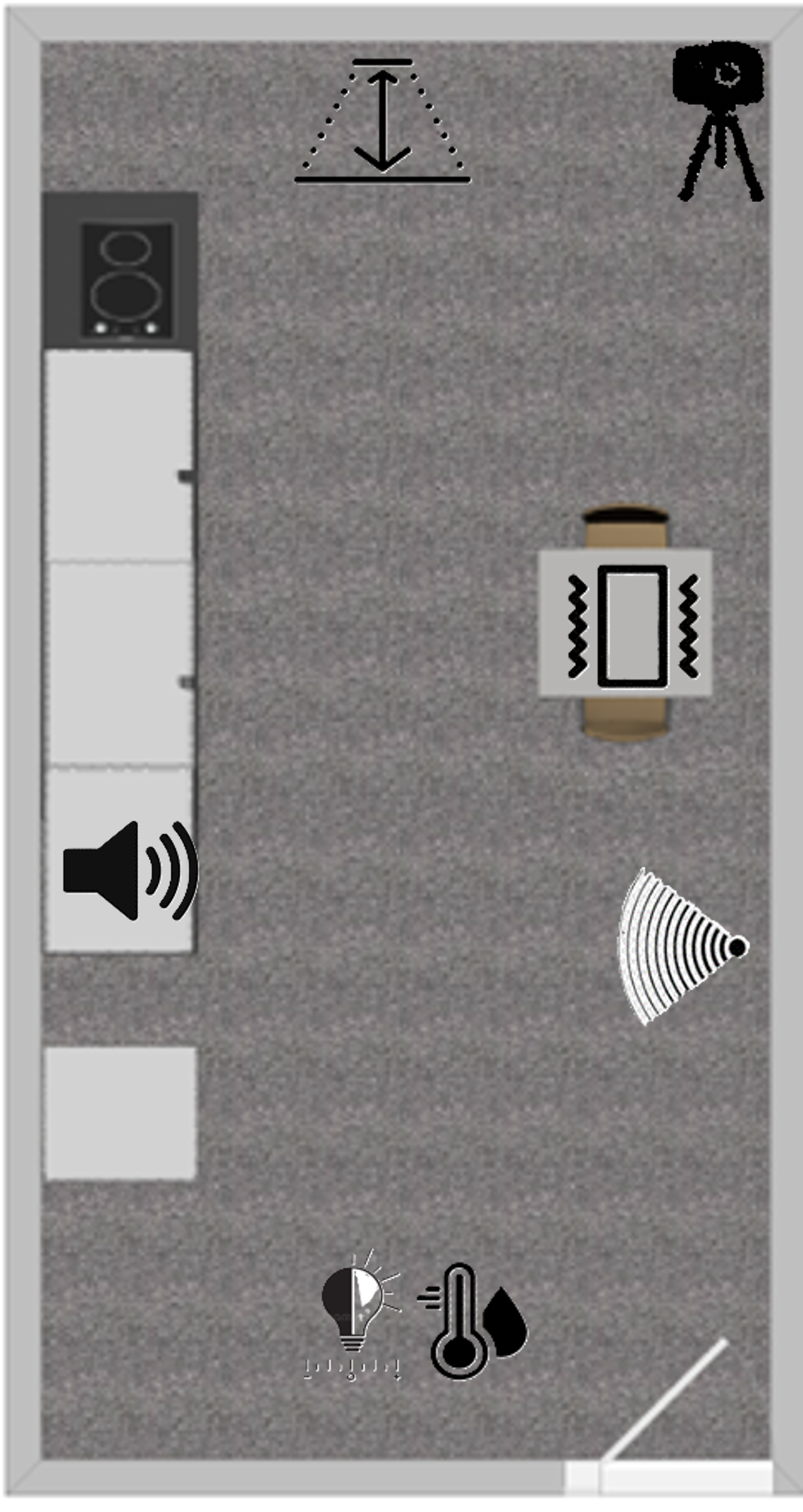}
        \caption{Overall layout of the kitchen and sensor setup in the environment}
        \label{fig:layout_with_sensor_kitchen}
    \end{minipage}\hfill
    \begin{minipage}{0.48\textwidth}
        \centering
        \includegraphics[width=\textwidth]{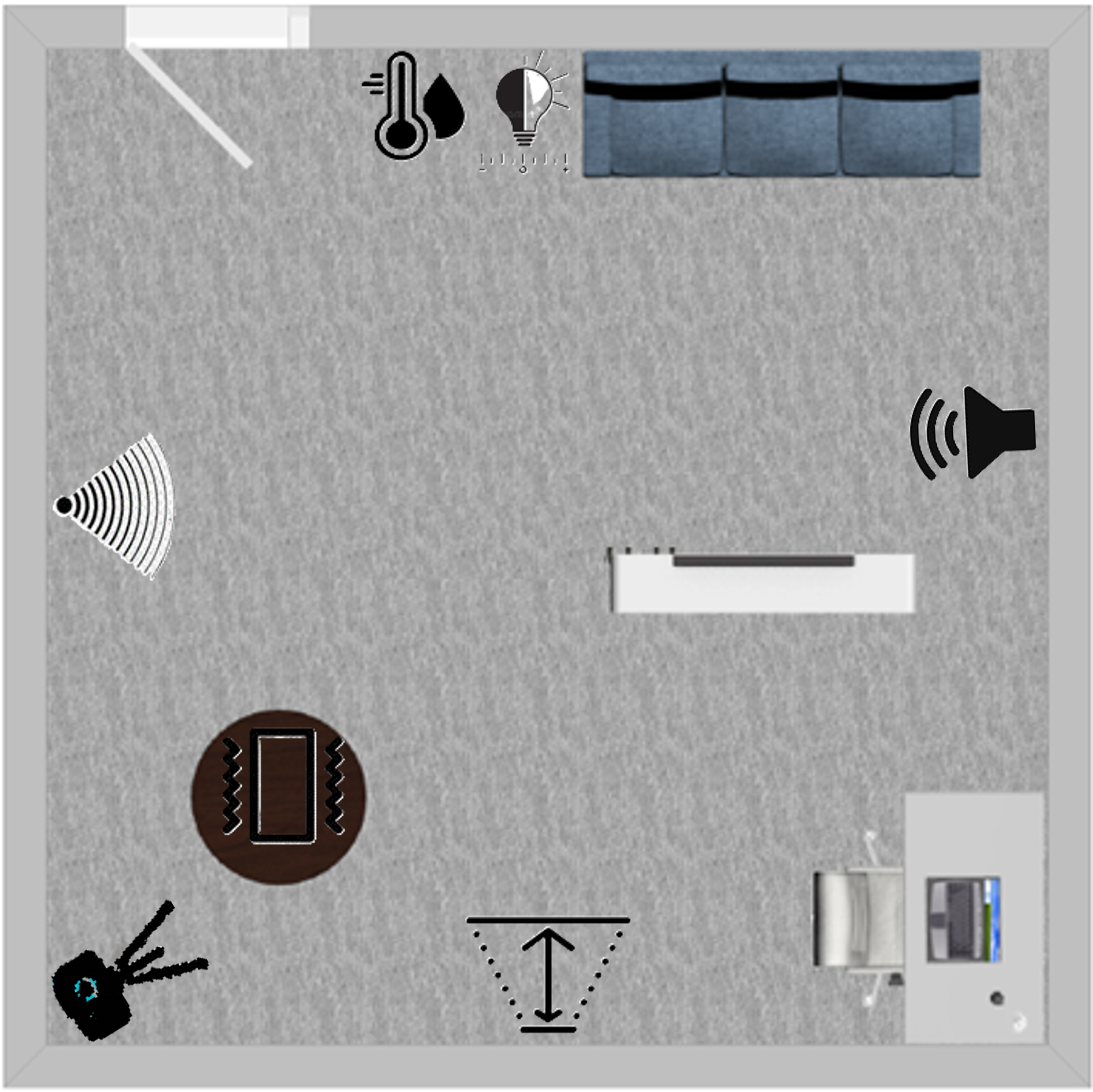}
        \caption{Overall layout of the livingroom and home-office with sensor setup}
        \label{fig:layout_with_sensor_livingroom}
    \end{minipage}
\end{figure}
The remainder of the sensors were attached to the participants' bodies. Each subject wore two wearable IMUs on both hands, and EMG electrodes were installed on both arms. We placed insole pressure sensors in their shoes and provided them with an eye tracker to wear. The biomonitor sensors, specifically a wearable smartlead, require skin contact on the chest. We educated each subject on how to install these sensors, and for privacy reasons, they had the option to self-install them.
The following schematic illustrates the room layouts where we collected our data. As previously mentioned, our setup included varying environmental conditions, such as different background noises, lighting conditions, and air quality across the rooms, totaling six distinct conditions in the data collection process. In Figure~\ref{fig:light condition}, we present a sample of the different lighting conditions in the living room.

\subsection{Hardware Setup and Calibration}
\label{Hardware}
We selected 12 distinct sensors for our experiments and collected data using a total of 16 sensors. Each sensor has its own resolution and sampling rate. We have summarized the hardware specifications of our sensor setup in the Table~\ref{tab:sensors spec}.
\begin{table}[ht]
\centering
\footnotesize
\caption{Hardware specifications}
\begin{tabular}{@{}llll@{}}
\toprule
\textbf{Sensor} & \textbf{Resolution} & \textbf{Sampling rate} \\ \midrule
Kinect Camera\cite{azure_kinect_dk} & 720p , WFOV\_UNBINNED & 15 fps \\
                          Lidar Camera\cite{intel_realsense_l515_datasheet} & 720p, 360 & 30 fps \\ \addlinespace
Microphone & - & 16 kHz \\
                          Wearable IMU & 3-axis Acc , Gyro , Magno & 12 Hz \\
                          Forearm EMG & - & 4000 Hz \\
                          Eye Tracker Camera & 720p & 30 fps \\
                          Gaze & - & 200 Hz \\
                          IMU head & 3-axis & 200 Hz \\
                          Insole Pressure & 8 point pressure & 500 Hz \\
                          Wearable Biomonitors & - & 1000 Hz \\ \addlinespace
 Stationary IMU & 3-axis & 40 Hz \\
                                     Radar & 3 antennas & 1000000 Hz \\
                                       CO2, Humidity, Temperature & - & 0.2 Hz \\
                                       Light & 465  - 615 nm  & 0.5 Hz \\ \addlinespace

\bottomrule
\end{tabular}
\label{tab:sensors spec}
\end{table}
\subsubsection{Cameras}
\begin{figure}[h]
    \centering
    \includegraphics[width=0.5\textwidth]{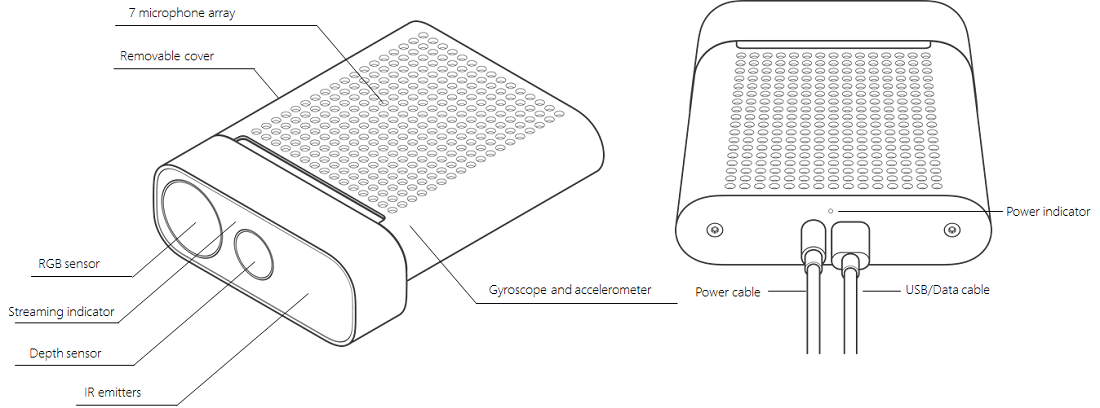}
    \caption{Azure Kinect DK}
    \label{fig:kinect spec}
\end{figure}

\begin{figure}[h]
    \centering
    \includegraphics[width=0.5\textwidth]{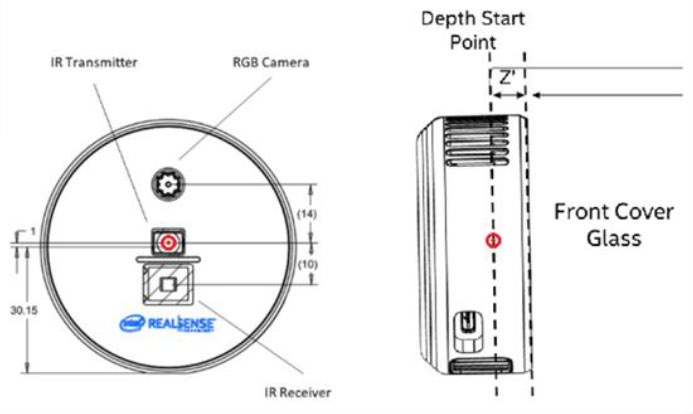}
    \caption{Lidar camera L515}
    \label{fig:intel spec}
\end{figure}

\paragraph{Kinect Camera}
The Kinect camera features an RGB sensor, a time-of-flight depth sensor (which differs from laser-emitting depth sensors), an IMU sensor, and an array of microphones. In our setup, as the cameras are fixed, we did not collect any IMU data from these cameras. Unfortunately, a limitation of the Kinect camera development kit is that it does not allow simultaneous recording of visual and audio data with the provided SDK. You can see the sensor setup of the Kinect DK in Figure~\ref{fig:kinect spec}. We can calibrate this camera by adjusting sensor parameters and modes, balancing exposure, and setting white points, among other settings. The optimal configuration is then passed as parameters during recording.
We have tested the camera in each room, recording data with minimal artificial adjustments to the input of the lenses.
\paragraph{Lidar Camera}
Our Lidar camera, the L515, is a compact model designed for indoor use and manufactured by Intel Figure~\ref{fig:intel spec}. It features a separate RGB sensor and a laser emitter for depth calculation. Calibration of this camera involves adjusting the RGB and depth settings separately and then aligning the outputs of the two sensors to achieve higher quality depth and RGB frames. We tested these cameras in each room, saved the settings as configuration files, and later loaded these profiles during recording sessions to ensure optimal performance.
\subsubsection{Wearable Biomonitors}
The BioMonitor SmartLead captures five different signals: ECG signal, heart rate, respiration, respiration rate, and R-R interval.

\paragraph{ECG:}
This signal is directly measured using two electrodes and one ground. The placement of sensors and electrodes can slightly alter the ECG waveform. For more details on electrode configurations and the types of signals they produce, see Appendix B.

\paragraph{Heart Rate:}
This signal is derived from the ECG waveform by counting the peaks of each QRS complex.

\paragraph{R-R Interval:}
This signal, calculated as the inverse of heart rate, is defined as the time elapsed between two successive R-waves of the QRS complex.

\paragraph{Respiration:}
Respiration is measured by detecting changes in tissue impedance between two electrodes. The change is quantified as the ratio of torso impedance change to overall impedance. Note: This change is subtle, and proper skin preparation is crucial for accurate measurement.

\paragraph{Respiration Rate:}
This signal is determined by counting the peaks of the respiration signal.

These sensors require direct skin contact with the chest area. participants were given the option to install the sensor themselves or have it installed by a technician. Most participants preferred to self-install the sensors.

\begin{figure}[h]
    \centering
    \includegraphics[width=0.6\textwidth]{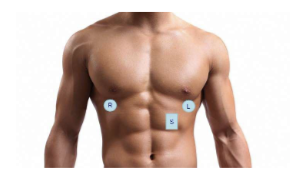}
    \caption{Electrode placement}
    \label{fig:biomonitors}
\end{figure}

We used the sensor placement depicted in Figure~\ref{fig:biomonitors} to collect respiration and ECG signals. Although the ECG waveform does not represent a standard ECG lead configuration, it closely resembles it.
Calibration of biomonitoring sensors involves recording inactive values for a few seconds and setting these as the baseline for subsequent recordings. This calibration must be performed for each subject at every recording session.

\subsubsection{Eye Tracker}
We employed a Noen eye tracker, which does not require calibration and can be powered and operated via a USB-C connection to an Android mobile phone. The device integrates multiple features into a single module, including an egocentric camera, an eye camera, an IMU sensor, and a microphone.
\begin{figure}[h]
    \centering
    \includegraphics[width=0.4\textwidth]{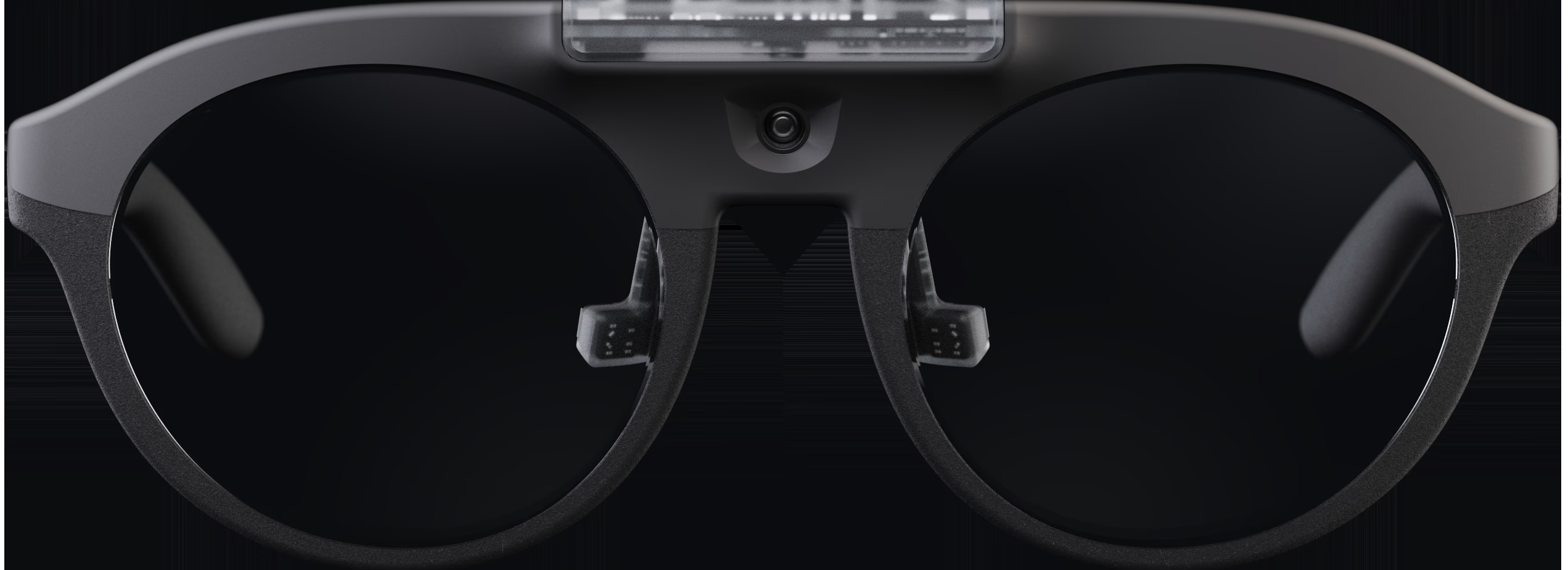}
    \caption{PupilLabs Neon calibration-free eye tracking module}
    \label{fig:eye tracker}
\end{figure}
\subsubsection{Insole Pressure}
\begin{figure}
    \centering
    \includegraphics[width=0.4\textwidth]{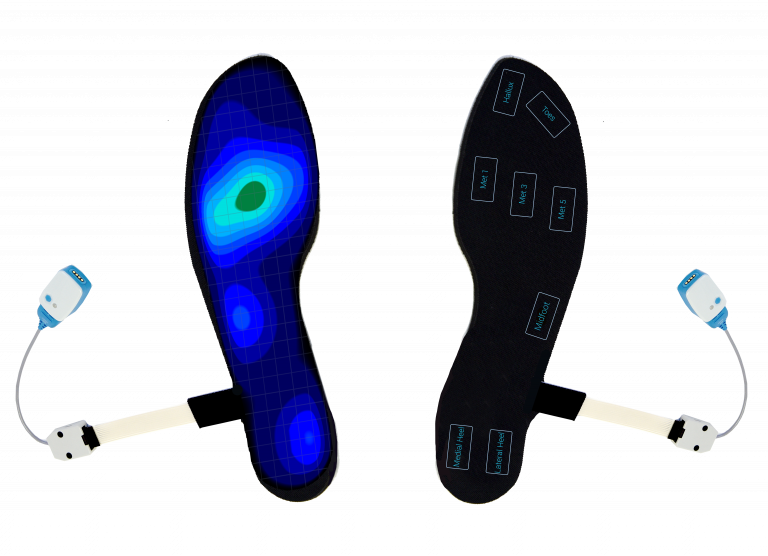}
    \caption{Insoles with 8 point pressure sensors}
    \label{fig:insoles}
\end{figure}

Insole pressure sensors, as depicted in Figure~\ref{fig:insoles}, were placed in the participants' shoes. Calibration of these sensors involves measuring the pressure exerted by a single foot on each side. Before starting recordings at each session, participants were asked to stand on one foot while keeping the other foot in the air.
\subsubsection{Radar}
Our 60 GHz radar sensor operates in FMCW mode with a 5.5 GHz bandwidth. This compact yet powerful indoor device is used for presence detection, distance measurement, and trajectory tracking. Calibration of the device is necessary based on the intended application and the distances it needs to cover. Settings can be saved and applied during recordings. We have tested our radar settings in each room to ensure robust detection of subject movements. However, the field of view (FoV) of the radar is narrower than its nominal FoV, which limits its ability to capture presence over wider angles.

The rest of the sensors do not require specific configurations. They only need to initiate the stream and allow the buffer to warm up in an inactive mode before recording begins.


\subsection{Alternative Sensor Setup}
Table~\ref{tab:sensor_economy} outlines cost-effective hardware alternatives that meet or closely approximate our specifications.

\begin{table}[htb]
\centering
\resizebox{\textwidth}{!}{%
\begin{tabular}{@{}lllr@{}}
\toprule
\textbf{Sensor} & \textbf{Proposed Alternative} & \textbf{Key Specifications} & \textbf{Est. Price} \\ \midrule
Kinect Camera & \textbackslash{}productlink\{https://www.orbbec.com/femto-bolt/\}\{Orbbec Femto Bolt\} & 1080p RGB, 1MP Depth, 30 fps & \$450 \\
Lidar Camera (360\textbackslash{}textdegree) & \textbackslash{}productlink\{http://www.slamtec.com/en/Lidar/S2\}\{Slamtec RPLIDAR S2\} & 30m Range, 32,000 samples/s & \$450 \\
Microphone & Generic USB Microphone & \textgreater{}16 kHz Sampling Rate & \$50 \\
Wearable IMU (x2) & \textbackslash{}productlink\{https://mbientlab.com/product/metamotionc/\}\{MbientLab MetaMotionC\} & 9-DoF, up to 100 Hz, BLE & \$200 (\$100 ea.) \\
Forearm EMG & \textbackslash{}productlink\{https://shop.openbci.com/products/cyton-biosensing-board-8-channel\}\{OpenBCI Cyton Board\} & 8-channel, up to 2000 Hz & \$500 \\
Eye Tracker / Gaze & \textbackslash{}productlink\{https://pupil-labs.com/products/core/\}\{Pupil Labs Pupil Core\} & 200 Hz Gaze Tracking & \$2,400 \\
IMU head & \textbackslash{}productlink\{https://mbientlab.com/product/metamotionc/\}\{MbientLab MetaMotionC\} & 9-DoF, up to 200 Hz, BLE & \$100 \\
Insole Pressure & XOSensor Research Kit & 8-point, High Sample Rate & \$2,500 \\
Wearable Biomonitors & \textbackslash{}productlink\{https://bitalino.com/en/board-kit\}\{BITalino (r)evolution Kit\} & ECG, EDA, EMG, up to 4kHz & \$200 \\
Stationary IMU & Arduino + MPU-6050 Module & 6-DoF, \textgreater{}40 Hz & \$40 \\
Radar & \textbackslash{}productlink\{https://www.ti.com/tool/IWR6843AOPEVM\}\{TI IWR6843AOP EVM\} & 60-64 GHz mmWave & \$350 \\
CO2, Humidity, Temp & \textbackslash{}productlink\{https://www.adafruit.com/product/5190\}\{Sensirion SCD41 Breakout\} & High-accuracy NDIR CO2 & \$55 \\
Light & \textbackslash{}productlink\{https://www.adafruit.com/product/4698\}\{Adafruit AS7341\} & 11-channel Visible Light & \$15 \\ \midrule
\textbf{TOTAL} &  &  & \textbf{\textbackslash{}textasciitilde\$7,360} \\ \bottomrule
\end{tabular}%
}
\caption{Alternative sensor setup for smaller budget projects}
\label{tab:sensor_economy}
\end{table}
Based on the components selected, the total estimated cost is less than \textbf{\$10,000}.
\subsection{Recording Procedure}
Each recording session begins by explaining the sensors, the expectations, and the objectives of the study to the participants, followed by obtaining their consent. participants then review a list of activities and familiarize themselves with the environment. We ask them to perform each activity at a normal pace and avoid stopping any activity before ten seconds have elapsed. Next, we assist them in wearing the sensors and provide instructions on how to install the BioMonitor SmartLead.
After all sensor devices are connected to the program, we start the calibration of physiological and IMU sensors in an inactive mode; specific calibration procedures have been explained earlier.
Then, we ask participants to stand outside the room and enter only after we count from 1 to 3. During these three seconds, we initiate the streaming pipeline. We signal for the entrance of the subject to the room and begin recording the streams at that moment.
\subsection{Activity Ordering and 
Instructions}
This section presents a sample ordering of activities for one session in each environment (refer to Table~\ref{tab:kitchen_instructions} and~\ref{tab:livingroom_instructions}. The tasks listed are those we asked and expected the participants to perform, organized into the hierarchical taxonomy of our dataset.
The only instructions provided to participants included a fast-forwarded video of expected tasks prior to their recording schedule and a list of daily activity tasks without specific instructions on how to perform them. 
participants were allowed to alter the order of activities; they were only required to signal their intent to begin the next activity by saying "next" before starting the next instance.
During each session, participants were asked to perform similar tasks with slight variations, such as watching different programs, playing different games, or choosing to use the dishwasher instead of washing dishes by hand.
\begin{table}[h]
\centering
\footnotesize 
\resizebox{\textwidth}{!}{%
  \begin{tabular}{ll}
    \begin{minipage}{0.48\textwidth}
      \centering
      \begin{tabular}{cl}
        \hline
        \textbf{Order} & \textbf{Activity} \\
        \hline
        1  & Enter the room \\
        2  & Turn on the light \\
        3  & Watch TV \\
        4  & Play video games \\
        5  & Use cellphone \\
        6  & Read Magazine \\
        7  & Carry and move heavy objects \\
        8  & Carry and move large objects \\
        9  & Move chair \\
        10 & Play games on PC \\
        11 & Prepare an email \\
        12 & Present \\
        13 & Take a quiz \\
        14 & Read instructions \\
        15 & Write on a paper \\
        16 & Sleep \\
        17 & Exercise - Jump rope \\
        18 & Exercise - Squat \\
        19 & Exercise - Jumping jacks \\
        20 & Exercise - Yoga Stretch \\
        21 & Exercise - Simple jump \\
        22 & Exercise - Punch \\
        23 & Turn off the light \\
        24 & Exit the room \\
        \hline
      \end{tabular}
      \caption{First Session Instructions}
      \label{tab:livingroom_instructions}
    \end{minipage}
    &
    \begin{minipage}{0.48\textwidth}
      \centering
      \footnotesize
      \begin{tabular}{cl}
        \hline
        \textbf{Order} & \textbf{Activity Description} \\
        \hline
        1  & Carry groceries \\
        2  & Put and organize items in the cabinet \\
        3  & Wash your hands \\
        4  & Open the utensil drawer and take what you need\\
        5  & Open the cabinet and take what you need \\
        6  & Take salad ingredients from the refrigerator \\
        7  & Wash ingredients as you see fit \\
        8  & Make a salad (chop, peel, and mix ingredients) \\
        9  & Put the salad in a plate or bowl \\
        10 & Open the can and add it to the salad \\
        11 & Cut a lemon and squeeze it into the salad \\
        12 & Take pancake ingredients from the cabinet \\
        13 & Make pancake batter \\
        14 & Put the batter in a pan on the stove \\
        15 & Fill the electric water boiler and make hot water \\
        16 & Take a cup and make instant coffee/tea \\
        17 & Fill a water pitcher and pour a cup of water \\
        18 & Set the table with the foods and drinks \\
        19 & Eat and drink \\
        20 & Throw out leftovers in the trash \\
        21 & Wash dishes in the sink \\
        22 & Dry dishes using a towel \\
        23 & Clean the table \\
        24 & Take the broom and clean the floor \\
        \hline
      \end{tabular}
      \caption{Third Session Instructions}
      \label{tab:kitchen_instructions}
    \end{minipage}
  \end{tabular}%
}
\end{table}

\section{Annotation Setup}
\label{Appendix: Annotation}
Initially, an experimenter reviews the recorded data to ensure there is no sensitive information or events that need to be removed and to verify that the activities performed by each subject align with the top activity labels. Subsequently, the data is segmented into Level 1 (L1) activity labels for each subject and session, serving as the basis for hierarchical annotations.

We instruct annotators to first view video footage from two camera perspectives to identify the high-level activity associated with the data under review. They then select the most fitting label for that video clip. Since the hierarchical labels are prepared in advance, annotators only need to decide which labels best describe each data level. They are expected to watch the video, review the annotation options, and choose an appropriate label.

Annotators are also tasked with providing a scene description, noting the subject and any interactions occurring within each video segment.

\subsection{Framework}

This section introduces the straightforward framework we developed for the hierarchical labeling of the data, along with a detailed step-by-step annotation procedure.

\begin{figure}[h]
    \centering
    \includegraphics[width=0.9\textwidth]{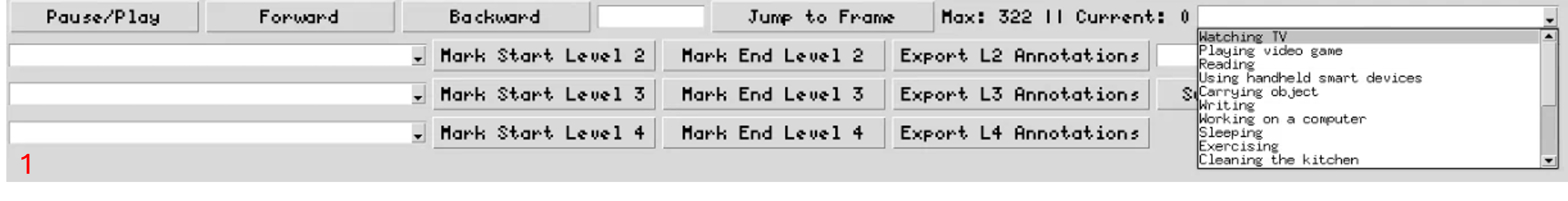}
    \caption{Step 1, selecting the Level 1 label associated with the video clip.}
    \label{fig:menu1}
\end{figure}

Annotators have the capability to play, pause, and navigate through each video clip. Hierarchical annotations are recorded at a frame-level basis. To determine the first-level label, annotators need to watch a few seconds of the video, then select the appropriate label from the first menu and mark the beginning of the activity Figure~\ref{fig:menu1}. 
    

    

    

\begin{figure}
    \centering
    \includegraphics[width=0.9\textwidth]{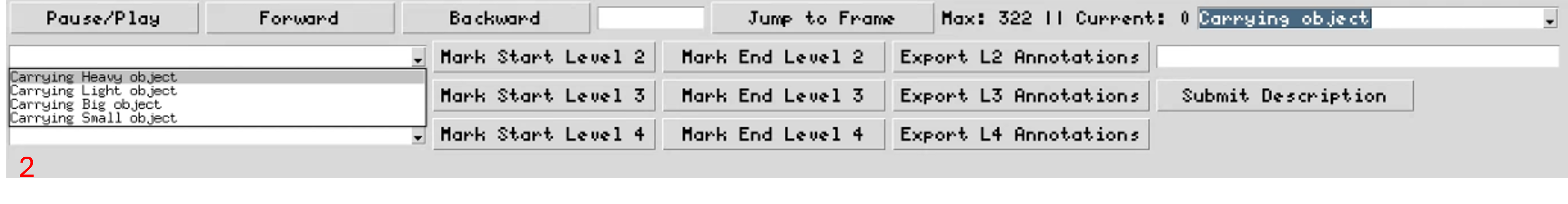}
    \caption{Step 2, selecting associated level 2 label with the video clip and the higher level label.}
    \label{fig:menu2}
\end{figure}

\begin{figure}
    \centering
    \includegraphics[width=0.9\textwidth]{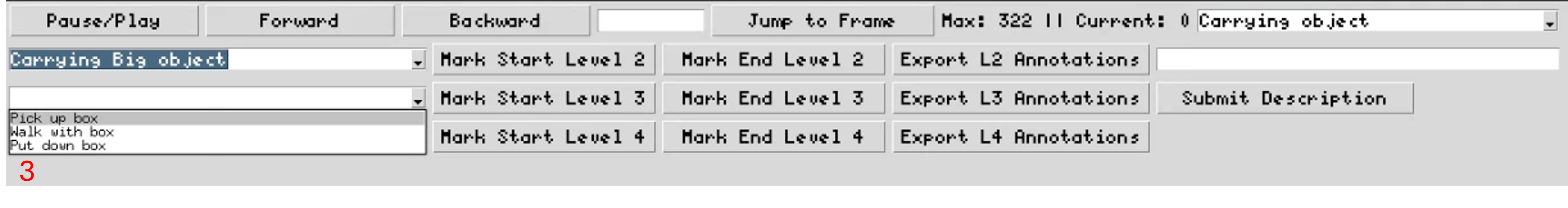}
    \caption{Step 3, selecting associated level 3 and then level 4 label with the video clip and the higher level label.}
    \label{fig:menu3}
\end{figure}

When annotators identify the L1 label, they will access hierarchical labels specific to that instance of the L1 label, as depicted in Figure~\ref{fig:menu2}. They are then required to mark the beginning of that specific label.
To streamline the annotation process and reduce the need for repeated viewings, we have restricted the option to mark the end of each level label until all its subordinate levels have been selected. This approach is analogous to the mathematical rule of solving expressions within inner parentheses before those in outer parentheses. In this context, the "outer parentheses" represent higher-level labels in the hierarchy. Annotators are allowed to select the end of each level label only after they have marked the lowest level label, as illustrated in Figure~\ref{fig:menu3}
Once the entire clip has been viewed, annotators must submit a description that will be associated with the Level 2 timestamps.
\subsection{Annotation Workflow}
As described in the previous section, our data consisted of untrimmed sets of activities performed by each subject. We first trimmed the recordings into activity clips, and annotations were performed on these trimmed parts of the data. There are two stages to annotating with this taxonomy. First, to minimize subjectivity, a list of possible elements for each concept is produced following the rules of the taxonomy. Annotators then select from these options dynamically while reviewing visual data.

We recruited multiple annotators and provided them with a hierarchical annotation program preloaded with a hierarchy list. Annotators were also asked to enter a description of the activity concept based on the visual scene and dynamics. The annotation program prevents annotators from selecting lower levels prior to higher-level annotations and from selecting start/end times outside of their higher-level parent's time span. Overlaps between same-level annotations for a specific file are allowed to ensure a smooth chain of actions and procedures.

The quality control process involves randomly selecting 50\% of the annotated data and segmenting clips based on each level of annotations for visual inspection. If there are undesirable time span selections or incorrect label selections, the file is put up for revision by a different annotator.

\subsection{Labels}

\begin{figure}[ht]
    \centering
    \scriptsize
    \begin{subfigure}{0.9\textwidth}
        \centering
        \includegraphics[width=\linewidth]{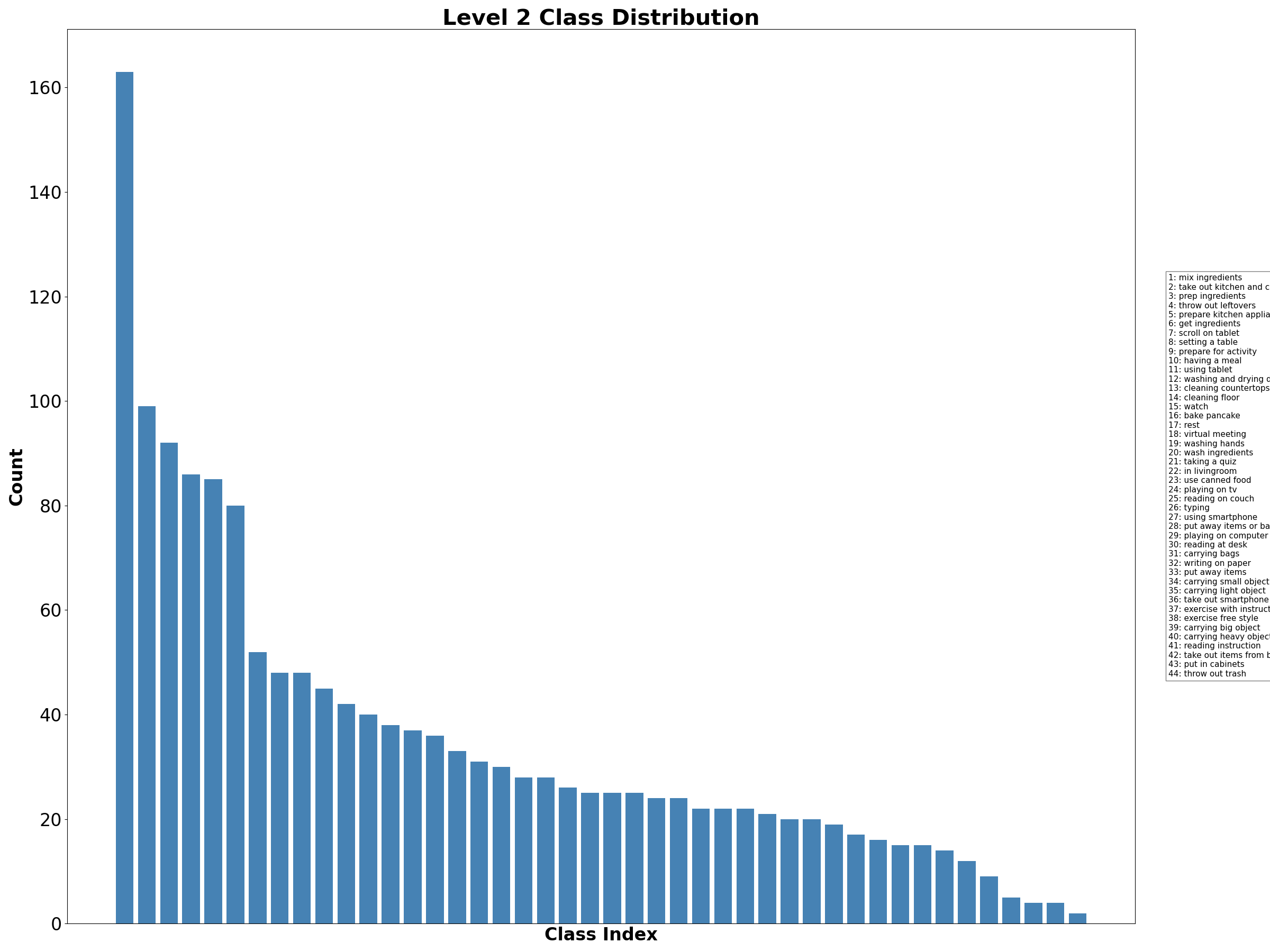}
        \label{fig:level2_distribution}
    \end{subfigure}
    \begin{subfigure}{0.9\textwidth}
        \centering
        \includegraphics[width=\linewidth]{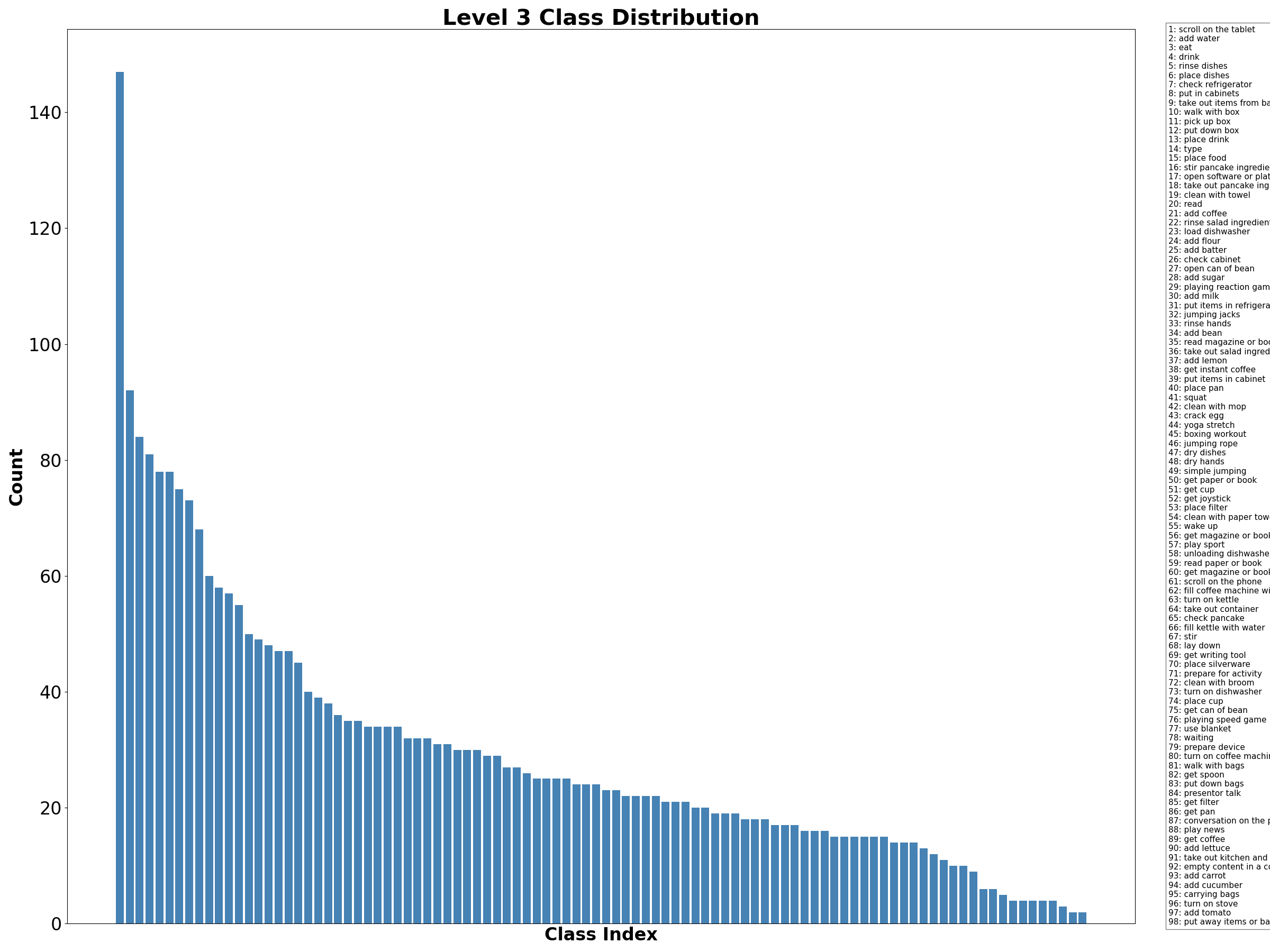}
        \label{fig:level3_distribution}
    \end{subfigure}
    \caption{ \scriptsize Class distributions of Level 2 actions (left) and Level 3 procedures (right) in the \texttt{DARai} hierarchy. The corresponding class labels are listed in Appendix \ref{Appendix: Annotation} in the same order as presented in the plot.}
    \label{fig:Class distributions}
\end{figure}

The labels in the \texttt{DARai} dataset are not balanced across classes, as shown in Figure~\ref{fig:Class distributions}. This imbalance arises from naturally occurring variations in human activities. Some classes, such as “Sleeping” or “Making coffee,” occur rarely, which can skew typical accuracy metrics. Additionally, each individual performs certain activities differently from others \ref{fig:hierarchy_seg}. Datasets with predefined structures, such as those referenced in Table~\ref{tab:dataset_multi}, often record class samples individually and aim for similar sample sizes per class and subject, thereby mitigating such imbalances. Unscripted datasets, however, can only artificially enforce class balance in their data. The \texttt{DARai} dataset contains unscripted daily activity recordings and exhibits class imbalance across all three hierarchy levels. For example, \textsf{Using utensil} is a procedure shared between many food preparation and eating activities, whereas \textsf{Cleaning the floor with a mop} is an action specific to fewer cleaning tasks. The unbalanced distribution of fine-grained actions and procedures makes recognition at lower-levels of hierarchy challenging. The distribution of labels shown in Appendix \ref{Appendix: Annotation} Figure~\ref{fig:Class distributions}.  \texttt{DARai} offers a valuable way to study modality specific' strengths and limitations in complex, everyday scenarios across three level of details.

\section{Participants}
\label{Appendix: Participants}
\subsection{Participation Call}

We recruited participants for an in-person research study aimed at generating public datasets through wearable sensors and video recordings of everyday activities. Recruitment methods included email and flyers.
We obtained approval from an Institutional Review Board (IRB) to conduct this study and collect data from human subjects.

\subsubsection{Scheduling}
To organize recording sessions, we utilized a scheduling platform that accommodated the available timings of the volunteers.

\begin{figure}[h]
    \centering
    \includegraphics[width=0.5\columnwidth]{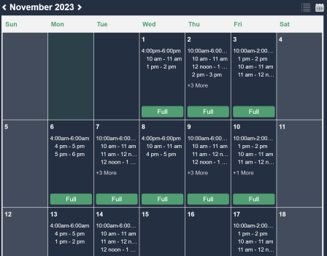}
    \caption{Schedule}
    \label{fig:schedule}
\end{figure}
\subsubsection{Eligibility Criteria}
\begin{itemize}
    \item Be between 18 – 89 years old
    \item Be able to consent without requiring surrogate consent from a legally authorized representative
\item Agree to allow your de-identified data (face blurred) to be disseminated for research purposes only
\item Be comfortable with performing everyday activities without wearing glasses. Contact lenses are permitted
\end{itemize}
\subsubsection{Expectation}

Participants were expected to engage in four sessions at the OLIVES lab, each lasting approximately 30 minutes. During these sessions, they wore state-of-the-art wearable sensors and performed common activities such as preparing food, watching TV, playing video games, and light exercise. Multiple sensor devices recorded their actions. Participants were closely monitored, and any sign of discomfort would immediately end the session. Each session concluded with a survey.

\subsubsection{Compensation}

Participants were compensated with a \$5 Amazon Gift Card for each session attended. Additionally, participants who attended all four sessions received a total of \$40 in Amazon Gift Cards, including a \$20 bonus for full participation.

\subsection{Personally Identifiable Information}

We maintained the anonymity of each participant by using an anonymous subject ID and removing names from our dataset. Each session was recorded using a combination of a subject ID and session ID. Access to a corresponding list of subject IDs and the scheduling list is restricted to experimenters only. To further protect privacy, we blurred subjects' faces in the video recordings using a face anonymization framework, which can be accessed at \url{https://github.com/ORB-HD/deface}, with settings that include a mask scale of 1.7, a detection threshold of 0.15, and a box blurring effect.

Additionally, we do not record real conversation audio. Instead, we anonymized subjects' audio in the dataset using a speaker anonymization framework introduced at \url{https://github.com/sarulab-speech/lightweight_spkr_anon}
\begin{figure}[h]
    \centering
    \includegraphics[width=0.3\columnwidth]{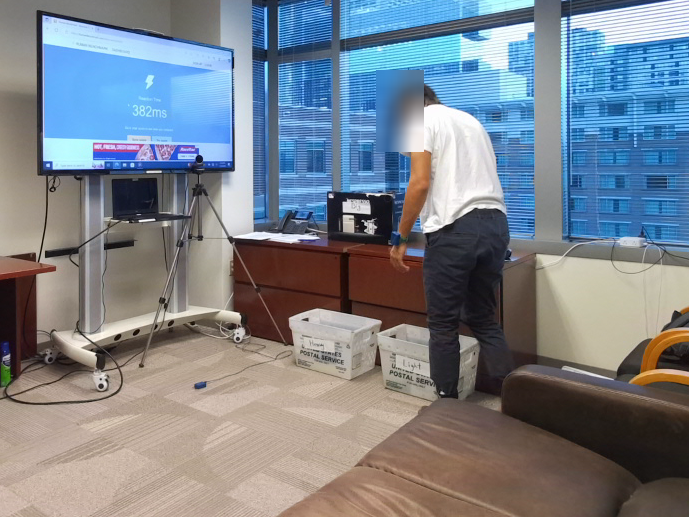}
    \caption{Blurred Face Example}
    \label{fig:enter-label}
\end{figure}

\subsection{Participant Demographics}
\begin{figure}[h]
    \centering
    \includegraphics[width =1\columnwidth]{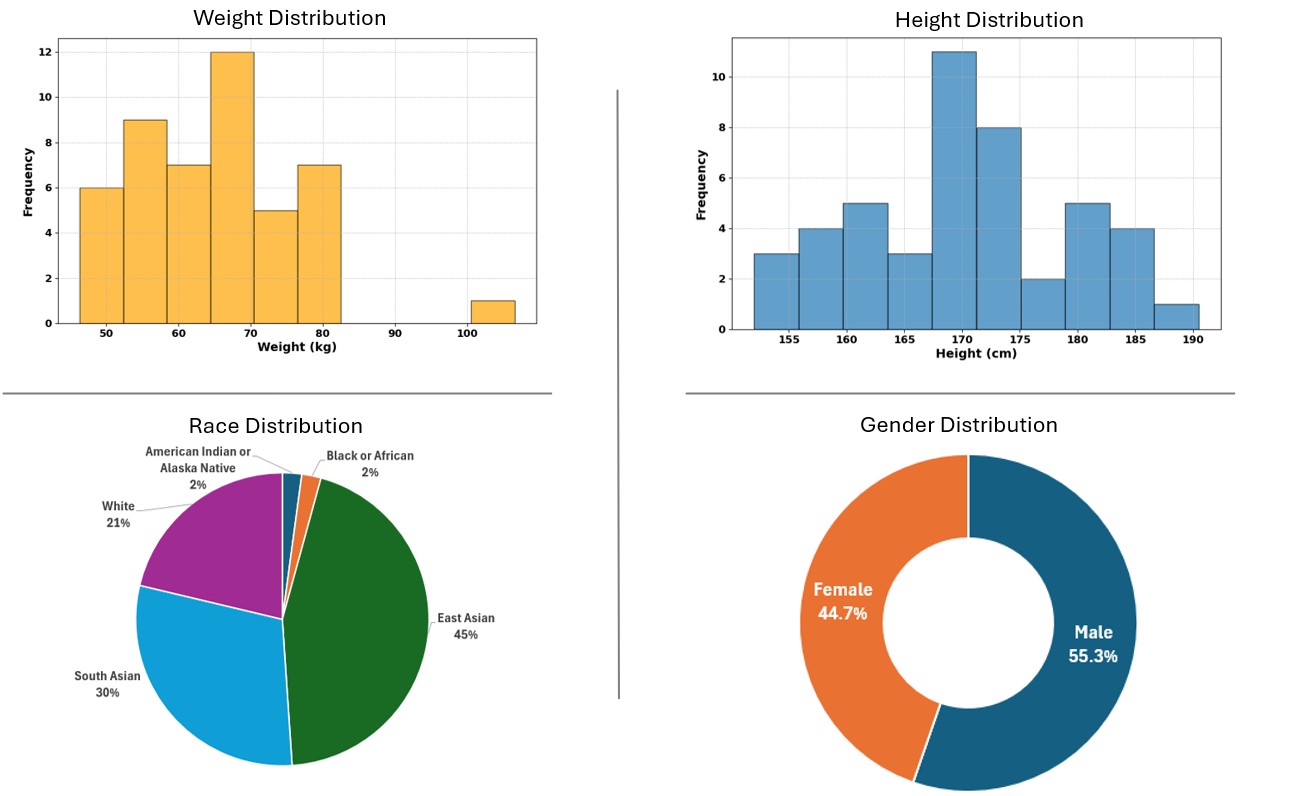}
    \caption{Demographic and BMI distribution of DARai's subjects}
    \label{fig:participanstat}
\end{figure}

We collected data from 50 participants, each of whom was asked to provide demographic information at the end of their session. This demographic data, which includes a variety of heights and weights impacting physiological signals, is primarily representative of a college campus demographic. The statistics collected from participants are illustrated in Figure~\ref{fig:participanstat}.

\subsection{Participant Surveys}
\label{Appendix: Surveys}

\begin{figure}[h]
    \centering
    \includegraphics[width =0.4\columnwidth]{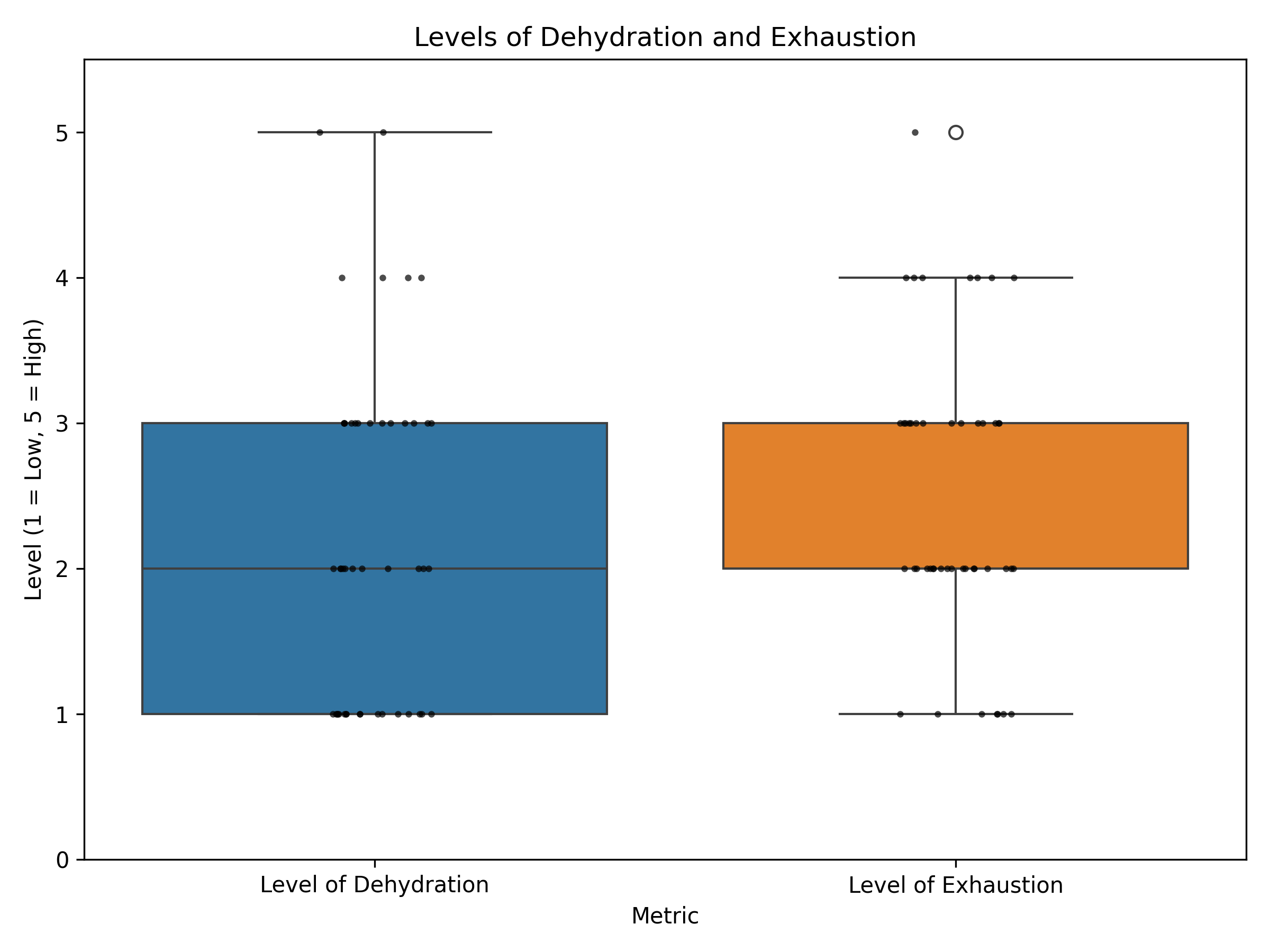}
    \caption{Participants exhaustion and dehydration levels after completing one recording session}
    \label{fig:participantexhaustion}
\end{figure}

At the end of each session, we asked participants to complete a survey. Initially, participants rated their levels of exhaustion and dehydration on a scale from 0 to 5, with this data displayed in Figure~\ref{fig:participantexhaustion}. 

\begin{figure}[h]
    \centering
    \includegraphics[width =0.7\columnwidth]{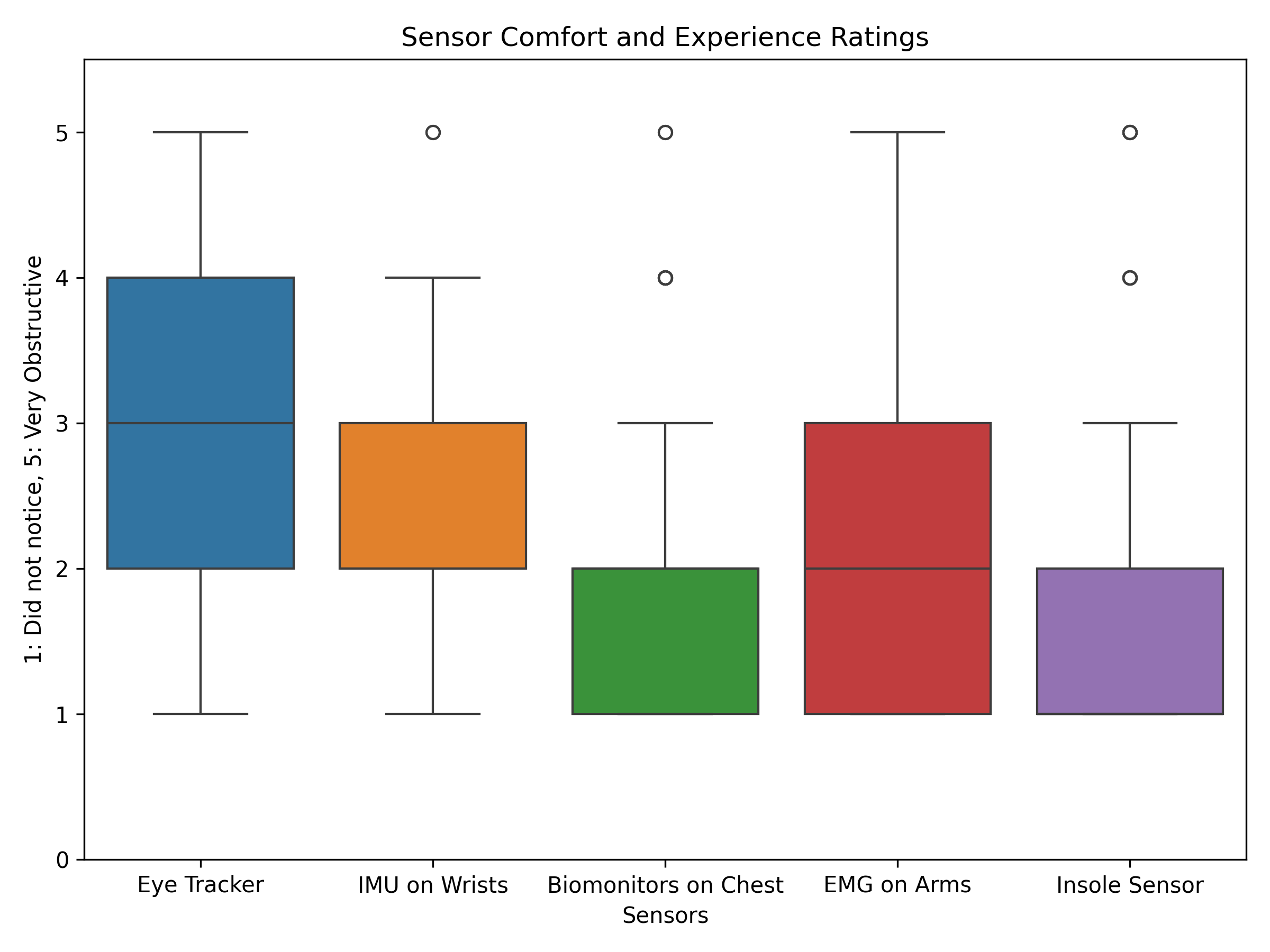}
    \caption{Participants' experience of comfortness with each sensor rated from 0 to 5. 0 means 'did not notice it' and 5 means 'very obtrusive'.}
    \label{fig:sensorexperience}
\end{figure}

Furthermore, participants evaluated their experience with the sensors, using a scale from 0 ('did not notice it') to 5 ('very obtrusive'). The results reflecting the obtrusiveness levels of the sensors are detailed in Figure~\ref{fig:sensorexperience} 

\section{Experimental Setup and Results}
\label{Appendix: ablation tables}
\subsection{Activity Recognition}
While the data presented in this paper is not specifically designed for activity recognition benchmarking, we chose end-to-end activity classification as the first experiment. In this experiment, we compare unimodal and multimodal setups and report the results using hierarchical labels.

\subsubsection{Unimodal analysis} 
Our data comprises both 1D signals and 3D frame sequences (activity clips). In the unimodal experiment setup, we perform action recognition on different modalities separately and each modality has its own experimental setup. In the following sections we explained and compared each setup in detail.
\paragraph{ Timeseries data} 
Our dataset contains 1D signal data with a wide range of sampling rates, from 12 Hz for IMUs to 10,000 Hz for ECG monitoring. Each signal is preprocessed to handle missing values and NaNs through simple interpolation, followed by normalization on a per-subject basis. To facilitate multimodal experiments, we resampled signals from forearm muscle EMG, insole pressure, and biomonitoring to a uniform sampling rate of 400 Hz. One approach to avoid overwhelming a model with longer sample sequences is how language models and specifically transformers treat input data. We create a set of fixed length subsequences of samples randomly selected by a temporal window and use them as input to a model.  Without any additional filtering applied on normalized data, we have trained an LSTM, BiLSTM network and a vanilla transformer from scratch on the set of 3-5 subsequences with a length of 10 seconds and aggregate the predication on entire set of subsequences for a sample. Average class-wise accuracy for these models are reported in Table~\ref{tab:l1_timeseries_result}.
For multichannel time series signals (IMU, EMG, insole, biomonitor, and gaze data), we use a transformer encoder~\citep{csshar}. The results for different architectures can be found in Appendix~\ref{Appendix: ablation tables}. All signals are resampled to a fixed-size sequence, with 100 to 2000 timesteps selected to achieve an average activity length of 10 seconds. They then pass through a 1D convolution layer, instead of a linear layer, before being fed into the transformer encoder. For the encoder transformer, we use a modified version of the one from \cite{csshar}. Initially, the input data is processed through three one-dimensional CNN blocks, which have [32, 64, 128] feature maps and a kernel size of 5. The resulting feature maps are then fed into two self-attention blocks, each with two heads.



\begin{table}[!h]
\centering
\footnotesize
\begin{tabular}{@{}ccc@{}}
\toprule
\multicolumn{3}{c}{Average Class Accuracy of L1 Activities for Time Series Data} \\ \midrule
\multicolumn{1}{c}{Data Modality} & \multicolumn{1}{c}{Training Model} & \multicolumn{1}{c}{Weighted Performance} \\
\midrule
IMU      & Transformer & 0.418 \\
EMG      & Transformer & 0.41  \\
Gaze     & Transformer & 0.31  \\
Bio      & Transformer & 0.39  \\
Insole   & Transformer & 0.5   \\
IMU      & LSTM        & 0.32  \\
EMG      & LSTM        & 0.14  \\
Gaze     & LSTM        & 0.16  \\
Bio      & LSTM        & 0.22  \\
Insole   & LSTM        & 0.18  \\
IMU      & BiLSTM      & 0.32  \\
EMG      & BiLSTM      & 0.14  \\
Gaze     & BiLSTM      & 0.15  \\
Bio      & BiLSTM      & 0.208 \\
Insole   & BiLSTM      & 0.12  \\
\bottomrule
\end{tabular}
\vspace*{.1mm}
\caption{Unimodal analysis of Activity detection based on L1 labels (Activities) for Time series data}
\label{tab:l1_timeseries_result}
\end{table}

\begin{table}[!h]
\centering
\footnotesize
\begin{tabular}{@{}ccc@{}}
\toprule
\multicolumn{3}{c}{Average Class Accuracy of L2 Activities for Time Series Data} \\ \midrule
\multicolumn{1}{c}{Data Modality} & \multicolumn{1}{c}{Training Model} & \multicolumn{1}{c}{Weighted Performance} \\
\midrule
IMU      & Transformer & 0.3 \\
EMG      & Transformer & 0.28  \\
Gaze     & Transformer & 0.1595  \\
Bio      & Transformer & 0.22  \\
Insole   & Transformer & 0.32   \\
\bottomrule
\end{tabular}
\vspace*{.1mm}
\caption{Unimodal analysis of Activity detection based on L2 labels (Actions) for Time series data}
\label{tab:l2_timeseries_result}
\end{table}

The choice of the number of subsequences was made based on the original data length and sampling rate.

\paragraph{ Timeseries Image Representation}
We extract features from these 1D data using Gramian Angular Field (GAF) to create a 2D representation from the signal sequences. This method is one of Imaging Time Series approach to use with non-recurrent deep learning models. The GAF algorithm encodes one-dimensional time series data as 2D vector using angular cosine and
time stamp as the radius \cite{wang2015encoding} and has proven effective in Human Activity Recognition (HAR) for various 1D signal types, including IMU and EMG data \cite{emggaf, imugaf}. Before generating GAF images, 1D sequence data is transformed into identifiable samples by dividing the data into equally sized segments known as temporal windows, with a 50\% overlap between each window to ensure maximum temporal relation per sequence. These generated GAF images are then fed into a ResNet model \cite{he2016deep} for training. For evaluation, we use majority voting based on the image prediction results from the same sequence, averaging the prediction results for each activity sequence across all images. A summary of results is reported in Tables~\ref{tab:uni_result},~\ref{tab:uni_result_L2}

\begin{figure}[h!]
    \centering
    \includegraphics[width=0.3\linewidth]{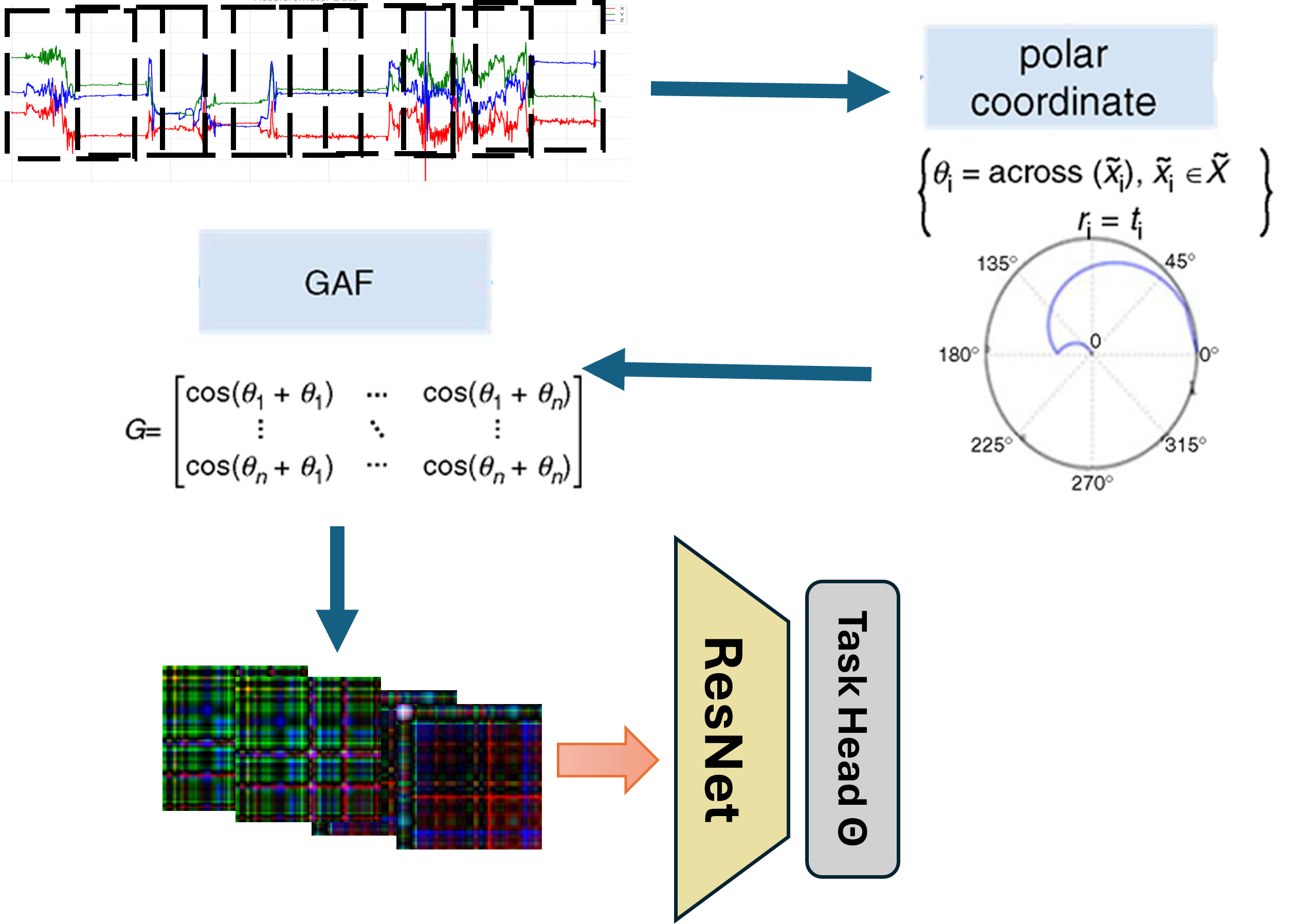}
    \caption{Overview of the GAF image pipeline}
    \label{fig:GAF-pipe}
    
\end{figure}

\begin{table}[!h]
\centering
\footnotesize
\begin{tabular}{@{}ccccc@{}}
\toprule
\multicolumn{3}{c}{ Average Class Accuracy of L1 Activities for ResNet34 GAF Images}   
\\ \midrule
\multicolumn{1}{c}{Data Modalities} & \multicolumn{1}{c}{Training Model} & \multicolumn{1}{c}{All Environments} \\
\midrule

IMU$_{Acc Gyro}$    & ResNet34      &   0.431     \\
EMG &  ResNet34       & 0.460    \\
Gaze &   ResNet34      & 0.442    \\
Insole total pressure &  ResNet34  &  0.422    \\
ECG & ResNet34   &   0.361  \\

\bottomrule
\end{tabular}
\vspace*{.1mm}
\caption{Unimodal benchmark results of L1 labels (Activities) for different modalities }
\label{tab:uni_result}
\end{table}
As evident from Figure~\ref{fig:Class distributions}, the granularity of activity recognition at the L2 level is significantly higher than at the L1 level. Consequently, the lower class-wise accuracy observed at the lower levels of the hierarchy is expected. This phenomenon has also been observed in other domains, such as multimodal medical image classification and disease detection \cite{10130287}, where increased granularity often leads to reduced accuracy.

\begin{table}[h]
\centering
\scriptsize
\begin{tabular}{@{}ccc@{}}
\toprule
\multicolumn{3}{c}{ Average Class Accuracy of L2 Activities for ResNet34 GAF Images}   
\\ \midrule
\multicolumn{1}{c}{Data Modalities}  & \multicolumn{1}{c}{Training Model} & \multicolumn{1}{c}{All Environments} \\
\midrule

IMU$_{Acc}$    & ResNet34 &     0.382     \\
EMG &    ResNet34 &  0.221    \\
Gaze &   ResNet34 &  0.172    \\
Insole total pressure &  ResNet34 &   0.234    \\
ECG & ResNet34 &  0.192  \\

\bottomrule
\end{tabular}
\vspace*{.1mm}
\caption{ Unimodal benchmark results of L2 labels (Actions) for different modalities }
\label{tab:uni_result_L2}
\end{table}
\paragraph{Visual data}
For RGB and Depth data, we used pretrained R3D\textunderscore18 \cite{tran2018closer}, MVit \cite{li2022mvitv2} and Swin Transformer \cite{liu2022video} Kinetics-400 weights, fine-tuning it for our dataset. The input to these models consists of 16 randomly selected RGB/Depth frame from each activity video sample stacked together as a 3D vector with temporal information. We have tested 24 frame for our subsequence length but the performance did not changed significantly. In Table~\ref{tab:visual-only-l1} we presented average class-wise accuracy on first level of DARai hierarchy.

\begin{table}[htb]
\centering
\footnotesize
\captionsetup[table]{font=scriptsize}
\resizebox{\textwidth}{!}{%
\begin{tabular}{cccccccccc}
\hline
& & & 
\multicolumn{1}{c}{\textbf{Class Category}}
& \multicolumn{2}{c}{\textbf{All}}
& \multicolumn{2}{c}{\textbf{Kitchen}}
& \multicolumn{2}{c}{\textbf{Living room}} \\
\cmidrule(lr){5-6}\cmidrule(lr){7-8}\cmidrule(lr){9-10}
\textbf{Data} & \textbf{Camera View} & \textbf{Level} & \textbf{Model} & \textbf{Finetuned} & \textbf{Trained from scratch} & \textbf{Finetuned} & \textbf{Trained from scratch} & \textbf{Finetuned} & \textbf{Trained from scratch} \\ \\
\multirow{6}{*}{\rotatebox[origin=c]{90}{RGB}} & $1$ & \multirow{6}{*}{L1} & \multirow{2}{*}{ResNet} & 0.89 & 0.89 $\pm$ 0.02 & 0.90 $\pm$ 0.004 & 0.39 $\pm$ 0.08 & 0.97 $\pm$ 0.01 & 0.75 $\pm$ 0.11 \\
 & $2$ &  &  & 0.75 & 0.72 $\pm$ 0.02 & 0.74 $\pm$ 0.02 & 0.49 $\pm$ 0.01 & 0.97 $\pm$ 0.003 & 0.82 $\pm$ 0.05 \\
 & $1$ &  & \multirow{2}{*}{Mvit-s} & 0.91 & 0.19 $\pm$ 0.05 & 0.86 & 0.18 $\pm$ 0.04 & 0.88 & 0.25 $\pm$ 0.02 \\
 & $2$ &  &  & 0.8 & 0.22 $\pm$ 0.11 & 0.64 & 0.35 $\pm$ 0.07 & 0.96 & 0.37 $\pm$ 0.14 \\
 & $1$ &  & \multirow{2}{*}{Swin-t} & 0.91 $\pm$ 0.005 & 0.49 $\pm$ 0.18 & 0.83 $\pm$ 0.02 & 0.56 $\pm$ 0.1 & 0.96 $\pm$ 0.008 & 0.43 $\pm$ 0.2 \\
 & $2$ &  &  & 0.77 $\pm$ 0.02 & 0.56 $\pm$ 0.06 & 0.54 $\pm$ 0.03 & 0.47 $\pm$ 0.04 & 0.87 $\pm$ 0.05 & 0.58 $\pm$ 0.1 \\
\midrule
\multirow{6}{*}{\rotatebox[origin=c]{90}{Depth}} & $1$ & \multirow{6}{*}{L1} & \multirow{2}{*}{ResNet} & \multirow{6}{*}{-} & 0.84 $\pm$ 0.04 & \multirow{6}{*}{-} & 0.77 $\pm$ 0.01 & \multirow{6}{*}{-} & 0.81 $\pm$ 0.01 \\
 & $2$ &  &  &  & 0.62 $\pm$ 0.01 &  & 0.49 $\pm$ 0.03 &  & 0.70 $\pm$ 0.01 \\
 & $1$ &  & \multirow{2}{*}{Mvit-s} &  & 0.49 $\pm$ 0.05 &  & 0.77 $\pm$ 0.1 &  & 0.36 $\pm$ 0.05 \\
 & $2$ &  &  &  & 0.50 $\pm$ 0.1 &  & 0.21 $\pm$ 0.02 &  & 0.54 $\pm$ 0.07 \\
 & $1$ &  & \multirow{2}{*}{Swin-t} &  & 0.59 $\pm$ 0.09 &  & 0.74 $\pm$ 0.03 &  & 0.64 $\pm$ 0.03 \\
 & $2$ &  &  &  & 0.48 $\pm$ 0.02 &  & 0.33 $\pm$ 0.03 &  & 0.51 $\pm$ 0.03 \\
\bottomrule
\end{tabular}%
}
\caption{ \scriptsize Top-1 accuracy results for visual data experiments using three model architectures on Activity level of DARai hierarchy. Results are reported separately for kitchen subset, living room subset, and both activity subsets together. Fine-tuned experiments are performed using models pretrained on Kinetics 400 dataset.}
\label{tab:visual-only-l1}
\end{table}

With increasing granularity of the data and shorter video samples in lower levels of the hierarchy we are observing a drastic performance drop in Table~\ref{tab:l2_visual_result}.
\begin{table}[!h]
\centering
\footnotesize
\begin{tabular}{@{}ccccc@{}}
\toprule
\multicolumn{5}{c}{ Average Class Accuracy of L2 and L3 Classess for visual data}   
\\ \midrule
\multicolumn{1}{c}{Data Modalities} & \multicolumn{1}{c}{Training Model} & \multicolumn{1}{c}{View} &\multicolumn{1}{c}{Level 2} & \multicolumn{1}{c}{Level3}  \\
\midrule

RGB    &   R3D & $cam_1$  &  0.30 & 0.029 \\
RGB    &   R3D &  $cam_2$ &  0.72 & 0.67 \\
RGB    &   mvit\_v2\_s & $cam_1$  & 0.027 &  0.017  \\
RGB    &   mvit\_v2\_s &  $cam_2$ &    0.55 &  0.60 \\
RGB    &   swin3d\_t &  $cam_1$ &   0.18 & 0.027 \\
RGB    &   swin3d\_t &  $cam_2$&    0.53 & 0.52\\

\bottomrule
\end{tabular}
\vspace*{.1mm}
\caption{Unimodal video understanding benchmark results on 50\% of L2 labels (33 class) and L3 labels (56 class)}
\label{tab:l2_visual_result}
\end{table}

\subsubsection{Multimodal analysis}
We utilized a subset of modalities from our previous experiments for multimodal activity recognition. In the first set of experiments, we used RGB sequences and depth sequences from the Kinect camera as inputs to the same video understanding model employed in the unimodal experiments. As anticipated, using both modalities together increased class-wise accuracy in the kitchen and merged environments. However, in the living room environment, the RGB results were already high, so combining it with depth only marginally improved the performance compared to using the depth modality alone.
In our other experiments using 1D IMU data and biomonitoring data in a multimodal setup, we tested window sizes on subsequences ranging from 3 to 6 seconds using a vanilla transformer. We aggregated features from each subsequence before making predictions. However, we were not successful in classifying L1 level classes. A significant challenge we encountered was the considerable variation in sequence lengths among L1 labels.
Additionally, these high-level activities often exhibited similar patterns in the time series data. For instance, many kitchen environment activities involve standing for most of the sequence duration. This similarity makes it difficult for transformers to develop robust feature representations at this level of the hierarchy, as the repetitive nature of the data does not provide distinctive features for effective classification. In vision there’s a lot of research done in fine grained variation analyses. For instance in explanations \cite{prabhushankar2020contrastive} , in prediction errors \cite{prabhushankar2024counterfactual}.
However, in multimodal time series activity classification, especially with high-level activities, there is often a lack of distinct temporal features, significant variability, and repetitive patterns. This makes it more challenging to develop effective models for capturing fine-grained variations that exist in datasets like DARai.

\begin{figure}[htbp]
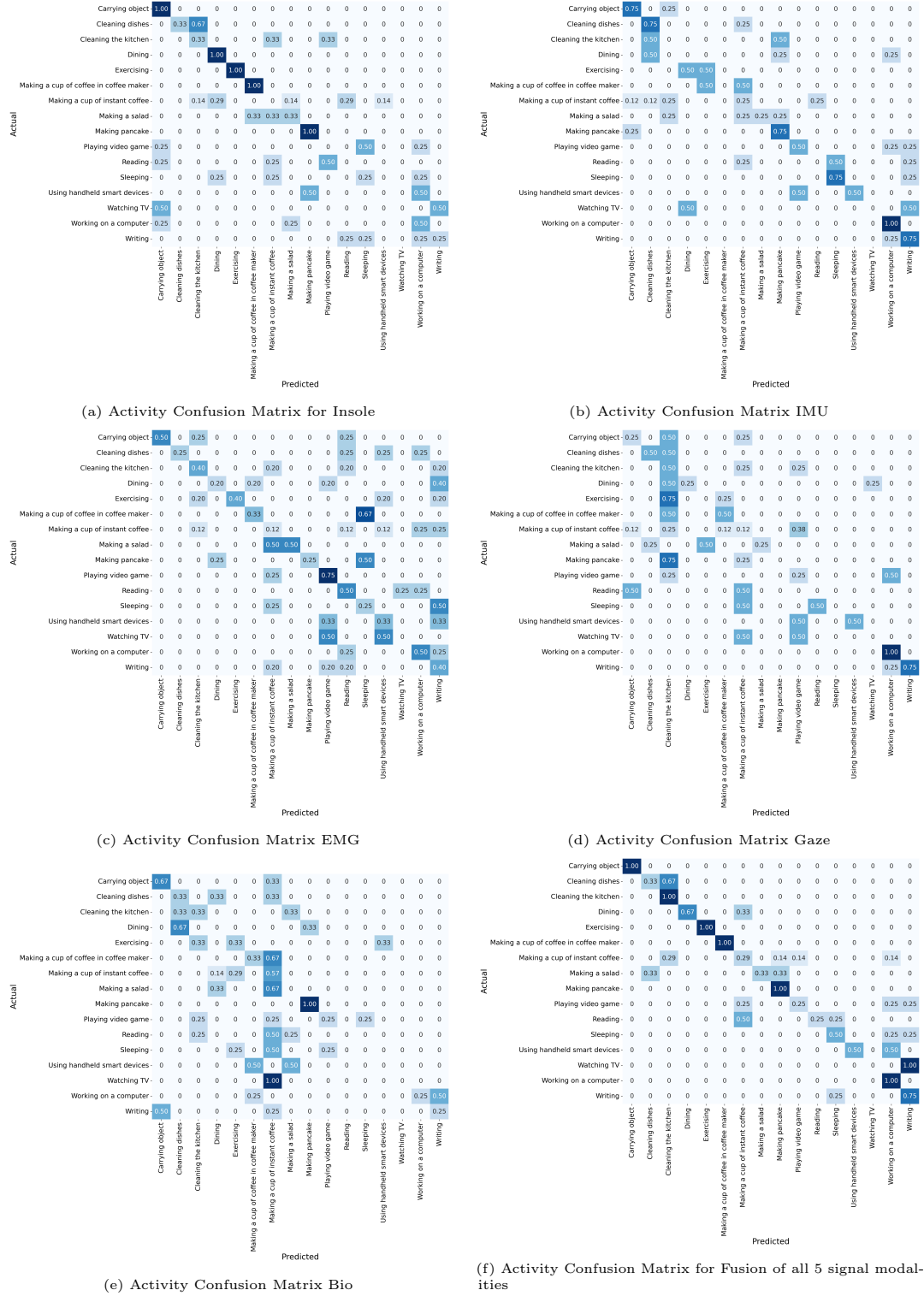

    \scriptsize
    \centering
    \begin{tabular}{cc}
        \begin{subfigure}[b]{0.45\textwidth}
            \centering
            \includegraphics[width=\textwidth]{figures/cm/fuse1/insole_cm.png}
            \caption{Activity Confusion Matrix for Insole}
            \label{fig:subfig1}
        \end{subfigure} &
        \begin{subfigure}[b]{0.45\textwidth}
            \centering
            \includegraphics[width=\textwidth]{figures/cm/fuse1/imu_cm.png}
            \caption{Activity Confusion Matrix IMU}
            \label{fig:subfig2}
        \end{subfigure} \\
        \begin{subfigure}[b]{0.45\textwidth}
            \centering
            \includegraphics[width=\textwidth]{figures/cm/fuse1/emg_cm.png}
            \caption{Activity Confusion Matrix EMG}
            \label{fig:subfig3}
        \end{subfigure} &
        \begin{subfigure}[b]{0.45\textwidth}
            \centering
            \includegraphics[width=\textwidth]{figures/cm/fuse1/gaze_cm.png}
            \caption{Activity Confusion Matrix Gaze}
            \label{fig:subfig4}
        \end{subfigure} \\
        \begin{subfigure}[b]{0.45\textwidth}
            \centering
            \includegraphics[width=\textwidth]{figures/cm/fuse1/bio_cm.png}
            \caption{Activity Confusion Matrix Bio}
            \label{fig:subfig5}
        \end{subfigure} &
        \begin{subfigure}[b]{0.45\textwidth}
            \centering
            \includegraphics[width=\textwidth]{figures/cm/fuse5/inertial_emg_insole_bio_gaze_cm.png}
            \caption{Activity Confusion Matrix for Fusion of all 5 signal modalities}
            \label{fig:subfig6}
        \end{subfigure} \\
    \end{tabular}
    \caption{Combining modalities to learn a shared representation space enables the model to learn from different patterns for the same activity classes and improves the accuracy of activity understanding.}
    \label{fig:unimodal-cm}
\end{figure}

\subsection{Temporal activity localization}
\label{sec:app-localization}
\paragraph{Motivation}
A key challenge of the DARAI dataset lies in its raw and unstructured nature, where activities unfold continuously without predefined boundaries. Temporal action segmentation focuses on identifying and segmenting distinct actions within these continuous sequences, even when transitions between actions are subtle or ambiguous. This task is particularly relevant for assessing the dataset's ability to support fine-grained temporal reasoning and model the complex, hierarchical structure of human behavior. For example, consider the actions of lifting a box and putting down a box as shown in Figure~\ref{fig:main-figure}. The majority of frames in these actions appear visually similar, making it challenging to determine precise boundaries. In addition to the shorter durations and imbalanced distribution of samples, a third challenge arises: the similarity of boundaries across samples at finer levels of hierarchy. By addressing this challenge, we aim to evaluate how well the dataset captures meaningful action transitions and provides sufficient cues for segmenting both high-level activities (L2 level) and detailed procedures (L3 level).

\begin{figure}
    \centering
    \includegraphics[width= 0.6\columnwidth]{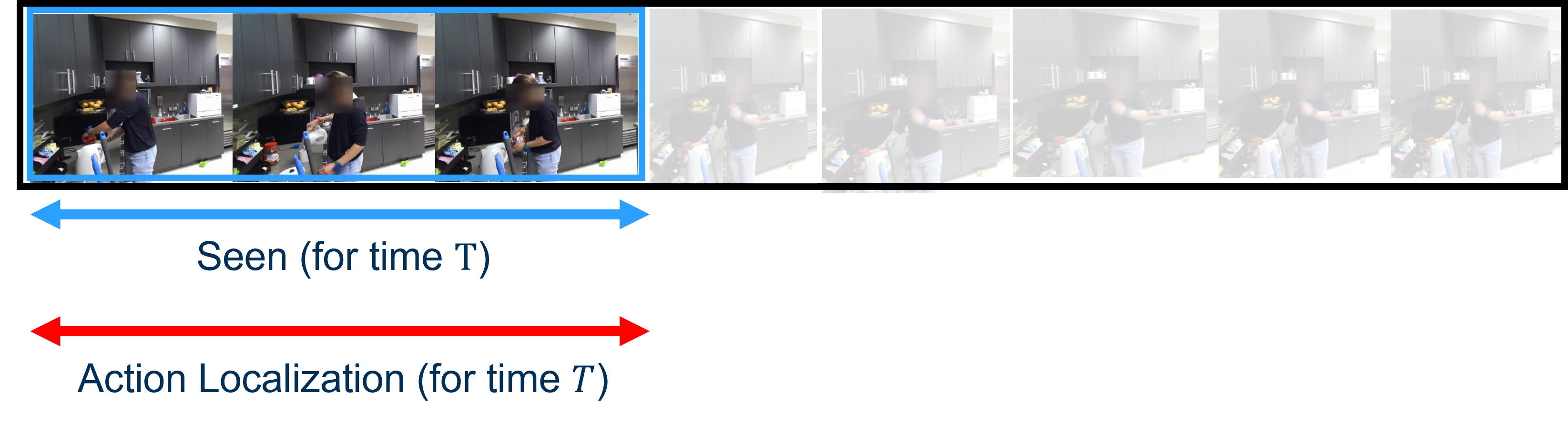}
    \caption{This visualization depicts the definition of segmentation task.}
    \label{fig:overview-segmentation}
\end{figure}

\paragraph{Task definitions}
\noindent We formulate the temporal activity localization problem as:
\[
\mathcal{X} = 
\begin{bmatrix}
x_{t-T} & \cdots & x_{t}
\end{bmatrix}^\top
= f(I_{t-T}, \cdots, I_{t}),
\]
where $\mathcal{X}$ is the temporal activity localization result, $T$ is the past time horizon, $x_t$ is the action label at time $t$, and $I_{t}$ is the image at time $t$. This is more illustrated in Figure~\ref{fig:overview-segmentation}. For this task, we used RGB frame features extracted from each sample video and a window size of eight consecutive frames to benchmark the decomposition of activity samples into lower-level steps (L2 and L3 labels) from different camera views. As the dataset was recorded at 15 frames per second (fps), we parsed the data by sampling one frame every 15 frames. This is because we want to reduce the input to 1 fps for computational efficiency while preserving the essential temporal structure. We used two different models: one is a Temporal Convolutional Network (TCN)-based model~\citep{singhania2021coarse}, and the other is an attention-based model~\citep{gong2022future}.

\noindent As noted in the main paper, we address two temporal activity localization tasks:
Action-level (L2 level) temporal activity localization and Procedure-level (L3 level) temporal activity localization.

\begin{figure}
    \centering
    \includegraphics[width= 0.6\columnwidth]{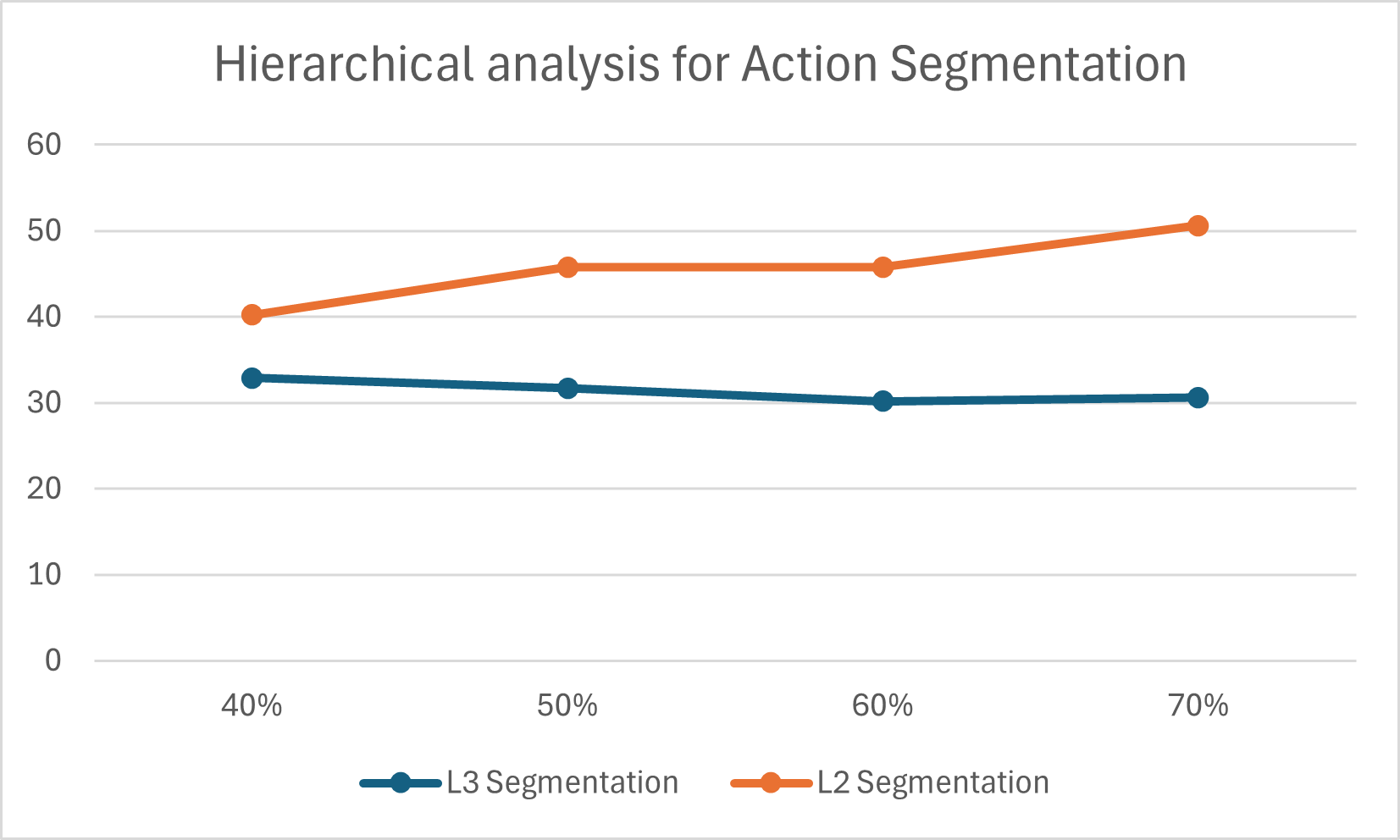}
    \caption{This visualization highlights the comparison between L2 segmentation performance and L3 segmentation performance.}
    \label{fig:hierarchical-segmentation}
\end{figure}

\begin{table}[htb]
\centering
\footnotesize
\resizebox{0.5\textwidth}{!}{%
\begin{tabular}{@{}ccccc@{}}
\toprule
\multicolumn{5}{c}{ Average MoF metric of L2 and L3 Classes for visual data}   
\\ \midrule
\multicolumn{1}{c}{Data Modalities} & \multicolumn{1}{c}{Model} & \multicolumn{1}{c}{View} &\multicolumn{1}{c}{Action (L2)} & \multicolumn{1}{c}{Procedure (L3)}  \\
\midrule
$RGB_{fine tuned}$  &   C2F-TCN & $1$  &  0.023 & 0.017 \\
$RGB_{fine tuned}$  &   C2F-TCN &  $2$ &  0.37 & 0.305 \\
\midrule
$RGB_{features}$   &   FUTR &  $1,2$ &    0.47 &  0.316  \\
\bottomrule
\end{tabular} %
}
\vspace*{.1mm}
\caption{ \scriptsize Comparison of MoF accuracy for localizing actions (L2) and procedures (L3) within their higher-level activity boundaries across different camera views. The model was fine-tuned on each camera view for activity recognition before the temporal localization experiment. The third-row multiview experiment was conducted using features extracted from both camera views prior to localization. `$1$' and `$2$' indicate different views and `$1,2$' indicates a multi views setup.}
\label{tab:segmentation_visual_result}
\end{table}

\begin{table}[htb]
\centering
\footnotesize
  \resizebox{0.8\textwidth}{!}{%
\begin{tabular}{@{}cccc@{}}
\toprule
\multicolumn{3}{c}{MoF Performance for Actions and Procedures with Partial Input Observation}   
\\ \midrule
\multicolumn{1}{c}{Input Sequence Length} &  \multicolumn{1}{c}{Action (L2)} & \multicolumn{1}{c}{Procedure (L3)}  \\
\midrule
$30\% $ of sequence  &  0.15 &  0.21  \\
$40\% $ of sequence   &  0.22 &  0.15 \\
$50\% $ of sequence   &   0.351 &  0.17 \\
$60\% $ of sequence   &   0.359&  0.23 \\
$70\% $ of sequence  &   0.45 &  0.26 \\
\bottomrule
\end{tabular} %
}
\vspace*{.1mm}
\caption{ \scriptsize Impact of input sequence length on localization MoF accuracy for actions (L2) and procedures (L3). The model was evaluated using different proportions of the entire input sequence of each activity sample. We used FUTR~\citep{gong2022future} model with multiple views to evaluate.}
\label{tab:segmentation_length_abalation}
\end{table}

\paragraph{Temporal Segmentation Challenges}
\noindent The results of these experiments are summarized in Table~\ref{tab:segmentation_visual_result}. Segmenting untrimmed procedural activities into lower-level action sequences is inherently difficult due to occlusion, camera viewpoints, and unstructured transitions. As seen in Table 1 of~\citep{10147035}, existing datasets such as Breakfast (MoF 0.69), 50Salads (MoF 0.78), and GTEA (MoF 0.77) achieve higher segmentation accuracy. In contrast, even under optimal camera views, DARai yields significantly lower MoF values (0.37 and 0.305), highlighting the increased difficulty of temporal segmentation in real-world, unstructured settings.

To further analyze temporal dependencies, we evaluated segmentation performance under varying observation lengths (Table~\ref{tab:segmentation_length_abalation}). We define "Input Sequence Length" as the proportion of an activity sample available for segmentation. The model was evaluated using 20\% to 70\% of the sequence, simulating cases where only partial observations are available. Results show a significant drop in segmentation accuracy when only a small portion of the sequence is observed, underscoring that actions and procedures are not learned in isolation but as interconnected components with strong temporal dependencies.

\paragraph{Hierarchical Analysis}
\noindent In Figure~\ref{fig:hierarchical-segmentation} and Tables~\ref{tab:l2-segmentation-short-term} and~\ref{tab:l3-segmentation-short-term}, we identify three key traits of the DARai dataset through a hierarchical analysis of segmentation performance.
\noindent First, L2 Segmentation consistently outperforms L3 Segmentation across all observation rates. This indicates that  L3 segmentation involves finer-grained actions, so it requires the model to identify detailed transitions and hierarchical dependencies.
\noindent Second, from 40\% to 70\%, L2 performance improves steadily as the observation rate increases. This indicates that dataset provides sufficient high-level cues for L2 segmentation as observation increases, so having access to more data allows the model to better understand and segment broader activities.
\noindent Third, L3 performance remains relatively flat across the observation rates. This lack of improvement highlights the inherent difficulty of segmenting fine-grained actions, even with additional observation data.
\noindent Lastly, while L2 segmentation benefits from general patterns and broader temporal structures, L3 segmentation struggles due to the intricate details and variability required to identify specific actions.

\begin{table}[h!]
\centering
\begin{tabular}{@{}lllllllll@{}}
\toprule
\textbf{Observed} & \textbf{20\%} & \textbf{30\%} & \textbf{40\%} & \textbf{50\%} & \textbf{60\%} & \textbf{70\%} & \textbf{80\%} & \textbf{90\%} \\ \midrule
10\% & 54.756 & 52.43  & 49.356 & 47.154 & 48.87  & 52.135 & 50.275 & 51.468 \\
20\% & 47.663 & 46.807 & 41.021 & 42.588 & 42.169 & 42.676 & 42.043 & 43.609 \\
30\% & 40.067 & 42.041 & 38.167 & 38.091 & 39.696 & 41.719 & 39.279 & 40.927 \\
40\% & 34.979 & 38.125 & 32.884 & 36.275 & 36.097 & 36.753 & 36.309 & 38.763 \\
50\% & 32.658 & 32.749 & 28.322 & 31.691 & 32.495 & 35.324 & 32.582 & 38.294 \\
60\% & 29.472 & 29.96  & 22.899 & 28.098 & 30.202 & 33.297 & 31.393 & 37.408 \\
70\% & 27.466 & 26.429 & 19.621 & 24.538 & 28.869 & 30.575 & 30.622 & 36.425 \\
80\% & 26.565 & 24.301 & 16.622 & 20.522 & 27.308 & 28.485 & 30.594 & 35.309 \\
90\% & 25.829 & 21.028 & 15.178 & 17.804 & 23.847 & 26.4   & 27.402 & 31.652 \\ \bottomrule
\end{tabular}
\caption{L2 level activity localization: Results obtained using FUTR and averaged over 3 fixed seeds for robustness. Each row corresponds to models trained on $n\%$ of the video sequence (e.g., 20\%, 30\%, ..., 90\%), and each column shows the inference performance when $m\%$ of the video sequence (e.g., 10\%, 20\%, ..., 90\%) is used as input. This evaluation highlights the performance across different combinations of training and observation rates.}
\label{tab:l2-segmentation-short-term}
\end{table}

\begin{table}[h!]
\centering
\begin{tabular}{@{}lllllllll@{}}
\toprule
\textbf{Observed} & \textbf{20\%} & \textbf{30\%} & \textbf{40\%} & \textbf{50\%} & \textbf{60\%} & \textbf{70\%} & \textbf{80\%} & \textbf{90\%} \\ \midrule
10\% & 29.558 & 31.641 & 24.706 & 29.867 & 24.935 & 22.796 & 26.955 & 18.611 \\
20\% & 40.079 & 32.944 & 27.441 & 34.169 & 26.453 & 27.238 & 28.165 & 18.744 \\
30\% & 33.826 & 37.039 & 34.789 & 42.592 & 37.249 & 35.288 & 33.995 & 25.42  \\
40\% & 29.353 & 29.968 & 40.209 & 47.453 & 40.977 & 40.652 & 39.126 & 28.955 \\
50\% & 19.472 & 25.892 & 32.547 & 45.782 & 43.382 & 44.664 & 42.311 & 36.045 \\
60\% & 17.751 & 24.318 & 29.496 & 43.547 & 45.794 & 49.354 & 47.4   & 37.66  \\
70\% & 15.765 & 17.041 & 28.266 & 39.629 & 39.235 & 50.616 & 49.501 & 46.832 \\
80\% & 13.834 & 16.559 & 26.006 & 36.096 & 36.707 & 43.021 & 49.095 & 49.735 \\
90\% & 10.871 & 15.86  & 22.887 & 35.192 & 35.944 & 45.694 & 43.52  & 47.696 \\ \bottomrule
\end{tabular}
\caption{L3 level activity localization: Results obtained using FUTR and averaged over 3 fixed seeds for robustness. Each row corresponds to models trained on $n\%$ of the video sequence (e.g., 20\%, 30\%, ..., 90\%), and each column shows the inference performance when $m\%$ of the video sequence (e.g., 10\%, 20\%, ..., 90\%) is used as input. This evaluation highlights the performance across different combinations of training and observation rates.}
\label{tab:l3-segmentation-short-term}
\end{table}

\subsection{Activity Anticipation}
\paragraph{Motivation} 
A defining characteristic of the DARAI dataset is its continuous and untrimmed nature. Participants were given a list of activities to perform but were not instructed on how or when to execute specific procedures. Due to this unstructured format, we aim to assess the dataset's ability to capture temporal dependencies and handle varying levels of abstraction—ranging from broad activity categories to fine-grained action details.
\begin{figure}
    \centering
    \includegraphics[width= 0.6\columnwidth]{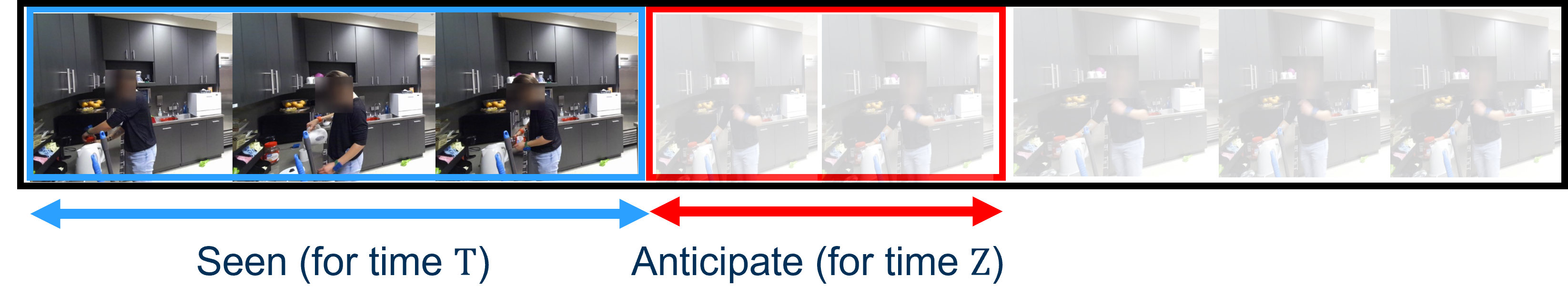}
    \caption{This visualization depicts the definition of anticipation task.}
    \label{fig:anticipation-overview}
\end{figure}
\paragraph{Formal tasks definitions}
\noindent We formulate the temporal activity localization problem as:
\[
\mathcal{X} = 
\begin{bmatrix}
x_{t+Z} & \cdots & x_{t+1}
\end{bmatrix}^\top
= f(I_{t-T}, \cdots, I_{t}),
\]
where $\mathcal{X}$ is the temporal activity anticipation result, $T$ is the past time horizon, $x_t$ is the action label at time $t$, and $I_{t}$ is the image at time $t$. As the dataset was recorded at 15 frames per second (fps), we parsed the data by sampling one frame every 15 frames, effectively reducing the input to 1 fps for computational efficiency while preserving the essential temporal structure. As illustrated in Figure~\ref{fig:anticipation-overview}, Action Anticipation involves using a portion of a video as input to predict future activities.
As noted in the main paper, we define four activity anticipation tasks:
short-horizon anticipation, long-horizon action anticipation, L2-level anticipation, L3-level anticipation.
\noindent First we compare short-horizon action anticipation performance with long-horizon action anticipation. For short-horizon action anticipation, we use $n\%, (n=20,30,…,90)$ of the video as input and anticipated the next 8 seconds as output. The long-horizon anticipation experiment extends the temporal scope to predict the final 8 seconds of the video given $n\%, (n=20,30,…,90)$ of the observed sequence as input. We then repeat the process with the ground truth of L2 level label and L3 level label. To ensure consistency, we report
average performance across 3 number of iteration, each with
fixed seeds 1, 10, 13452.

\begin{figure}
    \centering
    \includegraphics[width= 0.6\columnwidth]{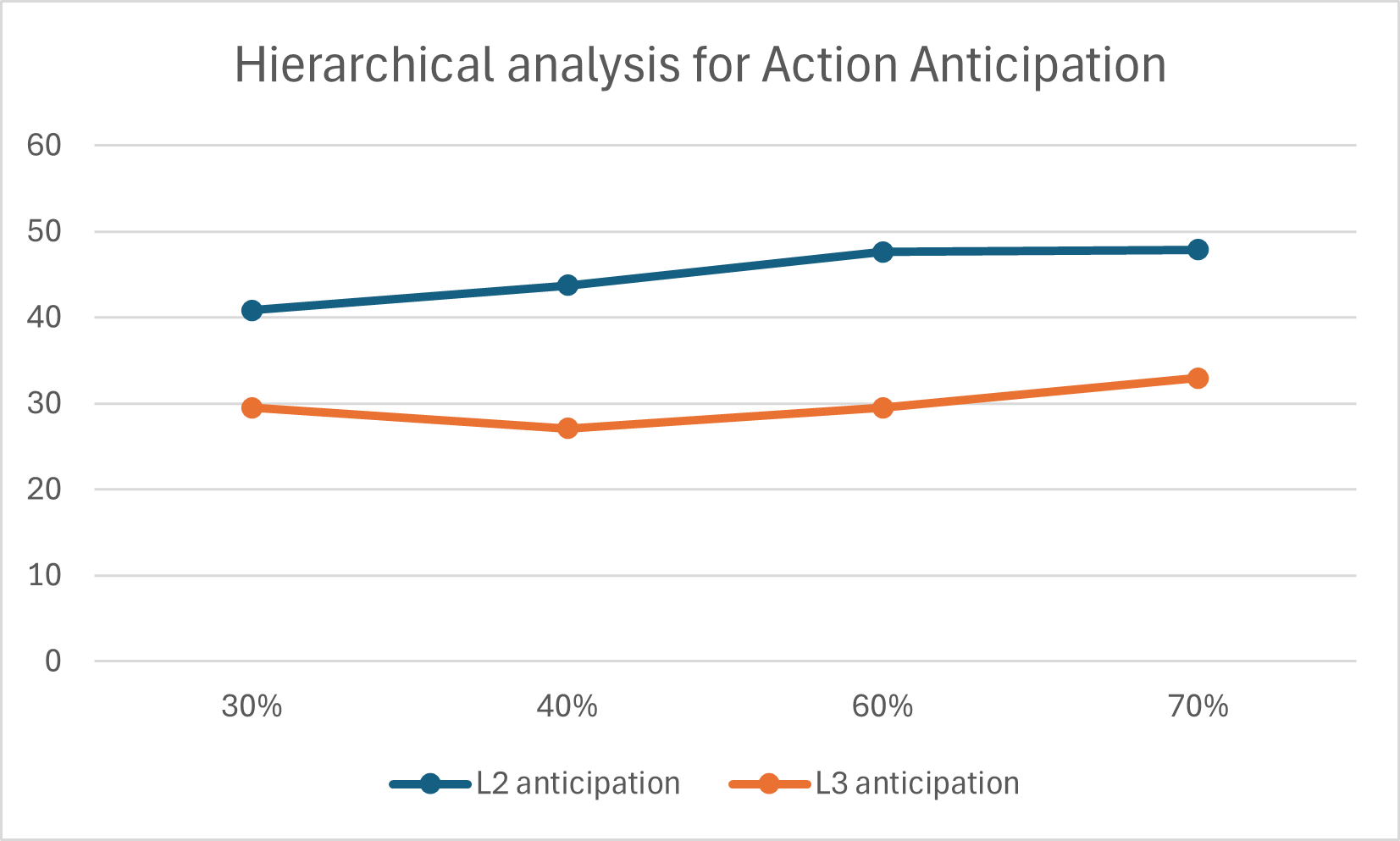}
    \caption{This visualization highlights the comparison between L2 anticipation performance and L3 anticipation performance.}
    \label{fig:hierarchical-anticipation}
\end{figure}

\begin{table}[h!]
\centering
\begin{tabular}{@{}lllllllll@{}}
\toprule
\textbf{Observed} & \textbf{20\%} & \textbf{30\%} & \textbf{40\%} & \textbf{50\%} & \textbf{60\%} & \textbf{70\%} & \textbf{80\%} & \textbf{90\%} \\ \midrule
10\% & 42.886 & 43.539 & 32.142 & 30.917 & 19.362 & 22.916 & 30.958 & 10.224 \\
20\% & 47.373 & 43.207 & 35.189 & 38.305 & 23.231 & 27.03  & 33.771 & 17.121 \\
30\% & 35.801 & 40.826 & 39.968 & 52.748 & 35.539 & 33.35  & 42.979 & 25.49  \\
40\% & 31.862 & 33.823 & 43.75  & 53.063 & 46.772 & 35.539 & 49.142 & 28.063 \\
50\% & 25.245 & 31.74  & 36.519 & 49.142 & 46.323 & 48.284 & 52.941 & 34.436 \\
60\% & 25.71  & 29.044 & 30.539 & 45.367 & 47.647 & 45.465 & 51.838 & 35.367 \\
70\% & 25.49  & 27.573 & 30.392 & 39.093 & 43.137 & 47.881 & 52.847 & 38.972 \\
80\% & 21.078 & 21.568 & 27.083 & 29.167 & 34.477 & 39.542 & 46.373 & 43.869 \\
90\% & 16.965 & 19.383 & 21.75  & 25.92  & 30.997 & 38.088 & 38.886 & 50.147 \\ \bottomrule
\end{tabular}
\caption{Short Horizon L2-level Anticipation: Results obtained using FUTR and averaged over 3 fixed seeds for robustness. Each row corresponds to models trained on $n\%$ of the video sequence (e.g., 20\%, 30\%, ..., 90\%), and each column shows the inference performance when $m\%$ of the video sequence (e.g., 10\%, 20\%, ..., 90\%) is used as input. This comprehensive evaluation highlights the performance across different combinations of training and observation rates.}
\label{tab:l2-short-anticipation}
\end{table}

\begin{table}[h!]
\centering
\begin{tabular}{@{}lllllllll@{}}
\toprule
\textbf{Observed} & \textbf{20\%} & \textbf{30\%} & \textbf{40\%} & \textbf{50\%} & \textbf{60\%} & \textbf{70\%} & \textbf{80\%} & \textbf{90\%} \\ \midrule
10\% & 40.441 & 42.034 & 34.436 & 34.558 & 28.501 & 28.308 & 32.86  & 12.934 \\
20\% & 36.274 & 39.338 & 36.887 & 31.372 & 23.757 & 28.431 & 32.247 & 14.093 \\
30\% & 22.671 & 32.72  & 34.558 & 38.48  & 32.58  & 33.21  & 37.027 & 19.485 \\
40\% & 21.813 & 23.039 & 30.637 & 34.926 & 32.948 & 32.72  & 40.441 & 26.467 \\
50\% & 18.522 & 20.955 & 23.039 & 30.024 & 34.646 & 35.784 & 42.051 & 30.654 \\
60\% & 19.187 & 21.2   & 20.71  & 27.888 & 34.751 & 39.828 & 47.303 & 30.707 \\
70\% & 18.207 & 20.36  & 21.271 & 23.109 & 28.448 & 39.46  & 46.936 & 42.909 \\
80\% & 17.717 & 18.627 & 23.231 & 23.354 & 28.693 & 33.578 & 44.905 & 44.87  \\
90\% & 16.001 & 18.259 & 19.187 & 23.371 & 26.855 & 35.049 & 37.447 & 46.025 \\ \bottomrule
\end{tabular}
\caption{Long Horizon L2-level Anticipation: Results obtained using FUTR and averaged over 3 fixed seeds for robustness. Each row corresponds to models trained on $n\%$ of the video sequence (e.g., 20\%, 30\%, ..., 90\%), and each column shows the inference performance when $m\%$ of the video sequence (e.g., 10\%, 20\%, ..., 90\%) is used as input. This comprehensive evaluation highlights the performance across different combinations of training and observation rates.}
\label{tab:l2-long-term-anticipation}
\end{table}

\begin{table}[h!]
\centering
\begin{tabular}{@{}lllllllll@{}}
\toprule
\textbf{Observed} & \textbf{20\%} & \textbf{30\%} & \textbf{40\%} & \textbf{50\%} & \textbf{60\%} & \textbf{70\%} & \textbf{80\%} & \textbf{90\%} \\ \midrule
10\% & 36.397 & 34.558 & 38.848 & 34.803 & 35.049 & 32.843 & 37.132 & 22.058 \\
20\% & 34.313 & 33.578 & 37.377 & 35.171 & 31.862 & 31.495 & 31.004 & 22.916 \\
30\% & 24.877 & 29.534 & 32.352 & 31.495 & 29.289 & 30.147 & 25.49  & 17.769 \\
40\% & 16.053 & 25.735 & 27.083 & 31.25  & 30.147 & 30.514 & 23.529 & 20.588 \\
50\% & 13.848 & 21.323 & 24.264 & 24.754 & 23.406 & 26.96  & 18.75  & 15.073 \\
60\% & 13.848 & 21.813 & 22.549 & 25.735 & 29.534 & 32.965 & 21.691 & 17.769 \\
70\% & 10.924 & 15.546 & 20.01  & 21.358 & 23.949 & 32.948 & 24.037 & 23.581 \\
80\% & 10.686 & 8.263  & 13.186 & 15.735 & 18.648 & 27.622 & 30.318 & 30.252 \\
90\% & 10.419 & 7.026  & 10.492 & 12.347 & 13.343 & 19.069 & 22.123 & 26.146 \\ \bottomrule
\end{tabular}
\caption{Short Horizon L3-level Anticipation: Results obtained using FUTR and averaged over 3 fixed seeds for robustness. Each row corresponds to models trained on $n\%$ of the video sequence (e.g., 20\%, 30\%, ..., 90\%), and each column shows the inference performance when $m\%$ of the video sequence (e.g., 10\%, 20\%, ..., 90\%) is used as input. This comprehensive evaluation highlights the performance across different combinations of training and observation rates.}
\label{tab:l3-short-term-anticipation}
\end{table}

\begin{table}[h!]
\centering
\begin{tabular}{@{}lllllllll@{}}
\toprule
\textbf{Observed} & \textbf{20\%} & \textbf{30\%} & \textbf{40\%} & \textbf{50\%} & \textbf{60\%} & \textbf{70\%} & \textbf{80\%} & \textbf{90\%} \\ \midrule
10\% & 21.813 & 22.181 & 25.49  & 22.303 & 23.529 & 23.529 & 25.735 & 13.97  \\
20\% & 22.303 & 22.549 & 25.49  & 25.735 & 22.916 & 23.774 & 25.122 & 14.95  \\
30\% & 17.647 & 19.73  & 25.49  & 25.49  & 23.529 & 26.593 & 25.367 & 20.22  \\
40\% & 13.112 & 15.563 & 22.794 & 23.774 & 22.916 & 27.205 & 26.47  & 24.632 \\
50\% & 12.622 & 14.828 & 18.75  & 21.568 & 22.916 & 26.593 & 27.696 & 27.696 \\
60\% & 10.049 & 11.887 & 16.544 & 19.975 & 21.568 & 27.328 & 26.966 & 28.676 \\
70\% & 10.359 & 9.558  & 13.602 & 16.299 & 14.215 & 24.632 & 26.102 & 29.166 \\
80\% & 11.642 & 8.823  & 11.764 & 13.97  & 12.132 & 21.446 & 24.019 & 30.882 \\
90\% & 12.009 & 7.72   & 9.681  & 12.132 & 11.764 & 18.995 & 19.607 & 27.328 \\ \bottomrule
\end{tabular}
\caption{Long Horizon L2-level Anticipation: Results obtained using FUTR and averaged over 3 fixed seeds for robustness. Each row corresponds to models trained on $n\%$ of the video sequence (e.g., 20\%, 30\%, ..., 90\%), and each column shows the inference performance when $m\%$ of the video sequence (e.g., 10\%, 20\%, ..., 90\%) is used as input. This comprehensive evaluation highlights the performance across different combinations of training and observation rates.}
\label{tab:l3-long-term-anticipation}
\end{table}

\paragraph{Benchmark} \noindent We aim to maintain similarities with other publicly available human action anticipation datasets, such as Breakfast and 50 Salads, which are widely used in the field. These datasets are typically collected in controlled environments, where participants perform predefined activities with relatively structured action sequences. In contrast, the DARai dataset introduces additional challenges due to its unstructured and naturally occurring activities. Unlike Breakfast and 50 Salads, where action transitions follow a more predictable order (e.g., sequential steps in preparing a meal or making a salad), DARai captures human activities in a more spontaneous and unconstrained setting. This results in higher variability and increased complexity in temporal reasoning, making action anticipation more difficult.

As shown in the table below, our dataset achieves comparable yet slightly lower performance than these more structured datasets. This performance gap highlights the unique challenges of DARai, where models must handle sparser observations, more diverse activity sequences, and greater unpredictability in human behavior. These factors make DARai a valuable benchmark for testing models' ability to generalize beyond controlled environments and anticipate actions in realistic, unconstrained settings. 
\begin{table}[h!]
\centering
\begin{tabular}{@{}lcc@{}}
\toprule
\textbf{Dataset}  & \textbf{Anticipate 10\%} & \textbf{Anticipate 20\%} \\ \midrule
Breakfast         & 27.70                   & 24.55                   \\
50Salads          & 39.55                   & 27.54                   \\
DARAI             & 25.75                   & 25.70                   \\ \bottomrule
\end{tabular}
\caption{Action Anticipation by observing 20\% of the video sequence.}
\label{tab:anticipation_comparison}
\end{table}
\subsection{Computational Resources}
We utilized an 8GB NVIDIA GeForce RTX 3060 and a 12GB NVIDIA TITAN RTX to run our models. Given the considerable size of our data, disk space may present a bottleneck for users who wish to use the entire dataset.

\end{document}